\documentclass{article}

\usepackage[main, final]{latex/neurips_2026}

\usepackage[utf8]{inputenc} %
\usepackage[T1]{fontenc}    %
\usepackage{hyperref}       %
\usepackage{url}            %
\usepackage{booktabs}       %
\usepackage{amsfonts}       %
\usepackage{nicefrac}       %
\usepackage{microtype}      %
\usepackage{xcolor}         %

\usepackage{soul}

\usepackage[utf8]{inputenc} %
\usepackage[T1]{fontenc}    %
\usepackage{xcolor}         %
\usepackage[dvipsnames]{xcolor}
\usepackage{amsmath}        %
\usepackage{amsthm}         %
\usepackage{amsfonts}       %
\usepackage{amssymb}        %
\usepackage{mathrsfs}       %
\usepackage{dsfont}         %
\usepackage{bm}             %
\usepackage{mathtools}      %
\usepackage{pgf,tikz}       %
\usepackage{pgfplots}       %
\pgfplotsset{compat=1.18}
\usepackage{subcaption}     %
\usepackage{float}          %
\usepackage{natbib}         %
\usepackage{parskip}        %
\usepackage{geometry}       %
\usepackage{hyperref}       %
\hypersetup{colorlinks=true, linkcolor=cyan, citecolor=blue}
\usepackage[capitalize]{cleveref}       %
\usepackage{enumitem}       %
\usepackage{thmtools}
\usepackage{thm-restate}
\usepackage[ruled, lined]{algorithm2e}
\usepackage{fvextra}
\usepackage{csquotes}

\usetikzlibrary{positioning,arrows,shapes,shapes.arrows,arrows.meta}
\tikzset{%
>={Latex[width=1.5mm,length=2mm]},
vertex/.style={draw,circle,inner sep=0mm,semithick,minimum width=4mm},
point/.style = {circle, draw, inner sep=0.04cm,fill,node contents={}},
uvertex/.style={outer sep=0},
bidir/.style={<->,dashed, line width=0.25mm},
dir/.style={->, line width=0.25mm},
regime/.style={shape=rectangle,fill=black,inner sep=0pt,minimum size=3pt,draw},
action/.style={shape=circle,fill=black,inner sep=0pt,minimum size=8pt,draw},
node distance=1cm,
font=\scriptsize\sffamily
}

\definecolor{betterred}{RGB}{228,26,28}
\definecolor{betterblue}{RGB}{55,126,184}

\def\*#1{\bm{#1}}
\def\1#1{\mathcal{#1}}
\def\2#1{\mathscr{#1}}
\def\3#1{\mathbb{#1}}
\def\4#1{\mathrm{#1}}
\newcommand{\R}{\mathbb{R}}

\newcommand{\pis}{\pi^{\star}}

\newcommand{\g}[1][\theta]{g_{#1}}
\newcommand{\giY}[1][\theta]{g_{#1}(X_i, Y_i)}
\newcommand{\giy}[1][\theta]{g_{#1}(x_i, y_i)}
\newcommand{\gY}[1][\theta]{g_{#1}(X, Y)}
\newcommand{\gy}[1][\theta]{g_{#1}(x, y)}
\newcommand{\giF}[1][\theta]{g_{#1}(X_i, f(X_i))}
\newcommand{\gF}[1][\theta]{g_{#1}(X, f(X))}
\newcommand{\gf}[1][\theta]{g_{#1}(x, f(x))}
\newcommand{\hiy}[1][\theta]{\nabla g_{#1}(X_i, Y_i)}
\newcommand{\hy}[1][\theta]{\nabla g_{#1}(X, Y)}
\newcommand{\hif}[1][\theta]{\nabla g_{#1}(X_i, f(X_i))}
\newcommand{\hf}[1][\theta]{\nabla g_{#1}(X, f(X))}

\newcommand{\angles}[2][]{#1\langle#2 #1\rangle}
\newcommand{\tuple}[2][]{\angles[#1]{#2}}

\newcommand{\cip}{\stackrel{p}{\rightarrow}}
\newcommand{\cid}{\stackrel{d}{\rightarrow}}

\newcommand{\ci}{\perp \!\!\!\! \perp }

\newcommand{\E}{\mathds{E}}
\newcommand{\V}{\mathds{V}}
\newcommand{\p}{\1P}
\newcommand{\cov}{\mathds{C}}

\newcommand{\tvar}{\text{Var}}
\newcommand{\bigparens}[1]{\bigg( #1 \bigg)}
\newcommand{\bigbracks}[1]{\bigg[ #1 \bigg]}

\newcommand{\commadots}{\ , \ldots  , \ }
\newcommand{\gap}{\hspace{.5em}}

\newcommand{\al}[1]{
	\begin{align*}
		#1
	\end{align*}
}

\newcommand{\flal}[1]{
	\begin{flalign*}
		#1
	\end{flalign*}
}

\declaretheoremstyle[
    spaceabove=4pt plus 3pt minus 2pt,
    spacebelow=4pt plus 3pt minus 2pt,
    headfont=\bfseries, bodyfont=\itshape, notefont=\normalfont, headpunct={.},
]{spacedplain}
\declaretheorem[style=spacedplain, name=Lemma]{lemma}
\declaretheorem[style=spacedplain, name=Theorem]{theorem}
\declaretheorem[style=spacedplain, name=Corollary]{corollary}

\declaretheorem[style=spacedplain, name=Assumption]{assumption}
\declaretheorem[style=spacedplain, name=Definition]{definition}
\declaretheorem[style=spacedplain, name=Remark]{remark}
\declaretheorem[style=spacedplain, name=Example]{example}

\Crefname{assumption}{Assumption}{Assumptions}

\title{Fixed-Size Active Statistical Inference}

\author{%
  Erik Skalnes \\
  Department of Computer Science\\
  Johns Hopkins University \\
  Baltimore, MD, 21218 \\
  \texttt{eskalne1@jhu.edu} \\
  \And
  Michael Oberst \\
  Department of Computer Science\\
  Johns Hopkins University \\
  Baltimore, MD, 21218 \\
  \texttt{moberst@jhu.edu} \\
}

\begin{document}
\maketitle

\begin{abstract}
	Variance reduction techniques improve the efficiency of drawing conclusions from data.
	Recent proposals for variance-reduction use two complementary techniques to reduce variance: incorporating machine learning predictions into estimators and weighted sampling.
	A popular weighted sampling algorithm is Poisson Sampling (PS), which independently flips a biased coin at each unit and includes all heads in the sample.
	A fundamental flaw with PS is that it has a random sample size, which presents practical challenges and introduces variance into the estimator.
	We show that a fixed-size alternative, known in survey sampling as Conditional Poisson Sampling (CPS), is a drop-in replacement for PS on general multivariate Z-estimation problems, which include convex M-estimation considered in prior work.
	Notably, we show that the asymptotic variance of CPS is no worse than that of PS, and can be substantially better in realistic scenarios where predictions are imperfect.
	We illustrate these advantages on both synthetic and real datasets, and release a Python package implementing CPS.
	In light of our contributions, there is no need to suffer a variable sample size.
\end{abstract}

\section{Introduction}\label{intro}

Science generalizes from a small slice of the world to the world itself, and statistical inference quantifies the uncertainty in doing so.
The key input is data, and data is often expensive.
A biologist meticulously experiments on rats to assess a novel treatment; a radiologist reviews scans to evaluate a cancer-prediction model.
The expense means that scientists often have less data than they would like, so they are motivated to derive more information per point.
The goal is to derive the same conclusion from a smaller dataset as one would from a larger dataset.

Several recent proposals for \textit{active statistical inference} use predictions, often from machine learning models, to draw conclusions with less labeled data \citep{zrnic2024asi, ao2024predictionasi, li2025robustasi, angelopoulos2025costoptimalasi}.
These techniques draw conclusions with fewer labels by using two compatible approaches for reducing the variance of an estimator.
The first uses a cheap predictor to generate predictions on all units, including unlabeled ones, and then uses the expensive labels to correct the error of the cheap predictions \citep{cassel1976difference, sarndal1992model, robins1994estimation, angelopoulos2023ppi}.
The second focuses on \enquote{active} selection of examples: rather than using Simple Random Sampling (SRS), one samples units according to a policy designed to prefer points whose labels contain more expected information.
Recent work relies on a weighted sampling design known as Poisson Sampling (PS)~\citep{Tille2020book, Hajek1964cps} for preferential sampling.
In PS, a \enquote{policy} assigns each unit a probability of being labeled, and the set of points to label is determined by independently flipping a weighted coin for each unit.

The issue with using PS to obtain a sample is that it gives a \textit{random sample size}, because each unit is included independently.
This is a practical problem.
Labeling budgets are fixed, so a design that overshoots or undershoots the budget is infeasible or wasteful.
This is also a statistical problem, as we will show, because the random sample size increases the variance of the estimator.

We fix the variable-size problem with a drop-in replacement: Conditional Poisson Sampling (CPS).
The idea is to reject samples from a PS design until we achieve one of exactly the intended budget, though modern implementations achieve a significant speedup with dynamic programming (see \cref{app:cps-implementation}).
CPS has a rich history in the survey sampling literature~\citep{Hajek1964cps, Chen1994cps, Chen1997poisbinom, Berger1998var, Berger1998norm, Deville2000cps, Brewer2003high, Matei2005variance, Henderson2006variance, Qualite2008comparison, bertail2017-a7, boistard2017-68, Tille2023compsci} because it
implements weighted inclusion probabilities like PS, but with a fixed sample size.

We show that in the context of \enquote{prediction-powered} statistical inference problems (e.g., convex M-estimation incorporating predictions, as considered in~\citet{zrnic2024asi}), CPS yields asymptotically normal estimators, and thus easy-to-construct confidence intervals.

Crucially, we show that the asymptotic variance of CPS is no worse than that of PS for \textit{any selection probabilities for sampling units} and can be substantially better---because it eliminates the variance contributed by the random sample size (see \cref{app:criticize_PS} for an example).
Proving asymptotic normality for CPS requires machinery beyond the Lindeberg CLT that suffices for PS, precisely because the fixed-size constraint introduces dependence between sampled units. Instead, we rely on the foundational framework of \citet{Hajek1964cps}, who relates PS and CPS asymptotically, and lift the result from mean estimation to Z-estimation and from finite populations to super-population inference via the joint design--model framework of \citet{rubin-bleuer2005-b9}.

We characterize the variance advantage of CPS over PS as depending entirely on the covariance between the estimator and the random sample size (\cref{corr:variance_reduction_cps}).
This covariance is a sum of the errors in the estimating equations, weighted by the probability of not sampling unit $i$.
For mean estimation, the advantage of CPS over PS is zero when the predictor $f$ is conditionally unbiased.
Our experiments match the theory: the variance of CPS is not significantly worse than that of PS on any dataset we test, and
the cases when it is significantly better overlap with the cases when $f$ is poorly calibrated.
A possibly counter-intuitive secondary point is that SRS can have lower variance than CPS (and thus PS) when the policy weights are chosen based on a very bad $f$.
We provide reproducibility code at \url{https://github.com/oberst-lab/fixed-size-asi} and a Python package implementing CPS at \url{https://pypi.org/project/cps-sampling/}.
In light of our contributions, there is no need to suffer a variable sample size.

\section{Related Work}\label{sec:related_work}

The goal of design-based techniques is to find a without-replacement and fixed-size sampling algorithm with weights $\*\pi$ that minimize the variance of the estimator.
Two algorithms with a rich history in the survey sampling literature \citep{Tille2020book} have recently been introduced to the machine learning community.
The first is often called 'stratified sampling' because it stratifies the population and samples uniformly within each stratum a number of times proportional to the stratum size times the (estimated) within-stratum standard deviation \citep{fogliato2024stratification, fisch2024stratifiedpredictionpoweredinferencehybrid, neyman1934allocation}.
The within-stratum sampling can be unequal or equal probability, making it compatible with other algorithms for weighted sampling.
The choice to stratify depends largely on whether one can sample preferentially towards high-variance strata, because it reduces variance in the estimator when high-variance strata are oversampled. 
Knowing the variance is non-trivial, but sometimes possible with heuristics.

The second algorithm is called Poisson Sampling (PS), but it is not fixed-size so its use is theoretical in the survey-sampling literature \citep{Hajek1964cps, Tille2023compsci}.
\citet{zrnic2024asi} introduce active statistical inference, combining PS with the difference estimator for M-estimation. 
\citet{ao2024predictionasi} extend the framework to experimental design.
\citet{li2025robustasi} address robustness to misspecified inclusion probabilities.
\citet{angelopoulos2025costoptimalasi} optimize the cost--accuracy tradeoff for AI model evaluation.
All rely on PS; our contribution is to show that CPS is a drop-in replacement that is never worse, and often better, for the sampling component.

On the estimator side, the idea of correcting a cheap predictor with expensive labels appears independently in survey sampling as the generalized difference estimator \citep{cassel1976difference, sarndal1992model}, in causal inference as the augmented inverse-probability-weighted estimator \citep{robins1994estimation}, and in machine learning as prediction-powered inference \citep{angelopoulos2023ppi, angelopoulos2023ppi++}.
\citet{mani2025ppi} provide non-asymptotic analysis. 
These estimator-based techniques are orthogonal to the design and compose naturally with weighted sampling, as demonstrated by \citet{zrnic2024asi}.

\section{A Survey of Sampling}\label{sec:survey}

Survey sampling and active statistical inference both estimate a parameter $\theta$ by sampling $n \ll N$ points from a (typically large) population of size $N$. We index the population by $\1U = \{1 \commadots N\}$, where each unit $i$ has observed covariates $\*x_i \in \1X \subseteq \3R^k$ and an initially unobserved label $y_i \in \1Y \subseteq \3R$. The estimate $\hat\theta$ aggregates the labels of sampled points, and the core challenge of \textit{sampling design} is choosing which examples to \enquote{sample} (i.e., obtain a label for) based on covariates and an estimator of the parameter to be used after sampling.

A \textit{sample} $\*s$ is a length-$N$ vector whose $i$th entry satisfies $0 \leq s_i \leq N$.
A \textit{without-replacement (WoR) sample} restricts $s_i \in \{0,1\}$.
A \textit{sampling design} $\p(\*s)$ is a distribution over samples.
A \textit{fixed-size design of size $n$} assigns positive probability only to samples with exactly $n$ units: $\p(\*s) > 0 \implies \sum_{i\in\1U} s_i = n$.
In the without-replacement designs we consider, each $s_i \in \{0,1\}$ indicates whether unit $i$ is in the sample.
The full distribution $\p(\*s)$ over all $2^N$ possible samples can be hard to characterize, but the variance of standard estimators depends only on a partial view: the marginal and joint inclusion probabilities defined below.

\begin{definition}[Inclusion Probabilities]\label{def:pi}
	The inclusion probabilities, or policy, $\*\pi\in[0,1]^N$ form a length-$N$ vector of probabilities with elements $\pi_i := P(i\in\*s) \equiv \sum_{\*s:i\in\*s} \p(\*s)$ that represent the marginal inclusion probability for unit $i$. Similarly, the joint inclusion probabilities $\*\Pi$ form an $N\times N$ matrix whose entries are $\pi_{ij} := P(i,j\in\*s) \equiv \sum_{\*s:i,j\in\*s} \p(\*s)$ where $\pi_{ii} \equiv \pi_i \equiv P(i\in\*s)$.
\end{definition}

\subsection{Poisson and Conditional Poisson Sampling}

We typically restrict ourselves to considering inclusion probabilities that sum to $n$, since the sum of inclusion probabilities yields the expected sample size over random draws of $\*s$. A \textit{weighted design} satisfies $\pi_i \neq \frac{n}{N}$ for some $i$. A simple weighted design is Poisson Sampling (PS).

\begin{definition}[Poisson Sampling (PS)]\label{def:poisson}
	Poisson Sampling w.r.t.\ $\*\pi$ has design distribution
	\al{
		& \p_{\textup{ps}(\*\pi)}(\*s) = \prod_{i=1}^N \pi_i^{s_i} (1-\pi_i)^{1 - s_i}
		&&\tag{design distribution}
	}
	which is implemented by sampling each $s_i \in \{0, 1\}$ independently with probability $P(s_i = 1) = \pi_i$.

\end{definition}

Poisson Sampling is the foundation of several recent proposals for \textit{active statistical inference} \citep{zrnic2024asi, ao2024predictionasi, li2025robustasi, angelopoulos2025costoptimalasi}.
However, Poisson Sampling has a notable drawback.
The realized sample size is a random quantity, even though the expected sample size is $\sum_{i\in\1U} \pi_i$.
Variable sample size creates two related problems.
First, it poses a practical challenge in real-world settings when sampling budgets are fixed.
Second, it introduces variance into estimators, as we discuss later on.
Conditional Poisson Sampling (CPS) \citep{Hajek1964cps} is a weighted and without-replacement design that is also fixed-size.

\begin{definition}[Conditional Poisson Sampling (CPS)]\label{def:cps}
	Conditional Poisson Sampling w.r.t.\ $\*\pi$, with Poisson parameters $\*p = \phi_n(\*\pi)$ (\cref{def:cps_implementation}), has design distribution
	\al{
		& \p_{\textup{cps}(\*\pi)}(\*s)
		= \p_{\textup{ps}(\*p)}\bigparens{\*s \mid |\*s| = n}
		\propto
		\begin{cases}
			\p_{\textup{ps}(\*p)}(\*s) & \text{if } \sum_{i\in\1U} s_i = n
			\\ 0 & \text{else}
		\end{cases}
		&&\tag{design distribution}
	}
	so it can be implemented by rejection sampling: draw $\*s \sim \p_{\textup{ps}(\*p)}$ until $|\*s| = n$.
\end{definition}

Note that, due to the rejective nature of the Conditional Poisson Sampling design, the probabilities used in the underlying Poisson Sampling design differ from $\*\pi$.  \citet{Hajek1964cps} showed that when $\*\pi$ is used in the underlying Poisson design, the resulting bias vanishes asymptotically, but \citet{Chen1994cps} provided an algorithm for computing $\*p$ for the underlying PS algorithm so that the rejective process implements the desired $\*\pi$, a \enquote{holy grail of survey sampling research} \citep{Tille2023compsci}.

\begin{definition}[Chen's CPS Calibration Map $\phi_n$]\label{def:cps_implementation}
	For a target inclusion probability vector $\*\pi \in (0,1)^N$ with $\sum_i \pi_i = n$,
	the \emph{CPS calibration map} $\phi_n : (0,1)^N \to (0,1)^N$ returns the
	Poisson parameter vector $\*p = \phi_n(\*\pi)$ such that Conditional Poisson Sampling
	with parameters $\*p$ and target size $n$ yields marginal inclusion probabilities $\*\pi$.
\end{definition}

Details of $\phi_n$, the extension of \citet{Deville2000cps} to map $\*\pi$ to $\*\Pi$, and a computationally efficient implementation of the rejection procedure based on dynamic programming are described in \cref{app:cps-implementation}. Computing $\*p$ costs $O(nN)$ per iteration of Chen's fixed-point procedure, and the dynamic program costs $O(nN)$ time and space to build and $O(N)$ per draw, whereas PS costs $O(N)$ per draw with no setup.
Further intuition for the challenges associated with any without-replacement, weighted, fixed-size sampling algorithm is discussed in \cref{app:all3_is_hard}.

\subsection{Design-Based Analysis of Sampling Designs}

There are two common frameworks for analysis in survey sampling.
The \textit{design-based} framework holds the population of $N$ units fixed and the target of inference is a function of this population, like the finite-population mean $\mu_N := \frac{1}{N} \sum_{i\in\1U} y_i$.
The \textit{model-based} (or \textit{super-population}) framework treats each $(X_i, Y_i)$ as an i.i.d.\ draw from some distribution, and the target of inference is a functional of that distribution, like the population mean $\mu := \E_m[Y]$, where $\E_m$ denotes expectation under the model.

We start with the design-based perspective on parameter estimation. Here all randomness in $\hat\theta$ comes from the sampling step, and standard asymptotic results consider the behavior of a sampling design over a sequence of fixed populations. %
For example, directly computing the variance of CPS is computationally expensive, so a body of work exists to form consistent estimators for its variance based on asymptotic analyses of this form \citep{Brewer2003high, Henderson2006variance, Matei2005variance}.
These works rely on the framework introduced by \citet{Hajek1964cps}.
\begin{definition}[Hájek's Asymptotic Framework (HAF)]\label{def:hajek-framework}
	For a CPS design w.r.t.\ $\*\pi$, HAF considers
	\al{
		& d(\*\pi) := \sum_{i\in\1U} \pi_i(1-\pi_i) \rightarrow \infty
	}
\end{definition}
\Cref{def:hajek-framework} considers the effect of growing populations as long as the number of units that we could conceivably sample grows without bound.  In particular, it implies $n \rightarrow \infty$ because $d\leq\sum_{i=1}^N \pi_i = n$ and $N-n\rightarrow \infty$ because $d\leq\sum_{i=1}^N (1-\pi_i) = N-n$.  Thus, it implies $N\to\infty$.  We make a different assumption from a slightly different perspective, informally stated here.
\begin{assumption}[Constraints on $\*\pi$]\label{asmp:fixed-ratio}\label{asmp:constraints_on_pi_informal}
	There exist integer budgets $n = n_N$ and a ratio $r \in (0, 1)$ with $n_N / N \to r$ (e.g., $n_N = \lfloor rN \rfloor$) such that, for any population of size $N$, $\sum_i \pi_i^N = n_N$, and there exists a value $\gamma \in (0, 1/2)$ such that, for any population of size $N$, $\*\pi^N$ satisfies $\gamma \leq \pi_i^N \leq 1 - \gamma$ for all $i$.
\end{assumption}
In other words, we assume that each $\*\pi$ for each population in the sequence as $N$ grows implements the budget $n_N \approx rN$ and is bounded. Growing $N$ along with bounded $\*\pi$ then implies~\cref{def:hajek-framework}.
With this assumption, we are ready to discuss asymptotic results for CPS, where we consider its behavior as the population grows.

\section{The Case for CPS: Theoretical Results}\label{sec:cps}

The variance of the estimator has two sources.
\textit{Model} randomness comes from modeling the data $(X_i, Y_i)$ as i.i.d.\ draws from a distribution $Q$; we write $\E_m$ for expectations under this model, i.e., with respect to $Q$.
\textit{Design} randomness comes from sampling the fixed population of size $N$ to create $\*s$.
Since the target parameter $\theta_0$ is a property of $Q$, considering only design randomness lets us state convergence to the moving target $\theta_N$.
Furthermore, $\*\pi$ depends on $X$, so the policy itself is a random function under $Q$.
Intuitively, PS introduces design variance through its random sample size, and CPS removes that variance by conditioning on $\tilde{n} = n$.

Consider the general statistical inference setting of Z-estimation, of which convex M-estimation, considered in~\citet{zrnic2024asi}, is a special case.   In this setting, $\theta_0 \in \mathbb{R}^{\ell}$ is a parameter we would like to estimate, $\g : \mathcal{X} \times \mathcal{Y} \to \R^{\ell}$ is a continuously differentiable function with respect to $\theta$, and the finite population of size $N$ consists of samples $(X_i, Y_i)$ drawn i.i.d.\ from $Q$.  Here, the parameter $\theta_0$ is identified as the solution to the population estimating equation $G(\theta) = 0$, and the finite-population estimand $\theta_N$ is identified as the solution to the empirical estimating equation $G_N(\theta) = 0$, where $G(\theta) := \E_m[\gY], G_N(\theta) := \frac{1}{N} \sum_{i=1}^{N} \giY$.  For instance, if $\gY = \nabla L_{\theta}(X, Y)$ for some convex loss function $L$, then this setting is equivalent to finding parameters $\theta$ that perform loss minimization.

In this setting,~\citet{zrnic2024asi} propose to use a fixed predictor $f(X)$ in two ways. First, they form predictions of $Y$ on unlabeled examples, and use those imputed values of $Y$ to construct estimates of $\giY$ on unlabeled examples via $\giF$.  Second, they use PS with unequal probabilities to choose which samples to label, and combine the two into a generalized difference estimator of $G_N(\theta)$ via
\begin{align}
	\hat{G}_N(\theta) & := \frac{1}{N} \sum_{i=1}^{N} \frac{s_i}{\pi_i^{N}} (\giY - \giF) + \giF. \label{eq:z-estimation-ghn}
\end{align}
They construct $\hat{\theta}_N$ as the solution to $\hat{G}_N(\theta) = 0$.  They show that this approach yields asymptotically normal estimates of $\theta_0$ for convex M-estimation problems, and derive the limiting variance.

In this section, we derive similar results for general \enquote{prediction-powered} Z-estimators defined through~\cref{eq:z-estimation-ghn} for both PS and CPS designs (\cref{theorem:main_asymptotic_result_informal}), which enables us to directly compare their asymptotic variances, and we demonstrate that the asymptotic variance of CPS is less than or equal to the asymptotic variance of PS, for any set of inclusion probabilities $\*\pi$.  There are two technical challenges in doing so:  First, the design distribution of CPS is complex, and simple arguments that work for convergence of PS (e.g., application of the Lindeberg Central Limit Theorem, which requires independence of sampling indicators) cannot be applied directly. Second, while CPS is well-studied in the survey sampling literature, many asymptotic results are formulated in the design-based framework with a focus on mean estimation, and it is non-obvious how to translate those results into statements about Z-estimators based on~\cref{eq:z-estimation-ghn} that incorporate both model and design uncertainty.  We build upon \citet{rubin-bleuer2005-b9} to establish conditions under which a Z-estimator $\hat\theta_N$ that solves $\hat{G}_N(\theta) = 0$ in~\cref{eq:z-estimation-ghn} converges to a normal distribution
\al{
	& \sqrt{n}(\hat\theta_N - \theta_0) \cid \1N(0, \Sigma)
	\quad\text{ where }\quad
	\Sigma = J^{-1}(\Sigma_d + r\,\Sigma_m) J^{-\top}
}
where $J=\nabla G(\theta_0)$, $r := \lim_{N\rightarrow \infty} n/N \in (0,1)$,
$\Sigma_m$ is the variance of $G_N(\theta_0)$ around $G(\theta_0)$, and $\Sigma_d$ is the variance of $\hat{G}_N(\theta_N)$ around $G_N(\theta_N)$ for the design.  A core insight of~\citet{rubin-bleuer2005-b9} is that these sources of variance contribute independently to the limiting distribution, under certain conditions.
$\Sigma_m$ depends only on the population, so it is design-agnostic: it is identical under PS, CPS, and SRS.
Thus, the PS vs.\ CPS vs.\ SRS comparison lives in $\Sigma_d$, and the comparison reduces to a design-level computation doable in the finite-population world.
The joint framework absorbs the design result into a statement about $\theta_0$.

\Cref{sec:cps:mean} establishes the comparison for mean estimation, where it is most transparent.
Mean estimation in the design framework contains the core intuition that CPS is no worse than PS for any $\*\pi$.
\Cref{sec:cps:mest} lifts it to Z-estimation and, by extension, convex M-estimation. Formal definitions, regularity conditions, and proofs are deferred to \cref{app:theory_appendix}.

\subsection{Mean Estimation}\label{sec:cps:mean}

The Horvitz--Thompson (HT) estimator with inclusion probabilities $\pi_i\in(0,1)$ is unbiased for finite population mean estimation.

\begin{definition}[Horvitz--Thompson (HT) Estimator]\label{def:expansion-estimator}
	Given a sampling design $\p(\*s)$ w.r.t.\ $\*\pi$ and any function $w_i := w(x_i, y_i)$ of the observed data, the Horvitz--Thompson estimator (HT) for the finite population mean $\mu^{w}_N := \frac{1}{N} \sum_i w_i$ is given by
	\al{
		& \hat{\mu}^{w}_{\*\pi}
		:= \frac{1}{N}\sum_{i=1}^{N} \frac{s_i}{\pi_i} \cdot w_i
	}
\end{definition}

HT estimates the mean when $w(x,y) = y$, and is in some sense the best linear estimator \citep{godambe1965admissibility, Hanurav1968admissibility, royall1970finite, wu2020sampling}. We use the more general formulation with $w$ to make the connection to the difference estimator below explicit.

A common augmentation of HT incorporates auxiliary information from the $N-n$ unsampled points via a label-predictor $f:\1X\rightarrow\1Y$.
The resulting estimator, also the focus of \citet{zrnic2024asi}, is known as the generalized difference estimator in survey sampling \citep{cassel1976difference}.

\begin{definition}[Generalized Difference Estimator (DE)]\label{def:augmented-expansion-estimator}
	The generalized difference estimator w.r.t.\ $\tuple{\*\pi, \p(\*s)}$ and a fixed predictor $f(X)$ is given by
	\begin{equation*}
		\hat{\mu}_{\*\pi}
		:= \frac{1}{N}\sum_{i=1}^{N} \frac{s_i (y_i - f(x_i))}{\pi_i} + f(x_i)
	\end{equation*}
\end{definition}

DE's error is HT's error for $w_i = y_i-f(x_i)$ because $\hat{\mu}_{\*\pi} - \mu^{y}_N = \hat{\mu}_{\*\pi}^{y - f} - \mu^{y-f}_N$.
Since the errors are equal, the variances are equal, so results below stated for HT with some $w$ hold for the DE.
In the design framework, $w_i$ is fixed for arbitrary $w$ because the $N$ units are fixed. 
We adapt the following lemma from~\citet{Berger1998norm, Hajek1964cps} to our setting to show that PS and CPS yield consistent and asymptotically normal HT estimators with respect to the realized finite-population mean $\mu^w_N$.

\begin{lemma}[Design Consistency and Asymptotic Normality of HT Under (C)PS]\label{lemma:ht_design_informal}
	Consider a sequence of size-$N$ populations $\{Y_i, X_i\}_{i=1}^{N}$ associated with $\*\pi^N$ satisfying \cref{asmp:constraints_on_pi_informal}.
	Let $w$ be a function with $\limsup_N \frac{1}{N}\sum_{i=1}^N w_i^4 < \infty$, and suppose $\sigma^2_\bullet(\*\pi, N)$ is bounded away from zero.
	Denote by $\mu^w_N \coloneqq \frac{1}{N} \sum_{i=1}^N w_i$ the $w$-population mean, and $\hat\mu^w_{\*\pi}$ the HT estimator formed under a PS design $\p_{\textup{ps}(\*\pi)}$ or CPS design $\p_{\textup{cps}(\*\pi)}$.  Then $\hat{\mu}^w_{\*\pi} - \mu^w_N \cip 0$ as $N \to \infty$, and
	\begin{equation*}
		\frac{\sqrt{N}}{\sigma_\bullet(\*\pi, N)} \big(\hat{\mu}^w_{\*\pi} - \mu^w_N\big) \cid \mathcal{N}(0, 1) \quad \text{as } N \to \infty,
	\end{equation*}
	where the design-specific scaling $\sigma_\bullet(\*\pi, N)$ is given by
	\begin{align*}
		\sigma^2_{\textup{ps}}(\*\pi, N)  & = \frac{1}{N} \sum_{i=1}^N w_i^2 \, \frac{1 - \pi_i}{\pi_i},                &
		\sigma^2_{\textup{cps}}(\*\pi, N) & = \frac{1}{N} \sum_{i=1}^N (w_i - \pi_i \beta)^2\, \frac{1 - \pi_i}{\pi_i},
	\end{align*}
	where $\beta = \frac{\sum_j w_j (1 - \pi_j)}{\sum_k \pi_k (1 - \pi_k)}$, and where $\pi$ above should be understood to be $\pi^N$.
\end{lemma}

The proof is given in~\cref{app:theory_appendix}, where the above is restated in the formal framework of~\citet{rubin-bleuer2005-b9} as~\cref{lemma:ht_design}. As a consequence of~\cref{lemma:ht_design_informal}, for mean estimation, the asymptotic design variance of CPS is no worse, and can be substantially better, than the asymptotic design variance of PS.

\begin{restatable}[Design Variance Reduction]{corollary}{VarianceReductionCPSMean}\label{corr:variance_reduction_cps}
	Under the conditions of~\cref{lemma:ht_design_informal}, the design variance scalings $\sigma^2_{\textup{ps}}$ and $\sigma^2_{\textup{cps}}$ from~\cref{lemma:ht_design_informal} satisfy
	\begin{equation*}
		\sigma^2_{\textup{ps}}(\*\pi, N) - \sigma^2_{\textup{cps}}(\*\pi, N)
		= \frac{\big(\tfrac{1}{N} \sum_i w_i (1 - \pi_i)\big)^2}{\tfrac{1}{N} \sum_i \pi_i (1 - \pi_i)}
		\;\geq\; 0.
	\end{equation*}
\end{restatable}
The proof of the above is given in~\cref{app:mean_variance_reduction}.
Asymptotics in the design framework have the disadvantage that they are stated with respect to \enquote{moving targets}---finite-sample versions of the population parameters.
A fundamental motivation for lifting to the model-based framework is the construction of confidence intervals for fixed parameters of the distribution that generates the population. 
The lift requires $\pi^N$ to concentrate around a fixed function of $X$, treated formally in \cref{asmp:design_assumptions}.
This condition is satisfied by any policy that depends on a consistent plug-in estimator of a population parameter.
For instance, the condition is mild when $\*\pi$ is proportional to a fixed function $u(X)$ and then clipped, as in the probability proportional to size (PPS) policy used in \citet{zrnic2024asi}.

\subsection{Z-Estimation}\label{sec:cps:mest}

We give our main result, \cref{theorem:main_asymptotic_result_informal}, here informally, deferring the full regularity conditions and setup to~\cref{app:theory_appendix}.

\begin{restatable}[Asymptotic Variance of Z-Estimation with (C)PS]{theorem}{MainAsymptoticResult}\label{theorem:main_asymptotic_result_informal}
	Under regularity conditions given in the appendix (see~\cref{asmp:design_assumptions,asmp:model_assumptions}), and assuming that $\Sigma_{\infty,\bullet}$ below is positive definite,
	\begin{equation*}
		\sqrt{n} (\hat{\theta}_N - \theta_0) \cid \mathcal{N}(0, \Sigma), \quad \Sigma = J^{-1}(\Sigma_d + r \cdot \Sigma_{m}) J^{-\top}
	\end{equation*}
	where $\Sigma_m = \tvar(\gY[\theta_0])$ is shared across designs and $\Sigma_d = r \Sigma_{\infty,\bullet}$ for $\bullet \in \{\textup{ps, cps}\}$ is given, with $w^\star := \gY[\theta_0] - \gF[\theta_0]$ and $\pis := \pis(X)$, by
	\begin{align*}
		\Sigma_{\infty,\textup{ps}}
		:= \E_m\bigbracks{w^\star w^{\star\top}(1-\pis) / \pis},
		\quad
		\Sigma_{\infty,\textup{cps}}
		 := \E_m \bigbracks{\bigl(w^\star-\pis \beta\bigr)\bigl(w^\star-\pis \beta\bigr)^\top\,\frac{1-\pis}{\pis}}
	\end{align*}
	where $\beta := \frac{\E_m\bigl[w^\star\bigl(1-\pis\bigr)\bigr]}{\E_m\bigl[\pis\bigl(1-\pis\bigr)\bigr]}, J := \nabla_\theta G(\theta_0)$,
	and $r := \lim_{N \rightarrow \infty} n / N \in (0,1)$.
\end{restatable}

The same algebra as~\cref{corr:variance_reduction_cps} gives $\Sigma_{\infty,\textup{ps}} \succeq \Sigma_{\infty,\textup{cps}}$ (i.e., $\Sigma_{\infty,\textup{ps}} - \Sigma_{\infty,\textup{cps}}$ is positive semi-definite), so CPS dominates PS in the joint design--model variance for any fixed $\pis$. The proof of the following is given in \cref{app:M_variance_reduction}.

\begin{restatable}[Design Covariance Reduction]{corollary}{MatrixVarianceReductionCPS}
	\label{corr:matrix_variance_reduction_cps}
	Write $M \succeq 0$ if $M$ is positive semi-definite, i.e., $v^\top M v \geq 0$ for any vector $v$.
	Under the conditions of~\cref{theorem:main_asymptotic_result_informal}, the design
	limiting covariance matrices $\Sigma_{\infty,\textup{ps}}$ and $\Sigma_{\infty,\textup{cps}}$ satisfy
	\begin{equation*}
		\Sigma_{\infty,\textup{ps}} - \Sigma_{\infty,\textup{cps}}
		=
		\frac{
			\E_m\bigl[w^\star(1-\pis)\bigr]
			\E_m\bigl[w^\star(1-\pis)\bigr]^{\top}
		}{
			\E_m\bigl[\pis(1-\pis)\bigr]
		} \ \succeq \ 0.
	\end{equation*}
\end{restatable}

Digging deeper, write the numerator in \cref{corr:matrix_variance_reduction_cps} as $vv^\top$, with $v := \E_m[w^\star(1-\pis)]$.
CPS has no advantage over PS exactly when $v = 0$.
A sufficient condition is $\E_m[w^\star \mid X] = 0$, because the tower property gives $v = \E_m[\E_m[w^\star \mid X](1-\pis(X))]$.
For mean estimation, $\E_m[w^\star \mid X] = \E_m[Y - f(X) \mid X]$, so this condition says that $f$ is the true regression function.
In practice we expect $\E_m[w^\star \mid X] \neq 0$.
A weaker condition is $\E_m[w^\star] = 0$, which holds for mean estimation when $f$ is fit by least squares with an intercept on in-distribution data.
If $\E_m[w^\star] = 0$, then $v = -\4{cov}(w^\star, \pis)$, so $v = 0$ when the error of $f$ is uncorrelated with the probability that a unit is sampled.

When $v \neq 0$, there is a gap, and the gap comes from the random sample size of PS.
Under PS, the covariance between the HT estimator and the sample size $\tilde{n} = \sum_i s_i$ is $\frac{1}{N}\sum_i w_i(1-\pi_i)$, which converges to $v$ (\cref{app:covariance}).
The variance of $\tilde{n}$ is $\sum_i \pi_i(1-\pi_i)$, and $\frac{1}{N}$ times it converges to the denominator $\E_m[\pis(1-\pis)]$.
So the gap divided by $N$ equals $\4{cov}(\hat\mu, \tilde{n})^2 / \V[\tilde{n}]$: the part of the PS variance of $\hat\mu$ that a linear function of $\tilde{n}$ explains.
For mean estimation, a biased $f$ makes $v$ large.
Each extra sampled unit then adds its residual $w_i$ to the HT sum, so the estimate moves with $\tilde{n}$.
CPS fixes $\tilde{n}$ and removes this variance.
We investigate the gap empirically in the next section.

\section{Experiments}\label{sec:experiments}

Our experiments bolster our theoretical results, which hold in asymptopia, with evidence that CPS beats PS in realistic scenarios. 
We compare SRS, PS, and CPS, which correspond to the methods proposed in \citet{angelopoulos2023ppi} (PPI), \citet{zrnic2024asi} (ASI), and this paper. 
In \cref{tab:rmse_significance}, we report, for each dataset, the percent change of CPS relative to SRS/PS in RMSE and expected CI width, averaged first over $5000$ seeds then over the budgets $n \in \{100, 300, 500, 700, 1000\}$.  
We report RMSE in addition to expected CI width because it depends neither on the variance estimator nor on coverage, but see \cref{app:coverage} and \cref{tab:variance_quality} for details on coverage and variance estimation at different $n$. 
Significance comes from a paired bootstrap over the $5000$ seeds: we resample the seeds with replacement 1000 times, using the same resampled seeds for every design and budget, and recompute CPS's percent change against SRS/PS. 
A change is not significant ($\dagger$) if it is less than two bootstrap standard errors from $0$.
When the bootstrap distribution is normal, significance means that at least $97.5\%$ of the bootstrap replicates lie on the same side of $0$ as the cell's value.
A negative cell value means a reduction in favor of CPS. 
The rows of \cref{tab:rmse_significance} report the results averaged over budgets, whereas \cref{app:per_budget} gives further details of each experiment and its results at each budget.

In the \textbf{Synthetic} rows, we consider simulated populations with $\pi_i = n/N$, so PS differs from SRS and CPS only in its random sample size.
The first two Synthetic rows use HT to estimate the mean of a population with mean $0$ and with mean $2$, and the $f_{lin}$ and $f_{bias}$ rows use DE to estimate the mean of DGP0 with an unbiased or a biased $f$.
We find that, with uniform weights, the variance of PS using HT increases between DGP0 and DGP1 and also increases using DE on DGP0 between $f_{lin}$ and $f_{bias}$, but that the variance of CPS (and SRS) does not increase.
Thus, we credit the increase to the only difference between the three algorithms: that PS has a variable sample size.

In the \textbf{Pew} rows with shifted $f$, we estimate the mean approval rate of Biden and Trump from the Pew post-election survey, using DE with a logit-shifted $f$.
This is the same dataset and estimand as \enquote{ASI Pew-Biden/Trump}, but with a shifted $f$. 
Estimation with this very poor $f$, which informs the DE estimate and the selection of $\*\pi$, causes both CPS and PS to underperform SRS in \enquote{Pew-Trump, shifted $f$} where the bias is relatively large, but does not cause CPS to underperform PS in either row. 

In the \textbf{SAIPE} rows, named for the Census Bureau's Small Area Income and Poverty Estimates program, we consider a more realistic source of bias for $f$.
We estimate mean income in four US states with a sampling rule determined by $f$ trained on California, so $f$ is biased by the distribution shift.
We show that CPS gains most over PS in the states where $f$ is most biased, and that it shows significant gains over SRS in all states. 

In the \textbf{ASI} rows, we consider the datasets and setup of \citet{zrnic2024asi}, but with SRS in place of unweighted PS as the baseline.
The Census row estimates a regression coefficient, the Pew rows estimate mean approval, and the AlphaFold row estimates an odds ratio.
CPS matches PS on all of them.

\begin{table}[t]
\centering
\small
\begin{tabular}{lrrrrrr}
\toprule
 & & \multicolumn{2}{c}{RMSE vs.} & \multicolumn{2}{c}{$\mathbb{E}$[width] vs.} & \\
\cmidrule(lr){3-4} \cmidrule(lr){5-6}
Dataset & $f$ bias & SRS & PS & SRS & PS & CPS cov. \\
\midrule
Synthetic DGP0, HT & -- & $-1\%$$^\dagger$ & $-.2\%$$^\dagger$ & $-.1\%$ & $-.1\%$$^\dagger$ & 0.899 \\
Synthetic DGP1, HT & -- & $-1\%$$^\dagger$ & $-26\%$ & $-.1\%$ & $-26\%$ & 0.899 \\
Synthetic DGP0, $f_{lin}$, DE & -- & $-.7\%$$^\dagger$ & $+.9\%$$^\dagger$ & $+.1\%$$^\dagger$ & $0\%$$^\dagger$ & 0.899 \\
Synthetic DGP0, $f_{bias}$, DE & -- & $-.7\%$$^\dagger$ & $-30\%$ & $+.1\%$$^\dagger$ & $-29\%$ & 0.899 \\
Pew-Biden, shifted $f$ & $+35\%$ & $-11\%$ & $-16\%$ & $-12\%$ & $-17\%$ & 0.890 \\
Pew-Trump, shifted $f$ & $+115\%$ & $+86\%$ & $-13\%$ & $+80\%$ & $-14\%$ & 0.877 \\
SAIPE AL & $+29\%$ & $-15\%$ & $-3\%$ & $-16\%$ & $-4\%$ & 0.894 \\
SAIPE OR & $+23\%$ & $-15\%$ & $-4\%$ & $-15\%$ & $-3\%$ & 0.899 \\
SAIPE MA & $+8\%$ & $-15\%$ & $-.6\%$$^\dagger$ & $-15\%$ & $-.2\%$$^\dagger$ & 0.897 \\
SAIPE TX & $+14\%$ & $-15\%$ & $+.4\%$$^\dagger$ & $-16\%$ & $-1\%$ & 0.891 \\
ASI Census & $-19\%$ & $-42\%$ & $-2\%$$^\dagger$ & $-41\%$ & $-1\%$ & 0.894 \\
ASI Pew-Biden & $-1\%$ & $-9\%$ & $-.7\%$$^\dagger$ & $-11\%$ & $-.2\%$$^\dagger$ & 0.895 \\
ASI Pew-Trump & $-2\%$ & $-10\%$ & $+1\%$$^\dagger$ & $-12\%$ & $-.1\%$$^\dagger$ & 0.891 \\
ASI AlphaFold & $-7\%$ & $-.5\%$$^\dagger$ & $0\%$$^\dagger$ & $-11\%$ & $-1\%$$^\dagger$ & 0.864 \\
\bottomrule
\end{tabular}
\caption{
	For every experiment, we report the percent change in RMSE and expected CI width for CPS against SRS and PS, with $\dagger$ indicating when a difference is not significant (within $2\times$ Monte Carlo SE of 0 with paired seed bootstrap). 
	CPS is not significantly worse than PS on any dataset, and beats SRS on all but \enquote{Pew-Trump, shifted $f$} because the policy $\*\pi$ depends on the heavily biased $f$. 
	We report RMSE to show that the expected width comparison is not an artifact of variance estimation, and we report coverage to show similarly that CPS has coverage approximately at the target of $0.9$ on all datasets. 
	We report the bias of $f$ to show that it approximately tracks the degree of improvement of CPS over PS, and that SRS beats CPS when it is too large. 
}
\label{tab:rmse_significance}
\end{table}

\paragraph{Summary of the Table:}
CPS never has significantly higher RMSE than PS in any experiment, but has significantly lower RMSE than PS in $6$ experiments.
The RMSE of CPS is significantly lower than that of PS in \enquote{Synthetic DGP1, HT} by $26\%$, in \enquote{Synthetic DGP0, $f_{bias}$, DE} by $30\%$, in \enquote{Pew-Biden, shifted $f$} by $16\%$, in \enquote{Pew-Trump, shifted $f$} by $13\%$, in \enquote{SAIPE AL} by $3\%$, and in \enquote{SAIPE OR} by $4\%$.
In each of the experiments where CPS significantly reduces RMSE versus PS, it also significantly reduces expected-width in an amount approximately tracking the reduction in RMSE. 
CPS also significantly reduces the expected-width in two experiments without a significant reduction in RMSE, in \enquote{SAIPE TX} by $1\%$ and in \enquote{ASI Census} by $1\%$.
Furthermore, CPS significantly reduces RMSE over SRS in every experiment except the \enquote{Synthetic} ones because they use uniform weights, \enquote{Pew-Trump, shifted $f$} because the weights are selected with an $f$ with bias of $+115\%$, and \enquote{ASI AlphaFold} where the reduction is insignificant for RMSE but for expected-width has a significant reduction of $11\%$ but at $.864$ coverage. 
These results indicate that CPS is never significantly worse than PS, is often better, and usually beats SRS.

\paragraph{Synthetic Data: PS Loses Only When the Residuals Have Non-Zero Mean.}
We estimate a population mean with $\pi_i = n/N$ for every unit, so PS differs from SRS and CPS only in its random sample size.
In DGP0, $X \sim \1N(0,1)$ and $Y = 2X + \epsilon$ with $\epsilon \sim \1N(0,1)$, so $\E[Y] = 0$; DGP1 is the same except $X \sim \1N(1,1)$, so $\E[Y] = 2$.
On both, we estimate the mean with HT (rows \enquote{Synthetic DGP0, HT} and \enquote{Synthetic DGP1, HT}).
On DGP0, we also estimate the mean with DE, once with the unbiased predictor $f_{lin}(X) = 2X$ and once with the biased predictor $f_{bias}(X) = 2X + 1$ (rows \enquote{Synthetic DGP0, $f_{lin}$, DE} and \enquote{Synthetic DGP0, $f_{bias}$, DE}).
When the mean is zero, CPS and PS are equal: on DGP0 the RMSE of CPS is within $0.2\%$ of PS, and with $f_{lin}$ within $1\%$, and neither difference is significant.
When the mean is non-zero, CPS reduces the RMSE of PS by $26\%$ on DGP1 and by $30\%$ with $f_{bias}$.
The two cases are the same case, because DE is HT applied to the residuals $w_i = y_i - f(x_i)$, so a biased $f$ shifts $w$ exactly as a non-zero mean shifts $y$.
With uniform $\*\pi$, CPS and SRS are the same design, and their RMSEs differ by at most $1\%$ in every synthetic row, with no significant difference.
\Cref{fig:synthetic} shows the same pattern at each budget.

\paragraph{Pew with a Shifted Predictor: A Miscalibrated $f$ Inflates the Variance of PS.}
The Pew post-election survey asks whether a respondent approves of Biden or of Trump, and we estimate the mean approval with DE.
We start from the probabilistic classifier $f$ of \citet{zrnic2024asi} and add $b = 3$ to its logits, which gives $f_{bias}(X) = \sigma(\mathrm{logit}(f(X)) + 3)$.
This shift changes only the calibration of $f$, not its ability to rank respondents, as measured by AUROC.
We sample with $\pi_i \propto \min(f_{bias}(x_i), 1 - f_{bias}(x_i))$, so the shifted predictor sets both the residuals and the inclusion probabilities.
The shift makes $f$ over-predict the mean by $35\%$ for Biden and by $115\%$ for Trump (rows \enquote{Pew-Biden, shifted $f$} and \enquote{Pew-Trump, shifted $f$}).
CPS reduces the RMSE of PS by $16\%$ for Biden and by $13\%$ for Trump, and the expected width by $17\%$ and $14\%$.
Against SRS, the comparison is different.
For Biden, CPS beats SRS by $11\%$ in RMSE.
For Trump, CPS loses to SRS by $86\%$ in RMSE, because $\*\pi$ is built from a very bad predictor $f$ (it predicts the mean approval is $0.65$, but the mean is actually $0.30$).
CPS can always match SRS by choosing uniform $\*\pi$, which PS cannot, but CPS can lose to SRS when the inclusion probabilities are chosen poorly.
\Cref{fig:pew_biden_bias_comparison} shows the widths at each budget, with and without the shift.

\paragraph{SAIPE: Under Distribution Shift, CPS Gains Most Where $f$ Is Most Biased.}
Here the bias of $f$ comes from deploying it on a population other than the one it was trained on.
Following the Census Bureau's SAIPE program, whose estimates are used to allocate Title I federal funds to school districts, we estimate the mean personal income of a target state from a sample designed with a predictor trained on California.
We split the 2019 California ACS data in half, and train an income predictor $f$ on the first half and a residual predictor $\hat{u}(x) \approx |y-f(x)|$ on the second half, and use $\hat{u}$ to set $\*\pi$ (\cref{exp:saipe}).
In all four target states, $f$ over-predicts the mean income: by $29\%$ in Alabama, $23\%$ in Oregon, $14\%$ in Texas, and $8\%$ in Massachusetts (rows \enquote{SAIPE AL} to \enquote{SAIPE TX}).
CPS significantly reduces the RMSE of PS in the two states where the bias is largest, by $3\%$ in Alabama and $4\%$ in Oregon, and the expected width by $4\%$ and $3\%$.
In Massachusetts, where the bias is smallest, CPS and PS do not differ significantly in RMSE or in expected width.
In Texas, the RMSE does not differ significantly, but CPS reduces the expected width by $1\%$, and this reduction is significant. Against SRS, CPS reduces the RMSE by $15\%$ in every state.

\paragraph{ASI Replication: With the Predictors and Policies of ASI, CPS Matches PS.}
We use the datasets, predictors, and inclusion probabilities of \citet{zrnic2024asi}, with SRS as the unweighted baseline (\cref{exp:asi}).
For AlphaFold, we estimate an odds ratio, with $f$ from AlphaFold and $\pi_i \propto \min(f(x_i), 1 - f(x_i))$.
For Census, we estimate the coefficient of age in a linear regression of income on age and sex, with $\pi_i \propto \hat u(x_i) \, \bigl|[(X^\top X)^{-1} X_i]_1\bigr|$, where $\hat u(x_i)$ predicts the residual $|y_i - f(x_i)|$ and $[(X^\top X)^{-1} X_i]_1$ is the influence of unit $i$ on the age coefficient, the first entry of the regression coefficient vector.
For Pew, we estimate the mean approval of Biden and of Trump, with $\pi_i \propto \min(f(x_i), 1 - f(x_i))$ (rows \enquote{ASI Census} to \enquote{ASI AlphaFold}).
On all four datasets, the RMSE of CPS is within $2\%$ of PS, and no difference in RMSE is significant.
In expected width, the one significant difference is on Census, where CPS is $1\%$ narrower than PS.
Our theory attributes these small gaps to a small $v$ for these predictors and policies (\cref{sec:cps:mest}).
Census shows that $v$ is not the bias of $f$ in the mean of $Y$: $f$ under-predicts the mean income by $19\%$, but CPS and PS still differ by at most $2\%$.
Against SRS, CPS reduces the expected width by $11$ to $41\%$ on every dataset.
On AlphaFold, the coverage of CPS is $0.86$, below the nominal $0.90$, and the coverage of PS is similar (\cref{tab:variance_quality}).

\paragraph{The Differences Hold at Small Budgets.}
\Cref{tab:rmse_significance} averages over budgets, and our theory holds as $n \to \infty$, so we also check each budget on its own (\cref{app:per_budget}).
In every row where CPS reduces the RMSE of PS on average, CPS also reduces it at the smallest budget, $n = 100$: by $27\%$ on DGP1, by $31\%$ with $f_{bias}$, by $17\%$ and $9\%$ on Pew with the shifted predictor, and by $1.5\%$ and $3\%$ in Alabama and Oregon.
In the other rows, the RMSE of CPS stays within $4\%$ of PS at every budget.
The point is that, although our theoretical result holds asymptotically, our empirical results show that the CPS advantage holds even at small budgets. 

\subsection{Discussion}

These experiments demonstrate that our theoretical claim, CPS has variance at most that of PS and hence no wider confidence intervals, holds on real datasets. 
Though the theory shows that the gap between CPS and PS is due to $v:=\E_m[w^\star(1-\pis)]$, the average residual of the estimating equation weighted by the probability that a unit is not sampled, we show that the bias of $f$ is an informative heuristic.
Since $w^\star = g(X,Y)-g(X,f(X))$, it is intuitive to translate the bias of $f$ into a statement about the deviation of $v$ from $0$. 
For mean estimation in general, $w^\star= Y-f(X)$ and the heuristic is better when the correlation between the residuals and the probability that a unit is not sampled is low. 
The use of the heuristic is supported in the SAIPE experiment (mean estimation), where the most biased states, AL and OR, are also the two that saw significant RMSE improvements of CPS over PS. 
Even for ASI Census, which estimates a regression coefficient, the heuristic correctly predicts that it would see the biggest reduction among ASI rows in RMSE and width (though only width is significant).
Importantly, PS is never significantly better than CPS in terms of RMSE or expected width, regardless of bias.
SRS, however, can be better when the predictor is very poor, as in \enquote{Pew-Trump, shifted $f$} (see \cref{exp:pew_biased_f} for details).

Our advice to the practitioner is to use CPS with weights determined by $f$ unless $f$ has a very high bias, in which case we recommend SRS or using a different $f_\pi$ to select the weights.
Together, the evidence in \enquote{Synthetic DGP0, $f_{bias}$, DE}, that with uniform weights CPS is the same as SRS, and \enquote{Pew-Trump, shifted $f$}, that very biased $f$ causes SRS to beat CPS, suggests that CPS can do no worse than SRS when it uses uniform weights.
Thus, $\*\pi$ can be selected to be arbitrarily close to uniform, only deviating in certain cases; however, choosing those cases is more art than science. 
That is, a predictor $f_\pi$ directionally correct for the residuals, even if worse than $f$ in metrics that determine usefulness for DE \citep{angelopoulos2023ppi, mani2025ppi}, can still be useful for determining weights. 
For example, $f$ might be a neural network and $f_\pi$ might be a linear regression.

\section{Conclusion}

Poisson Sampling has a random sample size, which is both practically inconvenient and a source of additional variance in the estimator. 
Conditional Poisson Sampling solves both problems by being fixed-size. 
CPS is a valid drop-in replacement for PS on 
general Z-estimation problems considered in prior work.
We establish its asymptotic normality via Hájek's framework rather than the Lindeberg CLT, which fails because CPS samples are dependent, and show that the asymptotic variance of CPS is no worse than that of PS for any policy $\pis$.
Our experiments indicate that this result holds in practice, for a range of estimands and datasets: CPS never performs significantly worse and can be significantly better than PS. 
There is no need to suffer a variable sample size.
More broadly, this work connects the ML evaluation literature to a rich body of survey sampling theory that has studied weighted fixed-size without-replacement sampling for decades. 
We hope it encourages further exchange between the two communities.

\pagebreak
\bibliographystyle{plainnat}
\bibliography{z.bib}

\pagebreak
\appendix

\section{Intro}

\subsection{Estimator-Based Techniques for Variance Reduction}\label{app:estimator-techniques}

A sample of size $n$ leaves $N-n$ units unlabeled. If a cheap label-predictor $f : \1X \to \1Y$ approximates $Y$ on $X$, we can evaluate $f(x_i)$ on all $N$ points for free and reserve the $n$ expensive human labels for correcting $f$'s error. This idea has been independently developed in survey sampling as the generalized difference estimator \citep{cassel1976difference, sarndal1992model}, in causal inference and missing data as the augmented inverse-probability-weighted estimator \citep{robins1994estimation}, and recently in machine learning as prediction-powered inference \citep{angelopoulos2023ppi, angelopoulos2023ppi++}. 

The mechanics start from a population-level decomposition that holds for any $f$:
\al{
    & \mu 
        = \underbrace{\frac{1}{N} \sum_{i=1}^N f(x_i)}_{\text{$\approx 0$ variance, all } N \text{ covariates}}
        + \underbrace{\frac{1}{N} \sum_{i=1}^N \bigparens{y_i - f(x_i)}}_{\text{estimated from } n \text{ labels}}.
}
The first term is free to compute exactly. The second is the population mean of the residuals $y - f(x)$, which we estimate unbiasedly on the sample, because we do not have access to all $N$ values of $y$. Under SRS (Simple Random Sampling), we get
\al{
    & \hat\mu^f_{SRS} 
        := \frac{1}{N}\sum_{i=1}^N f(x_i) 
        + \frac{1}{n}\sum_{i \in \*s}\bigparens{y_i - f(x_i)}.
}
The estimator is unbiased for any $f$, and its variance is governed by the variance of the residuals rather than the variance of $y$. So any $f$ sufficiently correlated with $y$ yields a strict variance reduction over the sample mean \citep{mani2025ppi}. It can be generalized to weighted designs by reweighting the residual sum by $1/\pi_i$.

\subsection{The Other Side of Flipping Coins: Criticisms of PS}\label{app:criticize_PS}

\textbf{Criticism 1: Variance Estimation}

The variance of HT for any design is:
\begin{lemma}[Design Variance of $\hat\mu_{\*\pi}$]\label{prop:expansion-variance}
	The variance of HT for a fixed finite population of $N$ units is
	\al{
		&   \V_{\p}[\hat{\mu}_{\*\pi}^w]
			= 	\E_{\p}[(\hat{\mu}_{\*\pi}^{w} - \mu_N^w)^2]
			= 	\frac{1}{N^2}\sum_{i=1}^{N} \sum_{j = 1}^{N}
				\frac{w_i}{\pi_i} \frac{w_j}{\pi_j} (\pi_{ij} - \pi_i\pi_j)
	}
\end{lemma}

Fixed-size designs can use the following unbiased estimator, but variable-size designs cannot. 

\begin{lemma}[SYG Variance \citep{YatesGrundy1953variance, Sen1953variance}]
    The Sen--Yates--Grundy (SYG) variance estimator is
    \al{
        &   \hat{\V}_{SYG}[\hat\mu] =
            \frac{1}{2N^2} \sum_{i\in\*s}\sum_{j\in\*s}
            \bigparens{\frac{y_i}{\pi_i} - \frac{y_j}{\pi_j}}^2 \frac{\pi_i\pi_j - \pi_{ij}}{\pi_{ij}}
    }
\end{lemma}

Any fixed-size design will exhibit dependence between units, even with uniform probabilities. 
Unless the design clusters points together, units have negative dependence $\pi_{ij}<\pi_i\pi_j$\footnote{An example of clustering: a design that always samples units $i$ and $j$ together has $\pi_{ij} = \pi_i = \pi_j > \pi_i \pi_j$.}. 
SRS exhibits negative dependence because of the fixed-size constraint. This is a problem (in general) because $\pi_{ij}<\pi_i\pi_j$ can cause the Horvitz--Thompson estimate of the variance in \cref{prop:expansion-variance} to be negative. The problem is exacerbated if $y$ can be negative. A solution is to use the SYG estimator, but it is only valid for fixed-size designs. 
It is non-negative when $\pi_{ij}<\pi_i\pi_j$, and so is in some sense better.

\textbf{Criticism 2: Strange Behavior}

On a population with zero variance, the variance-minimizing probability proportional to size (PPS) policy $\pi_i\propto |y_i|$ has positive variance:
\begin{example}[Scale-Dependent Variance]\label{ex:scale_dependent_variance}
    The mean estimator has positive variance when coupled with PS on a population with no variance. It has zero variance under SRS (proof in \cref{app:simple_variance}). Let $y_i = r\in\3R\setminus\{0\} \gap \forall i$. The variance-minimizing $\*\pi$ is uniform: $\pi_i = n/N \gap\forall i$. The variance of PS from \cref{prop:expansion-variance} is
    \al{
            &   \frac{1}{N^2}\sum_{i=1}^N \bigparens{\frac{y_i^2}{\pi_i} - y_i^2}
                = \frac{r^2}{N^2}\bigparens{\sum_{i=1}^N\frac{1}{\pi_i} - N}
                = 0 \iff \sum_{i=1}^N \frac{1}{\pi_i} = N
            \\& \text{Take $a_i=\sqrt{\pi_i}$, $b_i=1/\sqrt{\pi_i}$ and apply Cauchy--Schwarz: $\sum_{i=1}^N \frac{1}{\pi_i} \geq \frac{N^2}{\sum_i \pi_i} = \frac{N^2}{n} > N \impliedby n < N$}
    }
\end{example}

Intuitively, $\hat{\mu}_{\*\pi}$ gives a different mean estimate for different realizations of the random sample size $\tilde{n}_{\*\pi}$. It is unbiased in expectation, but only because high realizations balance low realizations. This intuition extends beyond constant populations, though it does not imply that one should always use SRS over PS. It does indicate that fixed-size designs have a fundamental advantage over PS.

\subsection{Variance of Simple Random Sampling}\label{app:simple_variance}

Simple Random Sampling (SRS), also known as uniform fixed-size without-replacement sampling, assigns $\pi_i = \frac{n}{N}$ for all $i$. Let $\hat{T}\equiv\hat\mu N$ be the total estimator.
\al{
    &   \V[\hat{T}] 
        = \sum_i\sum_j \frac{y_iy_j}{\pi_i\pi_j} (\pi_{ij} - \pi_i\pi_j)
    \\& = \sum_i \frac{y_i^2}{\pi_i^2}(\pi_i - \pi_i^2) 
        + \sum_i\sum_{j\neq i} \frac{y_iy_j}{\pi_i\pi_j} (\pi_{ij} - \pi_i\pi_j)
    \\& = \sum_i \frac{y_i^2}{\frac{n^2}{N^2}}\bigparens{\frac{n}{N} - \frac{n^2}{N^2}} 
        + \sum_i\sum_{j\neq i} \frac{y_iy_j}{\frac{n^2}{N^2}} \bigparens{\frac{(n-1)n}{(N-1)N} - \frac{n^2}{N^2}}
        &&\tag{$\pi_{ij} = \pi_{i\mid j}\pi_j = \frac{(n-1)n}{(N-1)N}$}
    \\& = \sum_i y_i^2\bigparens{\frac{N}{n} - 1} 
        + \sum_i\sum_{j\neq i} y_iy_j \bigparens{\frac{N(n-1)}{n(N-1)} - 1}
    \\& = \frac{N-n}{n}\sum_i y_i^2 
        - \frac{N-n}{n(N-1)} \sum_i\sum_{j\neq i} y_iy_j
    \\& = \frac{N-n}{n} \bigparens{
            \sum_i y_i^2 - \frac{1}{N-1} (T^2 - \sum_i y_i^2)
            }
            &&\tag{$T^2 := (\sum_i y_i)^2 = \sum_i\sum_j y_iy_j$} 
    \\& = \frac{N(N-n)}{n(N-1)}\sum_i y_i^2 - \frac{N-n}{n(N-1)}T^2 
    \\& = \frac{N-n}{n(N-1)}(N\sum_i y_i^2 - T^2)
}

Consider the constant population $\*Y=\{r\commadots r\}$, $r\in\3R$. The variance of $\hat{T}$ is then $0$:
\al{
    &   \V[\hat{T}] = \frac{N-n}{n(N-1)} (N(r^2N) - r^2N^2) = 0
}

By contrast, Poisson Sampling has non-zero variance on a constant population (see \cref{ex:scale_dependent_variance}).

\section{Survey}\label{app:survey}

\subsection{CPS Implementation}\label{app:cps-implementation}

\citet{Chen1994cps} show how to map $\*\pi$ to $\*p$ to implement CPS with rejective sampling on $\p_{\textup{ps}(\*p)}$.
\begin{align*}
    &   \text{let } \pi_k(\*p, n) :=
        n \dfrac{
            \frac{p_k}{1-p_k}
            (1-\pi_k(\*p, n-1))
        }{
            \sum_{l\in\1U}
                \frac{p_l}{1-p_l}
                (1-\pi_l(\*p, n-1))
        }
        \text{  with  }
        \*\pi(\*p, 0) = 0
    \\& \text{initialize } \hspace{1em} \*p^0 = \*\pi
    \\&
        \text{iterate until convergence }
        \*p^{i} = \*p^{i-1} + \*\pi - \*\pi(\*p^{i-1}, n)
\end{align*}
Then \citet{Deville2000cps} show how to get $\pi_{ij}$ recursively. 
\begin{align*}
    & \text{let } \pi_{ij}(0) = \pi_{ij}(1) = 0
    \\& \pi_{ij}(n) = \frac{
        n(n-1) \frac{p_ip_j}{(1-p_i)(1-p_j)}
        [1 -\pi_i(n-2) -\pi_j(n-2) + \pi_{ij}(n-2)]
        }{
        \sum_{k\in\1U} \sum_{l\neq k}
        \frac{p_kp_l}{(1-p_k)(1-p_l)}
        [1 - \pi_k(n-2) - \pi_l(n-2) + \pi_{kl}(n-2)]
        }
\end{align*}

The implementation we follow is based on the function \texttt{UPmaxentropy} in \citet{tille2016package}\footnote{CPS is sometimes referred to as maximum entropy sampling.}. 
Both their codebase and our codebase use the sequential procedure of \citet[Sec.~5.6.9, p.~91]{tille2006sampling}, which 
implements CPS with dynamic programming instead of rejective sampling to achieve 
$O(nN)$ space and time. Here is how and why it works. 

The first observation to make is that any sampling design, CPS or otherwise, can be implemented by a sequential procedure.
Implementing a sampling design means drawing a sample $\*s\sim P(\*s := (s_1 \commadots s_N))$.
Write $P(\*s) = \prod_{i=1}^N P(s_i \mid s_1 \commadots s_{i-1})$ and sample from each conditional distribution $s_i \sim P(s_i \mid s_1 \commadots s_{i-1})$ to implement the design, and thus $\E_d[P(s_i = 1 \mid s_1 \commadots s_{i-1})] = \pi_i$ for all $i$. 
Intuitively, the fixed-size constraint zeros out later conditional probabilities, but this is balanced out by the cases where earlier units are not sampled. 

For general designs, the sequential procedure requires computing a conditional probability for each unit, based on all possible outcomes of previously sampled units. 
So unit $i$ requires a different conditional probability for each realization of $s_1 \commadots s_{i-1}$. 
The conditional probability for CPS is more tractable, because the sequential procedure only requires knowing the number of heads among $s_1 \commadots s_{i-1}$: it depends only on how many units were sampled, not which ones. 
That is, $\p_{\textup{cps}}(s_i \commadots s_N \mid s_1 \commadots s_{i-1})$ for size $n$ equals $\p_{\textup{cps}}(s_i \commadots s_N)$ for size $n - \sum_{k<i} s_k$.
Thus, $P(s_i \mid s_1 \commadots s_{i-1}) = P(s_i \mid \sum_{k<i} s_k)$, so the conditional probability only depends on the number of heads sampled thus far, not which units received heads. 
Consequently, to store the full conditional probabilities needed to sample unit $i$, we only need to store $\min(i,n+1)$ histories instead of $2^{i-1}$ histories.

To recap what we have shown:
First, we can sample any design in $O(N)$ sequentially if we have the conditional probabilities.
Second, CPS conditional probabilities at unit $i$ depend only on the number of heads in the history up to $i$ in the sequential procedure. 
So a natural question is why we cannot compute the conditional probabilities online; why do we need to precompute a table of size $O(nN)$ containing conditional probabilities?

The answer is that the coin probabilities are computed using recursion from the end. 
First, observe that $k$ heads in the past of unit $i$ implies $n-k$ heads in the future.
Let $Z(i,m) := \p_{\textup{ps}(\*p)}(\sum_{j\geq i} s_j = m)$ be the probability that coins in the Poisson Sampling design underlying CPS have exactly $m$ heads between unit $i$ and unit $N$. 
Second, derive the probability of heads at unit $i$:
\al{
    &   \p_{\textup{cps}(\*\pi)}(s_i=1 \mid s_1 \commadots s_{i-1})
        = \p_{\textup{cps}(\*\pi)}(s_i=1 \mid |s_{j<i}|=k)
    \\& = \p_{\textup{cps}(\*\pi)}(s_i=1 \mid |s_{j<i}|=k, |s_{j\geq i}|=n-k)
        &&\tag{they are the same event}
    \\& = \p_{\textup{ps}(\*p)}(s_i=1 \mid |s_{j<i}|=k, |s_{j\geq i}|=n-k)
        &&\tag{CPS is PS conditioned on sample size $n$}
    \\& = \p_{\textup{ps}(\*p)}(s_i=1 \mid |s_{j\geq i}|=n-k)
        &&\tag{PS is independent of the past, $j\ge i$ contains $s_i$}
    \\& = \frac{ 
            \p_{\textup{ps}(\*p)}(s_i=1, |s_{j\geq i}|=n-k)
        }{
            \p_{\textup{ps}(\*p)}(|s_{j\ge i}|=n-k)
        }
    \\& = \frac{ 
            p_i Z(i+1, n-k-1)
        }{
            Z(i, n-k)
        }
        &&\tag{same event as $s_i=1$ and $|s_{j\ge i}|=n-k$}
}
Now write $Z(i,m) = p_i Z(i+1,m-1) + (1-p_i) Z(i+1,m)$ to relate it to the same probability at unit $i+1$, with base case $Z(N+1, 0) = 1$ and $Z(N+1,m)=0$ for $m \neq 0$. 
This tells us that knowing the $Z$ values for $i+1$ gives us $Z$ values for $i$, hence the recursion runs backward, starting from unit $N$. 

Thus we can draw a CPS sample as follows. 
Build the table of coins in $O(nN)$ time and store it in $O(nN)$ space using the recursion from $N$, justified by the above derivation. 
Each row in the table is a number of heads in the history and each column is the unit that we are flipping at. 
Then move through the table and increment the row every time we get heads, to 
draw a sample in $O(N)$.

\paragraph{Runtimes.}
\Cref{tab:runtimes} times every experiment at $n = 1000$. The main point is that for these datasets, the difference is a matter of seconds at worst. The CPS setup (computing $\*p$ and filling the table) takes 0.4 to 17 seconds and grows with $N$, as the $O(nN)$ cost predicts. Once the table is built, a CPS draw takes 0.1 to 11 milliseconds. The overhead matters when both $n$ and $N$ are large, because the table has $nN$ entries: for example, $n = 10^4$ and $N = 10^7$ would need $10^{11}$ entries (about 800 GB at double precision), whereas PS needs no table.

\begin{table}[t]
\centering
\begin{tabular}{lrrrrr}
\toprule
Dataset & $N$ & $n$ & CPS build (s) & PS draw (ms) & CPS draw (ms) \\
\midrule
SAIPE AL & 26,607 & 1000 & 2.42 & 1.05 & 1.89 \\
SAIPE OR & 25,186 & 1000 & 1.95 & 0.93 & 0.60 \\
SAIPE MA & 45,372 & 1000 & 3.46 & 1.61 & 1.69 \\
SAIPE TX & 157,386 & 1000 & 8.68 & 5.54 & 9.40 \\
Census & 190,046 & 1000 & 16.82 & 7.22 & 10.95 \\
Pew-Biden & 6,189 & 1000 & 0.38 & 0.29 & 0.11 \\
Pew-Trump & 6,207 & 1000 & 0.39 & 0.30 & 0.11 \\
AlphaFold (2 groups) & 10,802 & 1000 & 0.36 & 0.49 & 0.17 \\
\bottomrule
\end{tabular}
\caption{Runtime of PS vs.\ CPS per experiment. $\*\pi$ is a shared input and is not timed. PS has no build step; the CPS build time is the $\*\pi\to \*p$ solve plus the table fill, averaged over 5 cold runs (cache wiped each run). Times are wall-clock on a single machine.}
\label{tab:runtimes}
\end{table}

\subsection{Why Fixed-Size, WoR, Weighted Sampling Is Hard}\label{app:all3_is_hard}

A sampling algorithm is an iterative procedure that includes some $i$ in $\*s$ at each iteration $1\leq t \leq n$. It tracks a state variable $S$ containing (1) weights $\*w$, (2) iteration $t$, and (3) included points $\*s_t$. A \enquote{correct} procedure must specify a conditional distribution for each possible state: $P(\text{select $i$ at iteration $t$} \mid \*w, \*s_t, t)$ that satisfies the desired inclusion probabilities: $\pi_i := \sum_{\*s:i\in\*s} \p(\*s)$. Naively, it must specify $O(2^N)$ conditional probabilities, because the probabilities are different for different sample paths. Each two-of-three algorithm, which achieves two of the three properties (weighted, fixed-size, and without replacement), simplifies the state by ignoring some aspect of it. SRS ignores (1) and is therefore unweighted. PS ignores (2) and is therefore variable-size. Multinomial sampling (MS) ignores (3) and therefore samples with replacement. An algorithm that achieves all three cannot ignore any aspect of the state. The challenge is to find a tractable all-three algorithm. 

Beyond aesthetics, the motivation for an all-three algorithm is that each of the three improves the variance under $\pis$. 
WoR reduces variance because sampling the same point twice can be non-optimal. 
Imagine a population with three typical points and one outlier: we want to sample the outlier once, but not twice. 
Weighting reduces variance when certain combinations of points estimate the mean better than others. 
The weights make it more likely that we include those combinations.
Fixed-size reduces variance because $\*\pi$ is chosen w.r.t.\ $n$, so deviations from $n$ sampled points can reduce the efficacy of the policy. 
However, showing that this holds for CPS is the least obvious of the three, because PPS is non-optimal for CPS~\citep{Qualite2008comparison}; doing so is the main contribution of our paper.

\section{Experiments}\label{app:experiments}

\subsection{Per-Budget Results and Experiment Details}\label{app:per_budget}

\Cref{tab:rmse_significance} averages over the budgets $n \in \{100, 300, 500, 700, 1000\}$.
This section gives the setup of each experiment in more detail and its results at each budget.

\subsubsection{PS Variance Depends on Population Mean}\label{exp:mean0-Y}

\begin{figure}[t]
\centering
\providecolor{srscolour}{HTML}{2CA02C}
\providecolor{pscolour}{HTML}{1A9EFF}
\providecolor{cpscolour}{HTML}{FF1F1F}
\begin{subfigure}[b]{0.48\textwidth}
\centering
\begin{tikzpicture}
		\begin{axis}[
			width=\linewidth, height=0.8\linewidth,
			xlabel={budget $n$}, ylabel={$\E[\text{width}]$},
			xtick={100,300,500,700,1000},
			grid=major, grid style={gray!25},
			tick label style={font=\footnotesize}, label style={font=\small},
			/pgf/number format/1000 sep={},
			scaled y ticks=false, yticklabel style={/pgf/number format/fixed, /pgf/number format/precision=3},
			legend style={font=\footnotesize, draw=none, at={(0.5,1.03)}, anchor=south},
			legend columns=3, legend cell align=left,
		]
			\addlegendimage{srscolour, line width=1pt}\addlegendentry{SRS}
			\addlegendimage{pscolour, line width=1pt}\addlegendentry{PS}
			\addlegendimage{cpscolour, line width=2.4pt, opacity=0.5}\addlegendentry{CPS}
			\addlegendimage{black, dotted, line width=1pt}\addlegendentry{$\E[Y]=0$}
			\addlegendimage{black, line width=1pt}\addlegendentry{$\E[Y]=2$}
			\addplot[srscolour, line width=1pt, dotted, forget plot] coordinates {(100,0.72462) (300,0.41486) (500,0.31771) (700,0.26567) (1000,0.21887)};
			\addplot[srscolour, line width=1pt, solid, forget plot] coordinates {(100,0.72462) (300,0.41486) (500,0.31771) (700,0.26567) (1000,0.21887)};
			\addplot[pscolour, line width=1pt, dotted, forget plot] coordinates {(100,0.72369) (300,0.41461) (500,0.31776) (700,0.26577) (1000,0.21876)};
			\addplot[pscolour, line width=1pt, solid, forget plot] coordinates {(100,0.97868) (300,0.56078) (500,0.42971) (700,0.35929) (1000,0.29568)};
			\addplot[cpscolour, line width=2.4pt, opacity=0.5, dotted, forget plot] coordinates {(100,0.72294) (300,0.41413) (500,0.31752) (700,0.26558) (1000,0.21865)};
			\addplot[cpscolour, line width=2.4pt, opacity=0.5, solid, forget plot] coordinates {(100,0.72294) (300,0.41413) (500,0.31752) (700,0.26558) (1000,0.21865)};
		\end{axis}
\end{tikzpicture}
\caption{Variance comparison of CPS, PS, and SRS using the HT estimator on two populations with different means. PS equals CPS and SRS on the mean-$0$ population, but is worse when the population has mean $2$. The solid line below contains the other 5 curves, and the single blue line is PS on DGP1.}
\label{fig:synthetic1}
\end{subfigure}
\hfill
\begin{subfigure}[b]{0.48\textwidth}
\centering
\begin{tikzpicture}
		\begin{axis}[
			width=\linewidth, height=0.8\linewidth,
			xlabel={budget $n$}, ylabel={$\E[\text{width}]$},
			xtick={100,300,500,700,1000},
			grid=major, grid style={gray!25},
			tick label style={font=\footnotesize}, label style={font=\small},
			/pgf/number format/1000 sep={},
			scaled y ticks=false, yticklabel style={/pgf/number format/fixed, /pgf/number format/precision=3},
			legend style={font=\footnotesize, draw=none, at={(0.5,1.03)}, anchor=south},
			legend columns=3, legend cell align=left,
		]
			\addlegendimage{srscolour, line width=1pt}\addlegendentry{SRS}
			\addlegendimage{pscolour, line width=1pt}\addlegendentry{PS}
			\addlegendimage{cpscolour, line width=2.4pt, opacity=0.5}\addlegendentry{CPS}
			\addlegendimage{black, dotted, line width=1pt}\addlegendentry{$f_{lin}$}
			\addlegendimage{black, line width=1pt}\addlegendentry{$f_{bias}$}
			\addplot[srscolour, line width=1pt, dotted, forget plot] coordinates {(100,0.32415) (300,0.1859) (500,0.14245) (700,0.11912) (1000,0.098091)};
			\addplot[srscolour, line width=1pt, solid, forget plot] coordinates {(100,0.32415) (300,0.1859) (500,0.14245) (700,0.11912) (1000,0.098091)};
			\addplot[pscolour, line width=1pt, dotted, forget plot] coordinates {(100,0.32459) (300,0.18582) (500,0.14246) (700,0.11917) (1000,0.098094)};
			\addplot[pscolour, line width=1pt, solid, forget plot] coordinates {(100,0.45968) (300,0.26322) (500,0.20181) (700,0.16879) (1000,0.13894)};
			\addplot[cpscolour, line width=2.4pt, opacity=0.5, dotted, forget plot] coordinates {(100,0.32475) (300,0.18584) (500,0.14254) (700,0.11915) (1000,0.098076)};
			\addplot[cpscolour, line width=2.4pt, opacity=0.5, solid, forget plot] coordinates {(100,0.32475) (300,0.18584) (500,0.14254) (700,0.11915) (1000,0.098076)};
		\end{axis}
\end{tikzpicture}
\caption{Variance comparison across the same population but with the difference estimator for two different $f$s. PS equals CPS and SRS when $f$ is unbiased, but is worse when $f$ is biased. The solid line below contains the other 5 curves, and the single blue line is PS for $f_{bias}$.}
\label{fig:synthetic2}
\end{subfigure}
\caption{PS variance is inflated when the residuals have non-zero mean---either from a non-zero population mean (left) or a biased predictor (right).}
\label{fig:synthetic}
\end{figure}
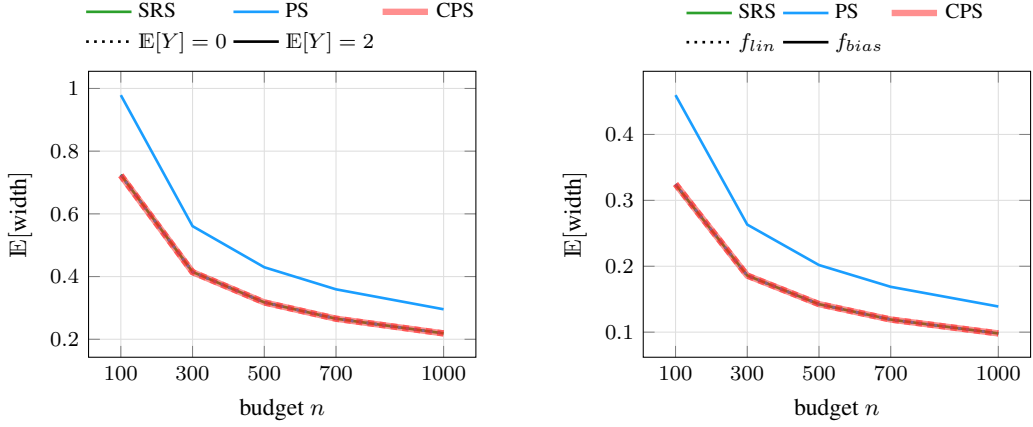

We compare SRS, CPS, and PS with $\pi_i = n/N$ for mean estimation first using 
\Cref{def:expansion-estimator} and then using \Cref{def:augmented-expansion-estimator}.
The left panel of \Cref{fig:synthetic} considers two data-generating processes, identical except that the first has $\E[Y] = 0$ and the second has $\E[Y] = 2$. The right panel uses only DGP0, with two predictors: one unbiased $f_{lin}(X) = 2 \cdot X$ and one biased $f_{bias}(X) = f_{lin}(X) + 1$.
\begin{align*}
    &   X \sim \1N(0,1) \quad Y \sim 2 \cdot X + \1N(0,1)
        &&\tag{DGP0}
    \\& X \sim \1N(1,1) \quad Y \sim 2 \cdot X + \1N(0,1)
        &&\tag{DGP1}
\end{align*}
In \Cref{fig:synthetic1}, we observe that SRS and CPS have identical variance under both data-generating processes, and that their variance is identical to that of PS under DGP0. 
However, the variance of PS is higher on DGP1, because the outcome $Y$ has non-zero mean. 
In particular, PS on DGP1 is the only algorithm that has higher variance. 
In \Cref{fig:synthetic2}, we see the same pattern for mean estimation with \Cref{def:augmented-expansion-estimator} when comparing a biased predictor to an unbiased predictor. 
Intuitively, the population labels under DE are $Y-f(X)$ instead of $Y$.
Again, we observe that only the variance of PS is shifted up.

\subsubsection{Pew Survey with a Biased Predictor}\label{exp:pew_biased_f}

\begin{figure}[t]
\centering
\providecolor{srscolour}{HTML}{2CA02C}
\providecolor{pscolour}{HTML}{1A9EFF}
\providecolor{cpscolour}{HTML}{FF1F1F}
\begin{subfigure}[b]{0.48\textwidth}
\centering
\begin{tikzpicture}
		\begin{axis}[
			width=\linewidth, height=0.8\linewidth,
			xlabel={budget $n$}, ylabel={$\E[\text{width}]$},
			xtick={100,300,500,700,1000},
			grid=major, grid style={gray!25},
			tick label style={font=\footnotesize}, label style={font=\small},
			/pgf/number format/1000 sep={},
			scaled y ticks=false, yticklabel style={/pgf/number format/fixed, /pgf/number format/precision=3},
			legend style={font=\footnotesize, draw=none, at={(0.5,1.03)}, anchor=south},
			legend columns=3, legend cell align=left,
		]
			\addlegendimage{srscolour, line width=1pt}\addlegendentry{SRS}
			\addlegendimage{pscolour, line width=1pt}\addlegendentry{PS}
			\addlegendimage{cpscolour, line width=2.4pt, opacity=0.5}\addlegendentry{CPS}
			\addlegendimage{black, line width=1pt}\addlegendentry{$f$}
			\addlegendimage{black, dotted, line width=1pt}\addlegendentry{$f_{bias}$}
			\addplot[srscolour, line width=1pt, solid, forget plot] coordinates {(100,0.10421) (300,0.059355) (500,0.045199) (700,0.037539) (1000,0.030534)};
			\addplot[srscolour, line width=1pt, dotted, forget plot] coordinates {(100,0.13416) (300,0.076262) (500,0.05807) (700,0.048195) (1000,0.039209)};
			\addplot[pscolour, line width=1pt, solid, forget plot] coordinates {(100,0.093827) (300,0.052747) (500,0.039686) (700,0.032747) (1000,0.026556)};
			\addplot[pscolour, line width=1pt, dotted, forget plot] coordinates {(100,0.1443) (300,0.079218) (500,0.060251) (700,0.050275) (1000,0.041242)};
			\addplot[cpscolour, line width=2.4pt, opacity=0.5, solid, forget plot] coordinates {(100,0.093203) (300,0.052835) (500,0.039715) (700,0.032799) (1000,0.026565)};
			\addplot[cpscolour, line width=2.4pt, opacity=0.5, dotted, forget plot] coordinates {(100,0.11952) (300,0.065427) (500,0.050036) (700,0.042009) (1000,0.034707)};
		\end{axis}
\end{tikzpicture}
\caption{Biden}
\label{fig:pew_bias_biden}
\end{subfigure}
\hfill
\begin{subfigure}[b]{0.48\textwidth}
\centering
\begin{tikzpicture}
		\begin{axis}[
			width=\linewidth, height=0.8\linewidth,
			xlabel={budget $n$}, ylabel={$\E[\text{width}]$},
			xtick={100,300,500,700,1000},
			grid=major, grid style={gray!25},
			tick label style={font=\footnotesize}, label style={font=\small},
			/pgf/number format/1000 sep={},
			scaled y ticks=false, yticklabel style={/pgf/number format/fixed, /pgf/number format/precision=3},
			legend style={font=\footnotesize, draw=none, at={(0.5,1.03)}, anchor=south},
			legend columns=3, legend cell align=left,
		]
			\addlegendimage{srscolour, line width=1pt}\addlegendentry{SRS}
			\addlegendimage{pscolour, line width=1pt}\addlegendentry{PS}
			\addlegendimage{cpscolour, line width=2.4pt, opacity=0.5}\addlegendentry{CPS}
			\addlegendimage{black, line width=1pt}\addlegendentry{$f$}
			\addlegendimage{black, dotted, line width=1pt}\addlegendentry{$f_{bias}$}
			\addplot[srscolour, line width=1pt, solid, forget plot] coordinates {(100,0.10082) (300,0.057515) (500,0.04381) (700,0.036392) (1000,0.029593)};
			\addplot[srscolour, line width=1pt, dotted, forget plot] coordinates {(100,0.10947) (300,0.062315) (500,0.047485) (700,0.039425) (1000,0.032066)};
			\addplot[pscolour, line width=1pt, solid, forget plot] coordinates {(100,0.090506) (300,0.050715) (500,0.038174) (700,0.031493) (1000,0.025495)};
			\addplot[pscolour, line width=1pt, dotted, forget plot] coordinates {(100,0.25875) (300,0.12903) (500,0.090511) (700,0.07141) (1000,0.055391)};
			\addplot[cpscolour, line width=2.4pt, opacity=0.5, solid, forget plot] coordinates {(100,0.090136) (300,0.050764) (500,0.038209) (700,0.031516) (1000,0.025514)};
			\addplot[cpscolour, line width=2.4pt, opacity=0.5, dotted, forget plot] coordinates {(100,0.23144) (300,0.11163) (500,0.076025) (700,0.058648) (1000,0.044367)};
		\end{axis}
\end{tikzpicture}
\caption{Trump}
\label{fig:pew_bias_trump}
\end{subfigure}
\caption{Expected width of confidence intervals (CIs) as a function of sample size budget for SRS, PS, and CPS on the Pew dataset.  Solid lines show the original predictor $f$, and dotted lines show the logit-shifted $f_{bias}$, with $b = 3$. For the original $f$, CPS and PS have comparable widths, lower than SRS.  For the biased $f$ on Biden, CPS has narrower CIs than SRS, which in turn is narrower than PS.  For the biased $f$ on Trump, CPS still dominates PS, but both are worse than SRS.}
\label{fig:pew_biden_bias_comparison}
\end{figure}
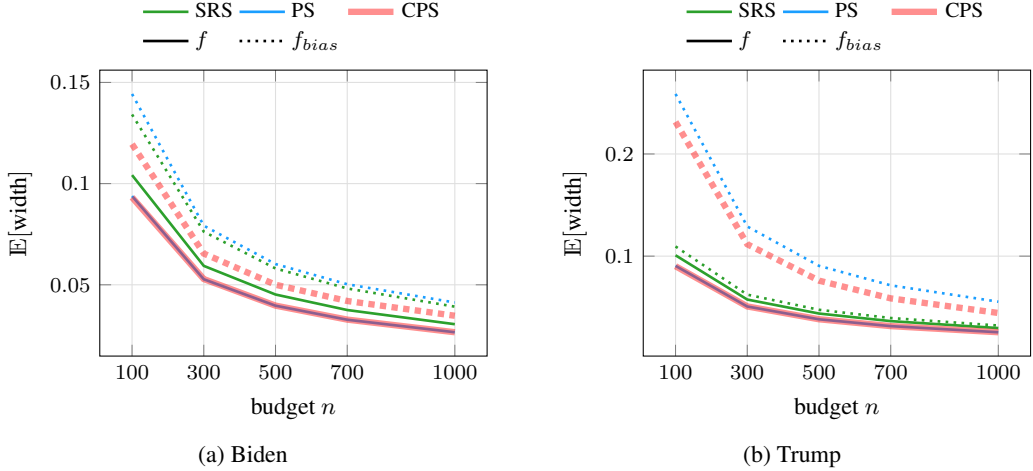

To demonstrate that PS is sensitive to bias in predictors on realistic datasets, we compare SRS, CPS, and PS using DE on the Pew dataset with two label predictors: the original predictor $f$ of~\citet{zrnic2024asi}, and $f_{bias}$, which shifts the pre-sigmoid logits of $f$ by a constant. In each case, the same predictor sets the weights, $\pi_i \propto \min(f(x_i), 1-f(x_i))$ or $\pi_i \propto \min(f_{bias}(x_i), 1-f_{bias}(x_i))$.
\begin{align*}
    &   f_{bias}(X) = \sigma(\mathrm{logit}(f(X)) + b)
\end{align*}
Notably, the constant bias in $f_{bias}$ only impacts the calibration of $f_{bias}$, and does not otherwise impact metrics that measure discrimination like AUROC.  Such a bias may be realistic in scenarios where a predictor is transferred from one population to another, or becomes poorly calibrated over time even when retaining discriminative ability. We study the first scenario directly in~\cref{exp:saipe}. In \Cref{fig:pew_biden_bias_comparison} we see that when we consider the original (unbiased) $f$, CPS and PS are equal to each other and better than SRS, but for biased $f$, CPS beats PS and is sometimes better than SRS.

The practical takeaway is twofold. When $f$ is a good predictor, CPS may perform similarly to PS, and both may tend to outperform SRS. However, when the predictor $f$ is poorly calibrated, leading to bias, CPS has a stronger advantage over PS and sometimes beats SRS. That said, the setting of a poorly calibrated predictor is also a setting where both weighted-sampling methods can underperform relative to SRS. In theory, the designs implementable by SRS are also implementable by CPS, simply by choosing uniform weights, which does not extend to PS. In practice, however, CPS can lose to SRS when the inclusion probabilities are chosen poorly.

\subsubsection{Distribution Shift: Income Estimation Across States}\label{exp:saipe}

We consider a setting where bias in $f$ occurs because it is deployed on a population with different statistics than the one on which it was trained. 
We are inspired by the Census Bureau's Small Area Income and Poverty Estimates (SAIPE) program, which uses income estimates to allocate Title I federal funding to school districts. We simulate a statistician at the Bureau designing a sample to estimate mean personal income in a target state using labeled data from California. We use the 2019 American Community Survey (ACS) data for employed individuals, the same source as the Census experiment in \citet{zrnic2024asi} and \Cref{exp:asi}.

We split the California sample in half and train two regressors.
The first, $f$, uses XGBoost to predict personal income from $17$ demographic and occupational covariates, but excludes geography since it will be deployed elsewhere. 
The second regressor, $\hat u(x)$, predicts the absolute residual $|y - f(x)|$ from $x$, and determines the weights according to the weight-mixing prescription of \citet{zrnic2024asi}: $\pi_i = \tfrac{1}{2}\, n \frac{\hat u(x_i)}{\sum_{j} \hat u(x_j)} + \tfrac{1}{2}\, \frac{n}{N}$.
Then we deploy on four target states: Alabama (AL), Oregon (OR), Massachusetts (MA), and Texas (TX) using $f$ and $\*\pi$ as defined above. 
In every state, $f$ over-predicts the mean income: by 28.6\% in AL, 23.2\% in OR, 14.0\% in TX, and 8.0\% in MA.

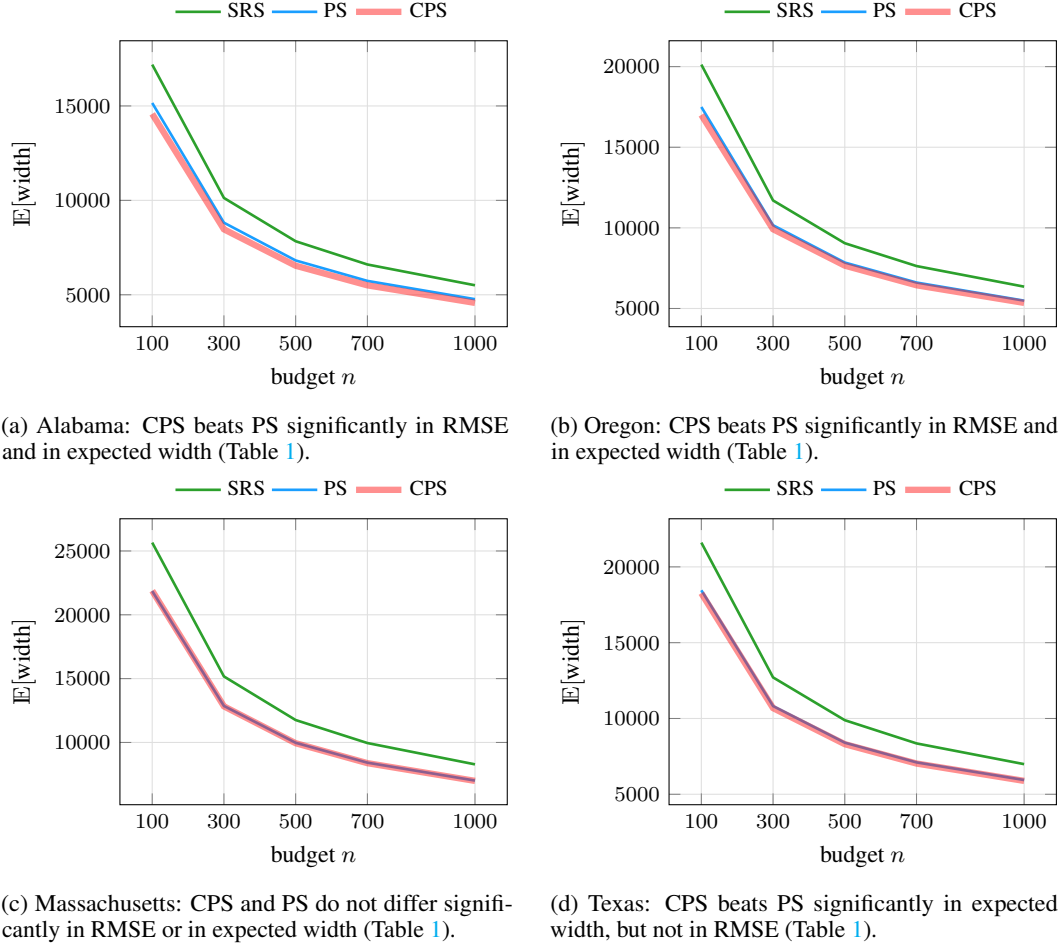
\begin{figure}[t]
\centering
\providecolor{srscolour}{HTML}{2CA02C}
\providecolor{pscolour}{HTML}{1A9EFF}
\providecolor{cpscolour}{HTML}{FF1F1F}
\begin{subfigure}[b]{0.48\textwidth}
\centering
\begin{tikzpicture}
		\begin{axis}[
			width=\linewidth, height=0.8\linewidth,
			xlabel={budget $n$}, ylabel={$\E[\text{width}]$},
			xtick={100,300,500,700,1000},
			grid=major, grid style={gray!25},
			tick label style={font=\footnotesize}, label style={font=\small},
			/pgf/number format/1000 sep={},
			scaled y ticks=false, yticklabel style={/pgf/number format/fixed, /pgf/number format/precision=3},
			legend style={font=\footnotesize, draw=none, at={(0.5,1.03)}, anchor=south},
			legend columns=3, legend cell align=left,
		]
			\addlegendimage{srscolour, line width=1pt}\addlegendentry{SRS}
			\addlegendimage{pscolour, line width=1pt}\addlegendentry{PS}
			\addlegendimage{cpscolour, line width=2.4pt, opacity=0.5}\addlegendentry{CPS}
			\addplot[srscolour, line width=1pt, solid, forget plot] coordinates {(100,17190) (300,10128) (500,7834.8) (700,6604.2) (1000,5503.7)};
			\addplot[pscolour, line width=1pt, solid, forget plot] coordinates {(100,15158) (300,8816.9) (500,6817.4) (700,5735.7) (1000,4767)};
			\addplot[cpscolour, line width=2.4pt, opacity=0.5, solid, forget plot] coordinates {(100,14581) (300,8467.2) (500,6538.7) (700,5500.8) (1000,4571.3)};
		\end{axis}
\end{tikzpicture}
\caption{Alabama: CPS beats PS significantly in RMSE and in expected width (\cref{tab:rmse_significance}).}
\label{fig:saipe_width_AL}
\end{subfigure}
\hfill
\begin{subfigure}[b]{0.48\textwidth}
\centering
\begin{tikzpicture}
		\begin{axis}[
			width=\linewidth, height=0.8\linewidth,
			xlabel={budget $n$}, ylabel={$\E[\text{width}]$},
			xtick={100,300,500,700,1000},
			grid=major, grid style={gray!25},
			tick label style={font=\footnotesize}, label style={font=\small},
			/pgf/number format/1000 sep={},
			scaled y ticks=false, yticklabel style={/pgf/number format/fixed, /pgf/number format/precision=3},
			legend style={font=\footnotesize, draw=none, at={(0.5,1.03)}, anchor=south},
			legend columns=3, legend cell align=left,
		]
			\addlegendimage{srscolour, line width=1pt}\addlegendentry{SRS}
			\addlegendimage{pscolour, line width=1pt}\addlegendentry{PS}
			\addlegendimage{cpscolour, line width=2.4pt, opacity=0.5}\addlegendentry{CPS}
			\addplot[srscolour, line width=1pt, solid, forget plot] coordinates {(100,20127) (300,11700) (500,9048) (700,7631.2) (1000,6353.2)};
			\addplot[pscolour, line width=1pt, solid, forget plot] coordinates {(100,17498) (300,10163) (500,7848.9) (700,6609.2) (1000,5485.9)};
			\addplot[cpscolour, line width=2.4pt, opacity=0.5, solid, forget plot] coordinates {(100,17006) (300,9915.6) (500,7657) (700,6440.2) (1000,5347.3)};
		\end{axis}
\end{tikzpicture}
\caption{Oregon: CPS beats PS significantly in RMSE and in expected width (\cref{tab:rmse_significance}).}
\label{fig:saipe_width_OR}
\end{subfigure}
\begin{subfigure}[b]{0.48\textwidth}
\centering
\begin{tikzpicture}
		\begin{axis}[
			width=\linewidth, height=0.8\linewidth,
			xlabel={budget $n$}, ylabel={$\E[\text{width}]$},
			xtick={100,300,500,700,1000},
			grid=major, grid style={gray!25},
			tick label style={font=\footnotesize}, label style={font=\small},
			/pgf/number format/1000 sep={},
			scaled y ticks=false, yticklabel style={/pgf/number format/fixed, /pgf/number format/precision=3},
			legend style={font=\footnotesize, draw=none, at={(0.5,1.03)}, anchor=south},
			legend columns=3, legend cell align=left,
		]
			\addlegendimage{srscolour, line width=1pt}\addlegendentry{SRS}
			\addlegendimage{pscolour, line width=1pt}\addlegendentry{PS}
			\addlegendimage{cpscolour, line width=2.4pt, opacity=0.5}\addlegendentry{CPS}
			\addplot[srscolour, line width=1pt, solid, forget plot] coordinates {(100,25657) (300,15171) (500,11755) (700,9954) (1000,8283.9)};
			\addplot[pscolour, line width=1pt, solid, forget plot] coordinates {(100,21859) (300,12860) (500,9972.7) (700,8415.9) (1000,7014.2)};
			\addplot[cpscolour, line width=2.4pt, opacity=0.5, solid, forget plot] coordinates {(100,21886) (300,12829) (500,9939.9) (700,8378.3) (1000,6989)};
		\end{axis}
\end{tikzpicture}
\caption{Massachusetts: CPS and PS do not differ significantly in RMSE or in expected width (\cref{tab:rmse_significance}).}
\label{fig:saipe_width_MA}
\end{subfigure}
\hfill
\begin{subfigure}[b]{0.48\textwidth}
\centering
\begin{tikzpicture}
		\begin{axis}[
			width=\linewidth, height=0.8\linewidth,
			xlabel={budget $n$}, ylabel={$\E[\text{width}]$},
			xtick={100,300,500,700,1000},
			grid=major, grid style={gray!25},
			tick label style={font=\footnotesize}, label style={font=\small},
			/pgf/number format/1000 sep={},
			scaled y ticks=false, yticklabel style={/pgf/number format/fixed, /pgf/number format/precision=3},
			legend style={font=\footnotesize, draw=none, at={(0.5,1.03)}, anchor=south},
			legend columns=3, legend cell align=left,
		]
			\addlegendimage{srscolour, line width=1pt}\addlegendentry{SRS}
			\addlegendimage{pscolour, line width=1pt}\addlegendentry{PS}
			\addlegendimage{cpscolour, line width=2.4pt, opacity=0.5}\addlegendentry{CPS}
			\addplot[srscolour, line width=1pt, solid, forget plot] coordinates {(100,21602) (300,12709) (500,9886.7) (700,8355.6) (1000,6987.9)};
			\addplot[pscolour, line width=1pt, solid, forget plot] coordinates {(100,18457) (300,10820) (500,8402.1) (700,7102.7) (1000,5941)};
			\addplot[cpscolour, line width=2.4pt, opacity=0.5, solid, forget plot] coordinates {(100,18264) (300,10685) (500,8305.1) (700,7028.8) (1000,5883.5)};
		\end{axis}
\end{tikzpicture}
\caption{Texas: CPS beats PS significantly in expected width, but not in RMSE (\cref{tab:rmse_significance}).}
\label{fig:saipe_width_TX}
\end{subfigure}
\caption{Expected width of confidence intervals as a function of the budget $n$ for SRS, PS, and CPS on SAIPE, where $f$ is trained on California and deployed in each target state. CPS is narrower than PS at every budget in Alabama and Oregon, the two states where $f$ is most biased, and CPS and PS are close in Massachusetts and Texas. Both weighted designs are narrower than SRS in every state.}
\label{fig:saipe_width}
\end{figure}

\begin{table}[t]
\centering
\small
\begin{tabular}{lrrrrrr}
\toprule
 & & \multicolumn{2}{c}{RMSE vs.} & \multicolumn{2}{c}{$\mathbb{E}$[width] vs.} & \\
\cmidrule(lr){3-4} \cmidrule(lr){5-6}
Target & $f$ bias & SRS & PS & SRS & PS & CPS cov. \\
\midrule
AL & $+29\%$ & $-15\%$ & $-3\%$ & $-16\%$ & $-4\%$ & 0.894 \\
OR & $+23\%$ & $-15\%$ & $-4\%$ & $-15\%$ & $-3\%$ & 0.899 \\
MA & $+8\%$ & $-15\%$ & $-.6\%$$^\dagger$ & $-15\%$ & $-.2\%$$^\dagger$ & 0.897 \\
TX & $+14\%$ & $-15\%$ & $+.4\%$$^\dagger$ & $-16\%$ & $-1\%$ & 0.891 \\
\bottomrule
\end{tabular}
\caption{SAIPE experiment where $f$ is trained on CA and used to design a sample and estimate the mean in a different state. 
	The two states with the highest bias show a significant advantage of CPS over PS and SRS in both RMSE and expected width, which aligns with the prediction of \cref{corr:matrix_variance_reduction_cps}. 
	Cells are averaged over budgets and indicated with a $\dagger$ when a difference is not significant. 
	We also note that the coverage of CPS is close to the target $0.90$, and that the RMSE results show that the difference in expected width is not an artifact of variance estimation.}
\label{tab:saipe_summary}
\end{table}

In \Cref{tab:saipe_summary}, we observe that CPS does not perform significantly worse than SRS or PS.
CPS performs significantly better in terms of width and RMSE in the two states, AL and OR, where the bias is largest.
In AL and OR, the respective biases of $f$ are 28.6\% and 23.2\%, and CPS reduces RMSE relative to PS by 3.1\% and 3.7\% and expected width by 4.0\% and 2.6\%. 
In MA and TX, the differences in RMSE are not significant, but in TX the 1.1\% reduction in expected width is significant.
CPS improves substantially on SRS with reductions across states in RMSE of 14.8--15.5\% and in expected width of 15.2--16.1\%. 
Lastly, CPS coverage exceeds 0.89 in all states.

\subsubsection{ASI Replication with CPS}\label{exp:asi}

Finally, we compare SRS, CPS, and PS using DE on the datasets from \citet{zrnic2024asi}.  We briefly summarize the tasks here, and point the interested reader to~\citet{zrnic2024asi} for more details. (1) \textbf{AlphaFold}: On this dataset, the task is to estimate an odds ratio on a protein-folding task, where $f$ is DeepMind's AlphaFold \citep{jumper2021alphafold} and $\*\pi$ is chosen as in~\citet{zrnic2024asi} via an \enquote{uncertainty} heuristic as $\pi_i \propto \min(f(x_i), 1-f(x_i))$.  (2)  \textbf{Census}: On this dataset, the task is to estimate a linear regression coefficient of age when regressing income on age and sex, where $f$ is XGBoost trained on a held-out labeled dataset ($N_{\text{train}} = 190{,}045$) and $\pi_i$ is proportional to a model trained to predict the residual $|Y-f(X)|$ times the influence on the slope $(1,0)(X^\top X)^{-1}X_i$.  (3) \textbf{Pew}: On this dataset, the task is to estimate the mean approval rating of Biden/Trump from a Pew survey, where $f$ is XGBoost trained on a held-out labeled dataset ($N_{\text{train}} = 6189$) and $\pi_i \propto \min(f(x_i), 1-f(x_i))$. 

In \Cref{tab:asi_width}, we show the average confidence interval width for each method for different sample size budgets, noting that for PS, the budget is the expected sample size, while for CPS, it is the fixed sample size.  We generally observe that CPS and PS have comparable variances in this setting, which indicates that $v$ is small for these predictors and policies.

\begin{table}[t]
\centering
\begin{tabular}{lllll@{\hskip 1em}lllll}
\toprule
Algorithm & $n=300$ & $500$ & $700$ & $1000$ & Algorithm & $n=300$ & $500$ & $700$ & $1000$ \\
\midrule
\multicolumn{5}{l}{\textbf{AlphaFold}} & \multicolumn{5}{@{}l}{\textbf{Census}} \\
SRS & 1.47 & 1.10 & .91 & .75 & SRS & 445 & 353 & 300 & 251 \\
PS & 1.37 & 1.03 & .84 & .68 & PS & 262 & 205 & 174 & 146 \\
CPS & 1.37 & 1.03 & .84 & .67 & CPS & 259 & 202 & 171 & 143 \\
\midrule
\multicolumn{5}{l}{\textbf{Pew (Biden)}} & \multicolumn{5}{@{}l}{\textbf{Pew (Trump)}} \\
SRS & .06 & .05 & .04 & .03 & SRS & .06 & .04 & .04 & .03 \\
PS & .05 & .04 & .03 & .03 & PS & .05 & .04 & .03 & .03 \\
CPS & .05 & .04 & .03 & .03 & CPS & .05 & .04 & .03 & .03 \\
\bottomrule
\end{tabular}
\caption{Expected confidence interval (CI) width on the datasets from ASI \citep{zrnic2024asi}. Smaller widths are better. For each dataset, we compare SRS, PS, and CPS using DE and the inclusion probabilities of \citet{zrnic2024asi} described in \cref{exp:asi}. Coverage is similar across all methods (see \cref{app:coverage}). Notably, PS and CPS have near-identical expected widths, which the theory predicts when $v = \E_m[w^\star(1-\pis)]$ is small (see \cref{sec:cps:mest}). This holds even for Census, where $f$ is biased in the mean (\cref{tab:rmse_significance}): $v$ weights the residuals $w^\star$ by $1-\pis$, so a bias in the mean of $Y$ need not make $v$ large.
	}
\label{tab:asi_width}
\end{table}

\subsection{Consistent Variance Estimation for CPS}
\label{app:matei-varapprox}

Every estimator in our experiments is either a Horvitz--Thompson mean $\hat\mu^w_{\*\pi} = \frac{1}{N}\sum_{i} \frac{s_i}{\pi_i} w_i$ of fixed values $w_i$, or a smooth function of such means.
For mean estimation (\cref{exp:mean0-Y,exp:pew_biased_f,exp:saipe}, and Pew in~\cref{exp:asi}), the values are $w_i = Y_i$ for HT and $w_i = Y_i - f(X_i)$ for DE, since $\frac{1}{N}\sum_i f(X_i)$ is computed exactly.
For Census, the target is a coordinate of a finite-population regression coefficient, which is a linear function of a vector-valued HT mean (\cref{rem:census_linear}).
For AlphaFold, the target is a log odds ratio, which is a smooth function of two group means (\cref{rem:alphafold_delta}).
In every case, the confidence interval is a Wald interval built from a design-based variance estimator: the Sen--Yates--Grundy (SYG) estimator under SRS, the HT estimator $\frac{1}{N^2}\sum_{k \in \*s} w_k^2 (1-\pi_k)/\pi_k^2$ under PS, and the estimator below under CPS.

The HT and SYG variance estimators under CPS require the pairwise inclusion probabilities $\pi_{ij}$, which take $O(nN^2)$ time to compute with the recursion of~\cref{app:cps-implementation}.
\citet{Matei2005variance} evaluate a family of approximate variance estimators that avoid this cost by replacing $\pi_{ij}$ with a formula that depends only on $\*\pi$.
We use the \enquote{Deville estimator}, which needs only the first-order inclusion probabilities of the sampled units, so it costs $O(n)$:
\al{
	& \widehat{\V}[\hat\mu^w_{\*\pi}]
	= \frac{1}{N^2}\sum_{k\in\*s} \frac{c_k}{\pi_k^2}\bigparens{w_k - \pi_k \hat\beta}^2,
	\quad c_k := \frac{n}{n-1}(1-\pi_k),
	\quad \hat\beta := \frac{\sum_{l\in\*s} c_l\, w_l/\pi_l}{\sum_{l\in\*s} c_l}.
}
For vector-valued $w_i \in \R^q$, we replace the square by the outer product $(w_k - \pi_k \hat\beta)(w_k - \pi_k\hat\beta)^\top$, with $\hat\beta \in \R^q$ defined by the same formula.
The estimator is non-negative (positive semi-definite) by construction.

We show that $N\,\widehat{\V}[\hat\mu^w_{\*\pi}]$ is a consistent plug-in estimator of the CPS design variance $\sigma^2_{\textup{cps}}(\*\pi, N)$ of~\cref{lemma:ht_design}, so the resulting Wald intervals have asymptotically correct coverage.
We use one observation about the proof of~\cref{lemma:ht_design}.

\begin{remark}[Consistency Needs Only Second Moments]\label{rem:ht_consistency_second_moment}
	The consistency part of~\cref{lemma:ht_design} holds under the weaker condition $\limsup_{N} \frac{1}{N}\sum_{i=1}^N w_i^2 < \infty$: under both designs, its proof uses the fourth-moment bound only to conclude that $\frac{1}{N}\sum_i w_i^2 = O(1)$, and then bounds the design variance by $\frac{2}{\gamma N}\cdot\frac{1}{N}\sum_i w_i^2$.
\end{remark}

\begin{lemma}[Consistency of the Deville Variance Estimator Under CPS]\label{lemma:deville_consistency}
	Assume the setting of~\cref{lemma:ht_design} under the CPS design $\p_{\textup{cps}(\*\pi)}$: fix $\omega$ such that~\cref{asmp:design_assumption_constraints_on_pi} holds at $\omega$, and assume $\limsup_{N} \frac{1}{N}\sum_{i=1}^N w_i^4 < \infty$.
	Then, in design probability,
	\begin{enumerate}
		\item \textup{(Consistency.)} $N\,\widehat{\V}[\hat\mu^w_{\*\pi}] - \sigma^2_{\textup{cps}}(\*\pi, N) \cip 0$.
		\item \textup{(Studentized CLT.)} If $\sigma^2_{\textup{cps}}(\*\pi, N)$ is bounded away from zero, then
		      \begin{equation*}
			      \frac{\hat\mu^w_{\*\pi} - \mu^w_N}{\sqrt{\widehat{\V}[\hat\mu^w_{\*\pi}]}} \cid \mathcal{N}(0,1),
		      \end{equation*}
		      so the interval $\hat\mu^w_{\*\pi} \pm z_{1-\alpha/2}\,\widehat{\V}[\hat\mu^w_{\*\pi}]^{1/2}$ covers $\mu^w_N$ with design probability tending to $1-\alpha$.
	\end{enumerate}
\end{lemma}

\begin{proof}
	The proof has three steps: we write both $\sigma^2_{\textup{cps}}$ and the estimator as the same continuous function of three population means, show that the estimator evaluates this function at the HT estimates of those means, and conclude by consistency of each HT estimate.

	\textit{Step 1: Population Quantities.}
	Define the population means
	\al{
		& A_N := \frac{1}{N}\sum_{i=1}^N w_i^2\,\frac{1-\pi_i}{\pi_i},
		\qquad B_N := \frac{1}{N}\sum_{i=1}^N w_i (1-\pi_i),
		\qquad C_N := \frac{1}{N}\sum_{i=1}^N \pi_i (1-\pi_i),
	}
	so that $\beta = B_N / C_N$ in~\cref{lemma:ht_design}.
	Expanding the square in $\sigma^2_{\textup{cps}}$,
	\flal{
		&   \sigma^2_{\textup{cps}}(\*\pi, N)
		    = \frac{1}{N}\sum_{i=1}^N \bigparens{w_i^2 \frac{1-\pi_i}{\pi_i} - 2\beta\, w_i(1-\pi_i) + \beta^2 \pi_i(1-\pi_i)}
		    = A_N - 2\beta B_N + \beta^2 C_N
		    = A_N - \frac{B_N^2}{C_N}.
		    &&\tag{$\beta = B_N/C_N$}
	}

	\textit{Step 2: The Estimator Is the HT Plug-In.}
	Define the HT estimators of $A_N$, $B_N$, and $C_N$,
	\al{
		& \hat A := \frac{1}{N}\sum_{k\in\*s} \frac{w_k^2(1-\pi_k)}{\pi_k^2},
		\qquad \hat B := \frac{1}{N}\sum_{k\in\*s} \frac{w_k(1-\pi_k)}{\pi_k},
		\qquad \hat C := \frac{1}{N}\sum_{k\in\*s} (1-\pi_k).
	}
	Each is $\frac{1}{N}\sum_i \frac{s_i}{\pi_i} v_i$ for the fixed values $v_i = w_i^2(1-\pi_i)/\pi_i$, $w_i(1-\pi_i)$, and $\pi_i(1-\pi_i)$, respectively.
	The factor $n/(n-1)$ cancels in $\hat\beta$, so $\hat\beta = \hat B / \hat C$, and expanding the square in the estimator,
	\flal{
		&   N\,\widehat{\V}[\hat\mu^w_{\*\pi}]
		    = \frac{n}{n-1}\cdot\frac{1}{N}\sum_{k\in\*s} \frac{1-\pi_k}{\pi_k^2}\bigparens{w_k^2 - 2\pi_k\hat\beta\, w_k + \pi_k^2\hat\beta^2}
		    = \frac{n}{n-1}\bigparens{\hat A - 2\hat\beta\hat B + \hat\beta^2 \hat C}
		    = \frac{n}{n-1}\bigparens{\hat A - \frac{\hat B^2}{\hat C}}.
		    &&
	}
	The estimator is undefined when $\hat C = 0$, i.e., when every sampled unit has $\pi_k = 1$; this cannot occur since $\pi_k \le 1-\gamma$.

	\textit{Step 3: Consistency.}
	By~\cref{asmp:design_assumption_constraints_on_pi}, $\gamma \le \pi_i \le 1-\gamma$, so the values $v_i$ of Step 2 satisfy
	\al{
		& \frac{1}{N}\sum_i \bigparens{w_i^2\frac{1-\pi_i}{\pi_i}}^2 \le \frac{1}{\gamma^2}\cdot\frac{1}{N}\sum_i w_i^4,
		\qquad \frac{1}{N}\sum_i w_i^2(1-\pi_i)^2 \le \frac{1}{N}\sum_i w_i^2,
		\qquad \frac{1}{N}\sum_i \pi_i^2(1-\pi_i)^2 \le 1,
	}
	which are $O(1)$ by the fourth-moment assumption (and Jensen's inequality for the middle term).
	By~\cref{rem:ht_consistency_second_moment}, $\hat A - A_N$, $\hat B - B_N$, and $\hat C - C_N$ all converge to zero in design probability.
	The population means are bounded: $|A_N| \le \frac{1}{\gamma}\frac{1}{N}\sum_i w_i^2 = O(1)$, $|B_N| \le (\frac{1}{N}\sum_i w_i^2)^{1/2} = O(1)$, and $C_N \ge \gamma(1-\gamma) > 0$.
	So with design probability tending to one, $\hat C \ge C_N/2 \ge \gamma(1-\gamma)/2$ and $|\hat B| \le |B_N| + 1$. On this event,
	\flal{
		&   \biggl|\frac{\hat B^2}{\hat C} - \frac{B_N^2}{C_N}\biggr|
		    \le \frac{|\hat B - B_N|\,(|\hat B| + |B_N|)}{\hat C} + \frac{B_N^2\,|\hat C - C_N|}{\hat C\, C_N}
		    \le \frac{2(2|B_N|+1)}{\gamma(1-\gamma)}|\hat B - B_N| + \frac{2 B_N^2}{\gamma^2(1-\gamma)^2}|\hat C - C_N|,
		    &&\tag{$\hat C \ge \gamma(1-\gamma)/2$, $C_N \ge \gamma(1-\gamma)$}
	}
	which converges to zero in design probability because $B_N = O(1)$.
	Hence $\hat A - \hat B^2/\hat C - \sigma^2_{\textup{cps}}(\*\pi, N) \cip 0$, and $\hat A - \hat B^2/\hat C$ is $O_P(1)$ because $\sigma^2_{\textup{cps}}(\*\pi, N) \le A_N = O(1)$.
	Finally, $n \to \infty$ under~\cref{asmp:design_assumption_constraints_on_pi}, so $\frac{n}{n-1} \to 1$, which proves (i).

	For (ii), (i) and $\sigma^2_{\textup{cps}}(\*\pi, N) \ge c > 0$ give $N\,\widehat{\V}[\hat\mu^w_{\*\pi}] / \sigma^2_{\textup{cps}}(\*\pi, N) \cip 1$.
	The claim follows from the asymptotic normality part of~\cref{lemma:ht_design} and Slutsky's lemma, since
	\begin{equation*}
		\frac{\hat\mu^w_{\*\pi} - \mu^w_N}{\sqrt{\widehat{\V}[\hat\mu^w_{\*\pi}]}}
		= \frac{\sqrt N\,(\hat\mu^w_{\*\pi} - \mu^w_N)}{\sigma_{\textup{cps}}(\*\pi, N)} \cdot \bigparens{\frac{\sigma^2_{\textup{cps}}(\*\pi, N)}{N\,\widehat{\V}[\hat\mu^w_{\*\pi}]}}^{1/2}.
	\end{equation*}
\end{proof}

Under PS, the HT variance estimator is $\hat A/N$, and the same argument (Step 3 alone) gives $\hat A - \sigma^2_{\textup{ps}}(\*\pi, N) \cip 0$.

The next lemma extends~\cref{lemma:deville_consistency} to vector-valued means and to the scalar functions of them that we report.

\begin{lemma}[Vector Means, Linear Functionals, and Smooth Functions]\label{lemma:deville_vector}
	Assume the setting of~\cref{lemma:ht_design_vector} under the CPS design, with $w_i \in \R^q$ and $\limsup_N \frac{1}{N}\sum_i \|w_i\|^4 < \infty$, and let $\widehat{\V}[\hat\mu^w_{\*\pi}] \in \R^{q\times q}$ be the matrix Deville estimator. Then, in design probability:
	\begin{enumerate}
		\item \textup{(Consistency.)} $N\,\widehat{\V}[\hat\mu^w_{\*\pi}] - \Sigma_{\textup{cps}}(\*\pi, N) \cip 0$ entrywise.
	\end{enumerate}
	Suppose in addition that $\Sigma_{\textup{cps}}(\*\pi, N) \to \Sigma_\infty$, a positive definite matrix. Then:
	\begin{enumerate}[resume]
		\item \textup{(Linear Functionals.)} Let $\ell_N \in \R^q$ be design-fixed vectors with $\ell_N \to \ell \neq 0$. Then
		      \begin{equation*}
			      \frac{\ell_N^\top(\hat\mu^w_{\*\pi} - \mu^w_N)}{\sqrt{\ell_N^\top\,\widehat{\V}[\hat\mu^w_{\*\pi}]\,\ell_N}} \cid \mathcal{N}(0,1),
		      \end{equation*}
		      and $\ell_N^\top\,\widehat{\V}[\hat\mu^w_{\*\pi}]\,\ell_N$ equals the scalar Deville estimator applied to the values $\ell_N^\top w_i$.
		\item \textup{(Smooth Functions.)} Suppose $\mu^w_N \to \mu_\infty$, and let $h : \R^q \to \R$ be continuously differentiable on a neighborhood of $\mu_\infty$ with $\nabla h(\mu_\infty) \neq 0$. Then
		      \begin{equation*}
			      \frac{h(\hat\mu^w_{\*\pi}) - h(\mu^w_N)}{\sqrt{\nabla h(\hat\mu^w_{\*\pi})^\top\,\widehat{\V}[\hat\mu^w_{\*\pi}]\,\nabla h(\hat\mu^w_{\*\pi})}} \cid \mathcal{N}(0,1).
		      \end{equation*}
	\end{enumerate}
\end{lemma}

\begin{proof}
	\textit{(i)} Steps 1 and 2 of the proof of~\cref{lemma:deville_consistency} go through verbatim with $A_N, \hat A \in \R^{q\times q}$ the means of $w_iw_i^\top(1-\pi_i)/\pi_i$, $B_N, \hat B \in \R^q$ the means of $w_i(1-\pi_i)$, and $C_N, \hat C$ as before, giving
	\al{
		& \Sigma_{\textup{cps}}(\*\pi, N) = A_N - \frac{B_N B_N^\top}{C_N},
		\qquad N\,\widehat{\V}[\hat\mu^w_{\*\pi}] = \frac{n}{n-1}\bigparens{\hat A - \frac{\hat B \hat B^\top}{\hat C}}.
	}
	Each entry of $\hat A$ and $\hat B$ is an HT estimator of fixed values bounded by $\|w_i\|^2/\gamma$ and $\|w_i\|$ in absolute value, so~\cref{rem:ht_consistency_second_moment} applies entrywise, and Step 3 applies to each entry.

	\textit{(ii)} By~\cref{lemma:ht_design_vector}, $\sqrt N(\hat\mu^w_{\*\pi} - \mu^w_N) \cid \mathcal{N}_q(0, \Sigma_\infty)$, and in particular it is $O_P(1)$. Write
	\flal{
		&   \sqrt N\,\ell_N^\top(\hat\mu^w_{\*\pi} - \mu^w_N)
		    = \ell^\top \sqrt N(\hat\mu^w_{\*\pi} - \mu^w_N) + (\ell_N - \ell)^\top \sqrt N(\hat\mu^w_{\*\pi} - \mu^w_N)
		    \cid \mathcal{N}(0, \ell^\top\Sigma_\infty\ell),
		    &&\tag{$(\ell_N - \ell)^\top O_P(1) = o_P(1)$; Slutsky}
	}
	where $\ell^\top \Sigma_\infty \ell > 0$ since $\Sigma_\infty$ is positive definite and $\ell \neq 0$.
	By (i) and $\Sigma_{\textup{cps}}(\*\pi, N) \to \Sigma_\infty$, $N\,\widehat{\V}[\hat\mu^w_{\*\pi}] \cip \Sigma_\infty$, so $N\,\ell_N^\top\widehat{\V}[\hat\mu^w_{\*\pi}]\ell_N \cip \ell^\top\Sigma_\infty\ell$, and Slutsky's lemma gives the studentized limit.
	For the last claim, $\ell_N^\top\hat\beta$ is the $\hat\beta$ of the values $\ell_N^\top w_i$, so $\ell_N^\top(w_k - \pi_k\hat\beta)(w_k - \pi_k\hat\beta)^\top\ell_N = (\ell_N^\top w_k - \pi_k\,\ell_N^\top\hat\beta)^2$.

	\textit{(iii)} Let $U$ be a convex neighborhood (a ball) of $\mu_\infty$ on which $h$ is continuously differentiable.
	By the consistency part of~\cref{lemma:ht_design_vector} and $\mu^w_N \to \mu_\infty$, we have $\hat\mu^w_{\*\pi} \cip \mu_\infty$, so with design probability tending to one both $\hat\mu^w_{\*\pi}$ and $\mu^w_N$ lie in $U$.
	On that event, the mean value theorem gives $h(\hat\mu^w_{\*\pi}) - h(\mu^w_N) = \nabla h(\tilde\mu_N)^\top(\hat\mu^w_{\*\pi} - \mu^w_N)$ for some $\tilde\mu_N$ on the segment between $\mu^w_N$ and $\hat\mu^w_{\*\pi}$.
	Then $\tilde\mu_N \cip \mu_\infty$, so $\nabla h(\tilde\mu_N) \cip \nabla h(\mu_\infty)$ by continuity, and the argument of (ii) with $\ell_N$ replaced by $\nabla h(\tilde\mu_N)$ and $\ell = \nabla h(\mu_\infty)$ gives $\sqrt N\bigparens{h(\hat\mu^w_{\*\pi}) - h(\mu^w_N)} \cid \mathcal{N}(0, \nabla h(\mu_\infty)^\top\Sigma_\infty\nabla h(\mu_\infty))$; here the step $(\ell_N - \ell)^\top O_P(1) = o_P(1)$ needs only $\ell_N \cip \ell$, not that $\ell_N$ is design-fixed.
	Likewise $\nabla h(\hat\mu^w_{\*\pi}) \cip \nabla h(\mu_\infty)$, so $N\,\nabla h(\hat\mu^w_{\*\pi})^\top\widehat{\V}[\hat\mu^w_{\*\pi}]\nabla h(\hat\mu^w_{\*\pi}) \cip \nabla h(\mu_\infty)^\top\Sigma_\infty\nabla h(\mu_\infty) > 0$, and Slutsky's lemma concludes.
\end{proof}

The conditions $\mu^w_N \to \mu_\infty$ and $\Sigma_{\textup{cps}}(\*\pi, N) \to \Sigma_\infty$ hold for $P$-almost every $\omega$ under our model and design assumptions: the first by the strong law of large numbers when $\E_m\|w\| < \infty$, and the second by the argument of~\cref{lemma:convergence_of_covariance} with $w_i(\theta_N)$ replaced by the fixed values $w_i$.
\Cref{lemma:deville_vector} covers the two estimands in our experiments that are not means.

\begin{remark}[Census Regression Coefficient]\label{rem:census_linear}
	Let $\*X \in \R^{N\times q}$ be the fully observed covariate matrix, $\*Y$ the labels, and $\*F = (f(X_i))_{i}$ the predictions.
	The target is a coordinate $\theta_N = e_1^\top(\*X^\top\*X)^{-1}\*X^\top\*Y$ of the finite-population least-squares coefficient, and the estimator of~\citet{zrnic2024asi} with squared loss is $\hat\theta = e_1^\top(\*X^\top\*X)^{-1}\bigparens{\*X^\top\*F + \sum_{i\in\*s} X_i(Y_i - f(X_i))/\pi_i}$.
	With $w_i := X_i(Y_i - f(X_i)) \in \R^q$ and $\ell_N := (\frac{1}{N}\*X^\top\*X)^{-1} e_1$,
	\begin{equation*}
		\hat\theta - \theta_N = \ell_N^\top\bigparens{\hat\mu^w_{\*\pi} - \mu^w_N}
	\end{equation*}
	exactly, and $\ell_N$ is design-fixed since $\*X$ is observed for every unit.
	If $\E_m[XX^\top]$ is invertible, $\ell_N \to \E_m[XX^\top]^{-1}e_1 \neq 0$ almost surely, so~\cref{lemma:deville_vector}(ii) applies when $\E_m[\|X\|^4 (Y - f(X))^4] < \infty$.
	The variance estimate is the scalar Deville estimator applied to the values $\ell_N^\top w_i$.
	No estimate of $\theta$ enters the variance estimate, because $X$ is observed for every unit and the estimating equation is linear in $\theta$.
\end{remark}

\begin{remark}[AlphaFold Log Odds Ratio]\label{rem:alphafold_delta}
	The two groups $g \in \{0,1\}$ are sampled with independent CPS designs, and each group mean is estimated with DE, so $\hat\mu_g - \mu_{g,N}$ is the HT error of the values $w_i = Y_i - f(X_i)$ in group $g$.
	The target is $h(\mu_{1,N}) - h(\mu_{0,N})$ with $h = \mathrm{logit}$, and $\nabla h(\mu) = 1/(\mu(1-\mu))$.
	If each $\mu_{g,N}$ converges to a limit in $(0,1)$,~\cref{lemma:deville_vector}(iii) applies to each group separately, and the variance estimate of the log odds ratio is the sum of the two delta-method variances $\widehat{\V}_g/(\hat\mu_g(1-\hat\mu_g))^2$.
	Because the designs are independent, the characteristic function of the sum of the two standardized errors factorizes, and the studentized log odds ratio is asymptotically $\mathcal{N}(0,1)$ by Slutsky's lemma.
\end{remark}

We also check the variance estimates empirically. For each experiment and design, \cref{tab:variance_quality} compares the average estimated variance over the 5000 seeds with the Monte Carlo variance of the point estimates across seeds.
The ratio is between 0.97 and 1.05 for every dataset except AlphaFold, where the variance is underestimated for both PS (0.92) and CPS (0.92). This matches the slight undercoverage of both designs on AlphaFold (0.87 and 0.86, against 0.88--0.91 for PS and CPS on every other dataset).

\begin{table}[t]
\centering
\begin{tabular}{llrrr}
\toprule
Dataset & Method & $\mathbb{E}[\hat v]/\mathrm{MCvar}$ & CI width / SRS & Coverage \\
\midrule
Synthetic DGP0, HT & SRS & 0.99 & 1.00 & 0.89 \\
 & PS & 1.00 & 1.00 & 0.90 \\
 & CPS & 1.00 & 1.00 & 0.90 \\
\midrule
Synthetic DGP1, HT & SRS & 0.99 & 1.00 & 0.89 \\
 & PS & 1.02 & 1.35 & 0.90 \\
 & CPS & 1.00 & 1.00 & 0.90 \\
\midrule
Synthetic DGP0, $f_{lin}$, DE & SRS & 1.01 & 1.00 & 0.90 \\
 & PS & 1.05 & 1.00 & 0.91 \\
 & CPS & 1.01 & 1.00 & 0.90 \\
\midrule
Synthetic DGP0, $f_{bias}$, DE & SRS & 1.01 & 1.00 & 0.90 \\
 & PS & 1.02 & 1.42 & 0.90 \\
 & CPS & 1.01 & 1.00 & 0.90 \\
\midrule
Pew-Biden, shifted $f$ (rate) & SRS & 1.01 & 1.00 & 0.90 \\
 & PS & 1.02 & 1.05 & 0.90 \\
 & CPS & 0.99 & 0.87 & 0.89 \\
\midrule
Pew-Trump, shifted $f$ (rate) & SRS & 0.99 & 1.00 & 0.90 \\
 & PS & 1.00 & 1.98 & 0.88 \\
 & CPS & 0.99 & 1.68 & 0.88 \\
\midrule
SAIPE AL (\$) & SRS & 1.02 & 1.00 & 0.90 \\
 & PS & 1.00 & 0.87 & 0.90 \\
 & CPS & 0.99 & 0.84 & 0.89 \\
\midrule
SAIPE OR (\$) & SRS & 1.01 & 1.00 & 0.90 \\
 & PS & 0.99 & 0.87 & 0.90 \\
 & CPS & 1.01 & 0.84 & 0.90 \\
\midrule
SAIPE MA (\$) & SRS & 1.02 & 1.00 & 0.90 \\
 & PS & 1.00 & 0.85 & 0.90 \\
 & CPS & 1.01 & 0.85 & 0.90 \\
\midrule
SAIPE TX (\$) & SRS & 1.00 & 1.00 & 0.90 \\
 & PS & 1.00 & 0.85 & 0.90 \\
 & CPS & 0.97 & 0.84 & 0.89 \\
\midrule
ASI Census (\$/yr) & SRS & 1.00 & 1.00 & 0.86 \\
 & PS & 0.99 & 0.59 & 0.89 \\
 & CPS & 1.02 & 0.58 & 0.89 \\
\midrule
ASI Pew-Biden (rate) & SRS & 1.00 & 1.00 & 0.90 \\
 & PS & 0.99 & 0.88 & 0.90 \\
 & CPS & 0.99 & 0.88 & 0.90 \\
\midrule
ASI Pew-Trump (rate) & SRS & 0.99 & 1.00 & 0.90 \\
 & PS & 1.02 & 0.88 & 0.90 \\
 & CPS & 0.98 & 0.88 & 0.89 \\
\midrule
ASI AlphaFold (log-OR) & SRS & 0.98 & 1.00 & 0.89 \\
 & PS & 0.92 & 0.91 & 0.87 \\
 & CPS & 0.92 & 0.90 & 0.86 \\
\bottomrule
\end{tabular}
\caption{
	Quality of the design-based variance estimators. For each dataset and design: the ratio of the average estimated variance to the Monte Carlo variance of the point estimate over 5000 seeds, the expected CI width relative to SRS, and the empirical coverage of nominal 90\% intervals.
	All ratios are close to 1 and all coverages are close to 0.90 (with slight undercoverage on AlphaFold for PS and CPS, and on Census for SRS), so comparing designs by RMSE and expected CI width is valid.}
\label{tab:variance_quality}
\end{table}

\subsection{ASI Replication}\label{app:coverage}
The following tables are for the ASI replication experiment. 
In \Cref{tab:asi_coverage} we show that CPS achieves valid confidence intervals, like the other two algorithms.
\Cref{tab:asi_n_variance} shows that the variance in the sample size for PS is non-trivial. 

\begin{table}[h]
\centering
\begin{tabular}{lrrrr}
\toprule
Algorithm & $n=300$ & $n=500$ & $n=700$ & $n=1000$ \\
\midrule
\multicolumn{5}{l}{\text{AlphaFold}} \\
SRS & 0.90 & 0.90 & 0.90 & 0.89 \\
PS & 0.85 & 0.87 & 0.89 & 0.89 \\
CPS & 0.84 & 0.86 & 0.88 & 0.89 \\
\midrule
\multicolumn{5}{l}{\text{Census}} \\
SRS & 0.86 & 0.87 & 0.88 & 0.88 \\
PS & 0.89 & 0.89 & 0.89 & 0.89 \\
CPS & 0.90 & 0.90 & 0.90 & 0.90 \\
\midrule
\multicolumn{5}{l}{\text{Pew (Biden)}} \\
SRS & 0.91 & 0.90 & 0.89 & 0.90 \\
PS & 0.90 & 0.90 & 0.90 & 0.90 \\
CPS & 0.89 & 0.90 & 0.90 & 0.89 \\
\midrule
\multicolumn{5}{l}{\text{Pew (Trump)}} \\
SRS & 0.90 & 0.91 & 0.90 & 0.90 \\
PS & 0.90 & 0.90 & 0.90 & 0.90 \\
CPS & 0.89 & 0.90 & 0.90 & 0.89 \\
\bottomrule
\end{tabular}
\caption{Empirical coverage of nominal 90\% confidence intervals for SRS, PS, and CPS on the datasets from ASI.}
\label{tab:asi_coverage}
\end{table}

\begin{table}[h]
\centering
\begin{tabular}{lrrrr}
\toprule
Algorithm & $n=300$ & $n=500$ & $n=700$ & $n=1000$ \\
\midrule
\multicolumn{5}{l}{\text{AlphaFold}} \\
SRS & 0 & 0 & 0 & 0 \\
PS & 278 & 443 & 602 & 804 \\
CPS & 0 & 0 & 0 & 0 \\
\midrule
\multicolumn{5}{l}{\text{Census}} \\
SRS & 0 & 0 & 0 & 0 \\
PS & 314 & 518 & 723 & 1025 \\
CPS & 0 & 0 & 0 & 0 \\
\midrule
\multicolumn{5}{l}{\text{Pew (Biden)}} \\
SRS & 0 & 0 & 0 & 0 \\
PS & 275 & 427 & 580 & 780 \\
CPS & 0 & 0 & 0 & 0 \\
\midrule
\multicolumn{5}{l}{\text{Pew (Trump)}} \\
SRS & 0 & 0 & 0 & 0 \\
PS & 267 & 425 & 590 & 793 \\
CPS & 0 & 0 & 0 & 0 \\
\bottomrule
\end{tabular}
\caption{Variance of the realized sample size $\tilde{n}$ (zero for SRS and CPS by design). Under PS, the exact variance is $\sum_i \pi_i(1-\pi_i) < n$. The Census entries slightly exceed $n$, which we attribute to Monte Carlo error over the seeds.}
\label{tab:asi_n_variance}
\end{table}

\clearpage
\section{The Case for CPS: Theoretical Results}\label{app:cps}\label{app:theory_appendix}

Recall the following theorem, which we stated informally in the main text.  

\MainAsymptoticResult*

We now demonstrate this result formally, including sufficient regularity conditions.  We prove the result for general Z-estimation, which includes M-estimation as a special case.

\renewcommand{\g}[1][\theta]{g_{#1}}
\renewcommand{\giY}[1][\theta]{g_{#1}(X_i, Y_i)}
\renewcommand{\giy}[1][\theta]{g_{#1}(x_i, y_i)}
\renewcommand{\gY}[1][\theta]{g_{#1}(X, Y)}
\renewcommand{\gy}[1][\theta]{g_{#1}(x, y)}
\renewcommand{\giF}[1][\theta]{g_{#1}(X_i, f(X_i))}
\renewcommand{\gF}[1][\theta]{g_{#1}(X, f(X))}
\renewcommand{\gf}[1][\theta]{g_{#1}(x, f(x))}
\newcommand{\h}[1][\theta]{\nabla g_{#1}}
\renewcommand{\hiy}[1][\theta]{\nabla g_{#1}(X_i, Y_i)}
\renewcommand{\hy}[1][\theta]{\nabla g_{#1}(X, Y)}
\renewcommand{\hif}[1][\theta]{\nabla g_{#1}(X_i, f(X_i))}
\renewcommand{\hf}[1][\theta]{\nabla g_{#1}(X, f(X))}

\subsection{Definitions}
\label{app:superpopulation_defs}

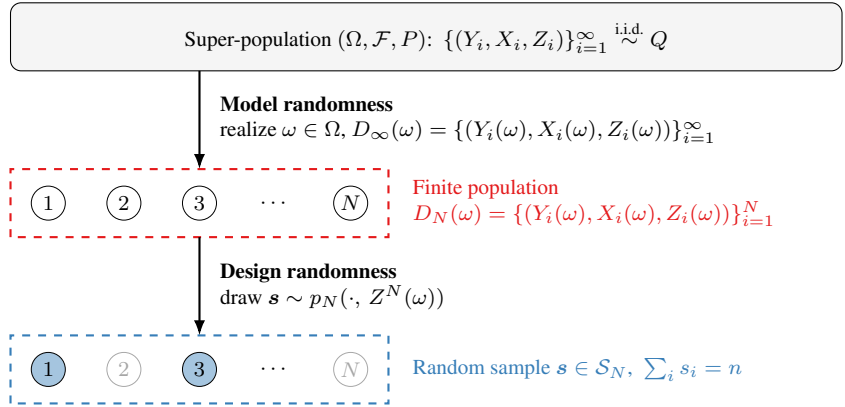
\begin{figure}[!htb]
	\centering
	\begin{tikzpicture}[
			every node/.style={font=\footnotesize},
			unit/.style={circle, draw, minimum size=4.5mm, inner sep=0pt, line width=0.3pt},
			selected/.style={circle, draw, fill=betterblue!45, minimum size=4.5mm, inner sep=0pt, line width=0.3pt},
			faded/.style={circle, draw, minimum size=4.5mm, inner sep=0pt, line width=0.3pt, opacity=0.35}
		]

		\node[draw, rounded corners, minimum width=11cm, minimum height=0.9cm, fill=gray!8]
		(P) at (0, 4.0) {Super-population $(\Omega, \mathcal{F}, P)$:\ \ $\{(Y_i, X_i, Z_i)\}_{i=1}^{\infty} \stackrel{\text{i.i.d.}}{\sim} Q$};

		\node[unit] (u1) at (-5.0, 1.8) {$1$};
		\node[unit] (u2) at (-4.0, 1.8) {$2$};
		\node[unit] (u3) at (-3.0, 1.8) {$3$};
		\node             at (-2.0, 1.8) {$\cdots$};
		\node[unit] (uN) at (-1.0, 1.8) {$N$};

		\draw[dashed, thick, betterred] (-5.5, 1.35) rectangle (-0.5, 2.25);
		\node[anchor=west, betterred, align=left] at (-0.3, 1.8)
		{Finite population\\ $D_N(\omega) = \{(Y_i(\omega), X_i(\omega), Z_i(\omega))\}_{i=1}^{N}$};

		\draw[->, thick] (-3.0, 3.55) -- (-3.0, 2.25);
		\node[anchor=west, align=left] at (-2.85, 2.90)
		{\textbf{Model randomness}\\ realize $\omega \in \Omega$, $D_{\infty}(\omega) = \{(Y_i(\omega), X_i(\omega), Z_i(\omega))\}_{i=1}^{\infty}$};

		\node[selected] (s1) at (-5.0, -0.4) {$1$};
		\node[faded]         at (-4.0, -0.4) {$2$};
		\node[selected] (s3) at (-3.0, -0.4) {$3$};
		\node                at (-2.0, -0.4) {$\cdots$};
		\node[faded]         at (-1.0, -0.4) {$N$};

		\draw[dashed, thick, betterblue] (-5.5, -0.85) rectangle (-0.5, 0.05);
		\node[anchor=west, betterblue, align=left] at (-0.3, -0.4)
		{Random sample $\*s \in \mathcal{S}_N$,\ \ $\sum_i s_i = n$};

		\draw[->, thick] (-3.0, 1.35) -- (-3.0, 0.05);
		\node[anchor=west, align=left] at (-2.85, 0.70)
		{\textbf{Design randomness}\\ draw $\*s \sim p_N(\cdot,\, Z^N(\omega))$};

	\end{tikzpicture}
	\caption{Two sources of randomness in the joint design--model framework formalized in~\cref{def:rb_product_space}. A realization $\omega \in \Omega$ determines an i.i.d.\ super-population sequence; truncating to the first $N$ elements yields the finite population $D_N(\omega)$ (red box). Conditional on this finite population, the sampling design $p_N$ draws a random sample $\*s$ of size $n$ (blue box; selected units shown filled), where the selection may depend on the auxiliary information $Z^N(\omega)$.}
	\label{fig:two-stage-randomness}
\end{figure}

We operate in the framework of~\citet{rubin-bleuer2005-b9}, and recall some definitions from their framework here, which is also illustrated in~\cref{fig:two-stage-randomness}.

\begin{restatable}[Super-Population and Finite Populations]{definition}{DefSuperpopulation}\label{def:rb_superpopulation}
	A \textit{super-population} is a probability space $(\Omega, \mathcal{F}, P)$ together with an i.i.d.\ sequence of random vectors $\{(Y_i, X_i, Z_i)\}_{i=1}^{\infty}$ defined on it, with a common distribution $Q$ on $\mathcal{Y} \times \mathcal{X} \times \mathcal{Z}$.  For each $N \geq 1$, the $N$th finite population associated with a realization $\omega \in \Omega$ is the tuple $D_N(\omega) \coloneqq \{(Y_i(\omega), X_i(\omega), Z_i(\omega))\}_{i=1}^{N}$.\footnote{Note that our definition here diverges slightly from the one used in~\citet{rubin-bleuer2005-b9}, in that we (a) directly incorporate the fact that the random vectors are i.i.d., and (b) explicitly define a common $\Omega$ and $\sigma$-algebra that links together different finite populations associated with the same super-population. The latter is implicitly needed to formalize assumptions stated in~\citet{rubin-bleuer2005-b9}, such as $P$-almost sure convergence of random variables.}
\end{restatable}

\begin{restatable}[Sampling Design]{definition}{DefSamplingDesign}\label{def:rb_sampling_design}
	Let $\mathcal{S}_N$ denote the set of all possible samples under a given sampling scheme with a finite population of size $N$, and $\mathcal{C}(\mathcal{S}_N)$ its power set. A \textit{sampling design} is a function $p_N : \mathcal{C}(\mathcal{S}_N) \times \mathcal{Z}^N \to [0, 1]$ such that for each $z^N \in \mathcal{Z}^N$, $p_N(\cdot, z^N)$ is a probability measure on $\mathcal{C}(\mathcal{S}_N)$, and for each $\*s \in \mathcal{S}_N$, $p_N(\{\*s\}, \cdot)$ is measurable; we abbreviate $p_N(\{\*s\}, \cdot)$ as $p_N(\*s, \cdot)$. We refer to $(\mathcal{S}_N, \mathcal{C}(\mathcal{S}_N), p_N)$ as the design probability space at population size $N$.
\end{restatable}

\begin{restatable}[Product Space]{definition}{DefProductSpace}\label{def:rb_product_space}
	The \textit{product space at population size $N$} combines the design space (\cref{def:rb_sampling_design}) with the super-population (\cref{def:rb_superpopulation}):
	\begin{equation*}
		\big(\mathcal{S}_N \times \Omega,\ \mathcal{C}(\mathcal{S}_N) \otimes \mathcal{F},\ P_{d, m}^{(N)}\big),
	\end{equation*}
	where the joint measure $P_{d, m}^{(N)}$ is determined on rectangles by
	\begin{equation*}
		P_{d, m}^{(N)}\big(\{\*s\} \times F\big) = \int_F p_N\big(\*s, Z^N(\omega)\big) \, dP(\omega), \qquad \*s \in \mathcal{S}_N,\ F \in \mathcal{F}.
	\end{equation*}
	We write $\E_d, \V_d$ for expectations and variances under the design (i.e., conditional on $Y^N, X^N, Z^N$) and $\E_m, \V_m$ for those under the super-population.
\end{restatable}

\begin{restatable}[Sample Estimator]{definition}{DefSampleEstimator}\label{def:rb_sample_estimator}
	A \textit{sample estimator} is a random variable $\hat{\theta}_N$ on the product space (\cref{def:rb_product_space}) of the form
	\begin{equation*}
		\hat{\theta}_N(\*s, \omega) = \hat{\theta}_N\big(\*s,\ Y^N(\omega),\ X^N(\omega),\ Z^N(\omega)\big), \qquad \*s \in \mathcal{S}_N,\ \omega \in \Omega.
	\end{equation*}
	Its randomness has two sources: the realization $\omega$ that determines the finite population, and the sample $\*s$ drawn under the design $p_N(\cdot, Z^N(\omega))$.
\end{restatable}

\subsection{Problem Setting and Regularity Assumptions}

We focus on the general setting of \enquote{prediction-powered inference} via estimating equations.  Convex M-estimation, considered in~\citet{zrnic2024asi}, is a special case.

\begin{definition}[Prediction-Powered Estimating Equations]\label{def:estimating_equations}
	For the remainder, we consider the following setup, where $\theta_0 \in \mathbb{R}^{\ell}$ is a parameter we would like to estimate, $\g : \mathcal{X} \times \mathcal{Y} \to \R^{\ell}$ is a function that is continuously differentiable with respect to $\theta$, $f$ is a fixed measurable function, and where the finite population of size $N$ consists of samples $(X_i, Y_i)$ drawn i.i.d.\ from $Q$, and other terms are defined as in~\cref{def:rb_superpopulation,def:rb_sampling_design,def:rb_product_space,def:rb_sample_estimator}:
	\begin{align*}
		G(\theta)         & := \E_m[\gY],                                                              \\
		G_N(\theta)       & := \frac{1}{N} \sum_{i=1}^{N} \giY, \text{ and }                           \\
		\hat{G}_N(\theta) & := \frac{1}{N} \sum_{i=1}^{N} \frac{s_i}{\pi_i^{(N)}} (\giY - \giF) + \giF
	\end{align*}
	where $\*s$ is drawn from either a Poisson or a Conditional Poisson Sampling design with inclusion probabilities $\*\pi^{(N)}(Z^{N}(\omega))$.  We further define $\hat{\theta}_N$ as the solution to $\hat{G}_N(\theta) = 0$, $\theta_N$ as the solution to $G_N(\theta) = 0$, and $\theta_0$ as the solution to $G(\theta) = 0$, where all solutions are presumed unique.
\end{definition}

We now make some assumptions, which should be read in the context of the definitions in~\cref{app:superpopulation_defs}. We collect them so as to distinguish between conditions on the \textit{model} (e.g., $G_N$, $G$, and the finite population of size $N$), and conditions on the \textit{design}.

\begin{assumption}[Model Assumptions]\label{asmp:model_assumptions}
	In what follows, let $B(\theta_0)$ be a compact neighborhood of the population parameter $\theta_0$.
	\crefalias{enumi}{assumption} %
	\begin{enumerate}[label=\theassumption(\alph*), ref=\theassumption(\alph*)]
		\item (Uniqueness) \label{asmp:model_unique_theta0} $G(\theta) = 0$ has a unique solution denoted by $\theta_0$, and $G_N(\theta) = 0$ has a unique solution denoted by $\theta_N$, $P$-almost surely.
		\item (Invertible Jacobian) \label{asmp:nonsingular_jacobian} The Jacobian $J(\theta_0) := \nabla G(\theta_0) \in \mathbb{R}^{\ell \times \ell}$ exists and is invertible.
		\item (Locally Lipschitz $\g$) \label{asmp:score_lipschitz} On $B(\theta_0)$, $\|\gy - \gy[\theta']\| \le M(x,y)\,\|\theta-\theta'\|$ with $\E_m[M(X,Y)^4] < \infty$ and $\E_m[M(X, f(X))^4] < \infty$.
		\item (Local Differentiability of $\g$) \label{asmp:score_C1} For each $(x,y)\in\mathcal X\times\mathcal Y$, the map $\theta \mapsto \gy$ is continuously differentiable on $B(\theta_0)$, with Jacobian denoted $\nabla g_{\theta}(x,y)$.
		\item (Locally Lipschitz $\h$) \label{asmp:jacobian_lipschitz} On $B(\theta_0)$, $\|\nabla g_{\theta}(x,y) - \nabla g_{\theta'}(x,y)\| \le M_J(x,y)\,\|\theta-\theta'\|$ with $\E_m[M_J(X,Y)] < \infty$ and $\E_m[M_J(X, f(X))] < \infty$.
		\item (Bounded Fourth Moments at $\theta_0$) \label{asmp:score_integrability_theta0} $\E_m\|\gY[\theta_0]\|^4 < \infty$ and $\E_m\|\gF[\theta_0]\|^4 < \infty$.
		\item (Consistency of $\theta_N$) \label{asmp:z_estimator_consistency} $\theta_N \xrightarrow{P\text{-a.s.}} \theta_0$ as $N \to \infty$, where $\theta_N$ is the solution to $G_N(\theta) = 0$ from~\cref{asmp:model_unique_theta0}.
	\end{enumerate}
\end{assumption}

We note that these assumptions are slightly stronger than the minimal conditions required for standard Z-estimation (cf.\ \citet{vanderVaart2000} Theorem 5.21, \citet{newey1994large} Theorem 3.4).  For instance, we require bounded fourth rather than second moments. These conditions are used to verify conditions of Theorem 6.1 of \citet{rubin-bleuer2005-b9} that require reasoning about uniform convergence over both the model and design probability space. 
The consistency assumption (\cref{asmp:z_estimator_consistency}) holds under standard Z-estimator theory (e.g.,~\citet[Theorem 5.9]{vanderVaart2000}) given~\cref{asmp:model_unique_theta0} (well-separated zero of $G$) together with uniform convergence $\sup_{\theta \in B(\theta_0)} \|G_N(\theta) - G(\theta)\| \xrightarrow{P\text{-a.s.}} 0$. %
We state it directly as an assumption to keep the proof chain modular.

We collect the following consequences of~\cref{asmp:model_assumptions}, which we will cite repeatedly in the sequel.

\begin{corollary}[Consequences of Model Assumptions]\label{cor:jacobian_representation}
	Under~\cref{asmp:model_assumptions}, the following hold:
	\crefalias{enumi}{corollary} %
	\begin{enumerate}[label=\thecorollary(\roman*), ref=\thecorollary(\roman*)]
		\item \emph{(Score Bound)}\label{cor:score_envelope} For each $(x,y)$, $\sup_{\theta \in B(\theta_0)} \|\gy\| \le \|\gy[\theta_0]\| + M(x,y)\,\mathrm{diam}(B(\theta_0))$ (and analogously for $\gf$). In particular, $\E_m[\sup_{\theta \in B(\theta_0)}\|\gY\|^4] < \infty$ (and analogously for $\gF$).
		\item \emph{(Jacobian Bound)}\label{cor:jac_bound} For each $(x,y)$, $\sup_{\theta \in B(\theta_0)} \|\nabla g_{\theta}(x,y)\| \le M(x,y)$, and $\sup_{\theta \in B(\theta_0)} \|\nabla g_{\theta}(x,f(x))\| \le M(x, f(x))$. In particular, $\E_m[\sup_{\theta \in B(\theta_0)}\|\hy\|^4] < \infty$ (and analogously for $\hf$).
		\item \emph{(Jacobian Representation)}\label{cor:jac_rep} The map $\theta \mapsto J(\theta) := \nabla G(\theta)$ is well-defined on $B(\theta_0)$ with $J(\theta) = \E_m[\hy]$.
		\item \emph{(Lipschitz Continuity of $J$)}\label{cor:jac_continuity} $J$ is Lipschitz (hence continuous) on $B(\theta_0)$, satisfying $\|J(\theta) - J(\theta')\| \le \E_m[M_J(X,Y)]\,\|\theta - \theta'\|$.
	\end{enumerate}
\end{corollary}
\begin{proof}
	\textit{(i)} On $B(\theta_0)$, the triangle inequality and~\cref{asmp:score_lipschitz} give $\|\gy\| \le \|\gy[\theta_0]\| + \|\gy - \gy[\theta_0]\| \le \|\gy[\theta_0]\| + M(x,y)\,\|\theta - \theta_0\|$, and $\|\theta - \theta_0\| \le \mathrm{diam}(B(\theta_0))$ since $\theta, \theta_0 \in B(\theta_0)$. The fourth-moment bound follows by combining with~\cref{asmp:score_lipschitz,asmp:score_integrability_theta0}.

	\textit{(ii)} Letting $\theta' \to \theta$ along any unit direction in~\cref{asmp:score_lipschitz} gives $\|\nabla g_{\theta}(x,y)\| \le M(x,y)$ on $B(\theta_0)$ (and analogously $\|\nabla g_{\theta}(x,f(x))\| \le M(x, f(x))$); the fourth-moment bounds then follow from~\cref{asmp:score_lipschitz}.

	\textit{(iii)} \Cref{asmp:score_C1,asmp:score_lipschitz} permit the interchange $\nabla G(\theta) = \E_m[\hy]$ via dominated convergence: the difference quotient $\bigl(\gy[\theta+he] - \gy\bigr)/h$ is bounded in norm by $M(x,y)$, which is integrable under~\cref{asmp:score_lipschitz}.

	\textit{(iv)} By~\cref{asmp:jacobian_lipschitz}, $\|J(\theta) - J(\theta')\| \le \E_m\|\hy - \hy[\theta']\| \le \E_m[M_J(X,Y)]\,\|\theta - \theta'\|$.
\end{proof}

\begin{assumption}[Design Assumptions]\label{asmp:design_assumptions}
	We make the following assumptions on the design.
	\crefalias{enumi}{assumption} %
	\begin{enumerate}[label=\theassumption(\alph*), ref=\theassumption(\alph*)]
		\item \label{asmp:design_assumption_constraints_on_pi} The finite-population inclusion probabilities $\*\pi^{(N)}$ are determined by the design as a measurable function of $Z^{N}$.  In this paper the design uses only the covariates, so we take $Z_i = X_i$. Let $n = n_N \in \3N$ be a deterministic budget sequence with $n_N / N \to r$ as $N \to \infty$, for a constant $r \in (0, 1)$ (e.g., $n_N = \lfloor rN \rfloor$). We assume that, with $P$-probability one, there exists $N_0$ (possibly depending on $\omega$) such that for all $N \geq N_0$, $\*\pi^{(N)}$ satisfies (i) $\sum_{i=1}^N \pi_i^{(N)} = n_N$ and (ii) $\gamma \leq \pi_i^{(N)} \leq 1 - \gamma$ for all $i \in \{1, \ldots, N\}$, with a deterministic constant $0 < \gamma < 1/2$.
		\item \label{asmp:design_assumption_pi_lim} There exists a measurable $\pis: \mathcal{X} \to [\gamma, 1-\gamma]$ satisfying $\E_m[\pis(X)] = r$, and $\frac{1}{N}\sum_{i=1}^N \bigl|\pi^{(N)}_i - \pis(X_i)\bigr|^2 \;\xrightarrow{P\text{-a.s.}}\; 0.$
	\end{enumerate}
\end{assumption}

\subsection{Proof Strategy}\label{app:proof_strategy}

We restate Theorem 6.1 of~\citet{rubin-bleuer2005-b9} below, and our proof will consist primarily of verifying that the conditions hold under our assumptions.  By applying a common framework to analyze both Poisson and Conditional Poisson Sampling, we will illustrate that the difference between the two is driven entirely by the difference in their design variances.

\begin{theorem}[Theorem 6.1 of \citet{rubin-bleuer2005-b9}]\label{thm:rb_main}
	Consider a sequence of super-populations of $N$ independent random vectors associated with design spaces, with sample sizes $n = n_N \to \infty$, and let $r := \lim_{N \to \infty} n / N \geq 0$. Suppose:
	\begin{enumerate}
		\item $G_N(\theta_0) \to 0$ with $P$-probability one.
		\item There is a compact neighborhood $B(\theta_0)$ on which, with $P$-probability one, all $G_N$ are continuously differentiable, and the Jacobians $\partial G_N(\theta) / \partial \theta$ converge uniformly in $\theta \in B(\theta_0)$ to a non-stochastic (in the model) limit $J(\theta)$ which is non-singular at $\theta_0$.
		\item $\sqrt{N}\, G_N(\theta_0) \cid \mathcal{N}(0, \Sigma_m)$ in the law of the super-population.
		\item There is a compact neighborhood $B(\theta_0)$ on which $\partial \hat{G}_N(\theta) / \partial \theta$ converges uniformly in design probability to a non-stochastic (in the design) limit which coincides with $J(\theta)$ at $\theta_0$, for $P$-almost every $\omega$.
		\item $\sqrt{n}\, \hat{G}_N(\theta_N) \cid \mathcal{N}(0, \Sigma_d)$ in the law of the design as $n \to \infty$, for $P$-almost every $\omega$, where $\Sigma_d$ is non-stochastic in the super-population.
	\end{enumerate}
	Then, setting $J := J(\theta_0)$ and $\Sigma := J^{-1} (\Sigma_d + r\, \Sigma_m) J^{-\top}$, we have, in the law of the product space (\cref{def:rb_product_space}),
	\begin{equation*}
		\sqrt{n}\,(\hat{\theta}_N - \theta_0) \cid \mathcal{N}(0, \Sigma).
	\end{equation*}
\end{theorem}

The \textit{model-based} conditions (1)--(3) are straightforward to verify under our assumptions using standard arguments, and we do so below as a starting point.

\begin{lemma}[Verification of Conditions (1)--(3) of \cref{thm:rb_main}]\label{lemma:model_condition_verification}
	For $G(\theta), G_N(\theta), \hat{G}_N(\theta), \theta_0, \theta_N, \hat{\theta}_N$ (see~\cref{def:estimating_equations}) defined on the design--model product space (\cref{def:rb_product_space}), under~\cref{asmp:model_assumptions} and~\cref{asmp:design_assumptions}, where $J := J(\theta_0)$, we have that
	\begin{enumerate}
		\item $G_N(\theta_0) \to 0$ with $P$-probability one.
		\item There is a compact neighborhood $B(\theta_0)$ on which, with $P$-probability one, all $G_N$ are continuously differentiable, and the Jacobians $\partial G_N(\theta) / \partial \theta$ converge uniformly in $\theta \in B(\theta_0)$ to a non-stochastic (in the model) limit $J(\theta)$ which is non-singular at $\theta_0$.
		\item $\sqrt{N}\, G_N(\theta_0) \cid \mathcal{N}(0, \Sigma_m)$ in the law of the super-population.
	\end{enumerate}
	where $\Sigma_m = \E_m[\gY[\theta_0] \gY[\theta_0]^\top]$.
\end{lemma}
\begin{proof}
	\textbf{Condition (1).} $G_N(\theta_0) \rightarrow 0$ with $P$-probability one (i.e., converges almost surely to zero) holds via the Strong Law of Large Numbers, since it is an empirical mean of random variables $\giY[\theta_0]$ that are mean-zero.

	\textbf{Condition (2).} Continuous differentiability holds because, by~\cref{asmp:score_C1}, for every $(x,y)$ the map $\theta \mapsto \gy$ is continuously differentiable on $B(\theta_0)$ with Jacobian $\nabla g_{\theta}(x,y)$, and since $G_N(\theta) = \frac{1}{N}\sum_{i=1}^N \giY$ is a finite sum, it is continuously differentiable on $B(\theta_0)$ with $\nabla G_N(\theta) = \frac{1}{N}\sum_{i=1}^N \hiy$, $P$-almost surely. By~\cref{cor:jacobian_representation}, $J(\theta) := \nabla G(\theta) = \E_m[\hy]$ is well-defined and continuous on $B(\theta_0)$, and is in particular non-stochastic in the model, and by~\cref{asmp:nonsingular_jacobian}, $J(\theta_0) = \nabla G(\theta_0)$ is invertible.  Uniform convergence then follows from the uniform SLLN of~\citet[Theorem 2]{jennrich1969-c8}, which requires (a) continuity of $\nabla g_{\theta}(x,y)$ in $\theta$ for each $(x,y)$, and (b) an integrable envelope $\E_m[\sup_{\theta \in B(\theta_0)} \|\hy\|] < \infty$. The first condition holds by~\cref{asmp:score_C1}, and the second holds via the Jacobian bound $\sup_{\theta \in B(\theta_0)} \|\nabla g_{\theta}(x,y)\| \le M(x,y)$ from~\cref{cor:jac_bound} together with $\E_m[M(X,Y)] < \infty$ by~\cref{asmp:score_lipschitz}. Hence, $\sup_{\theta \in B(\theta_0)}\|\nabla G_N(\theta) - J(\theta)\| \xrightarrow{P\text{-a.s.}} 0$.

	\textbf{Condition (3).} $\sqrt{N}\, G_N(\theta_0) \cid \mathcal{N}(0, \Sigma_m)$ in the law of the super-population holds via the standard central limit theorem, with $\Sigma_m = \E_m[\gY[\theta_0] \gY[\theta_0]^\top]$ existing because $\E_m\|\gY[\theta_0]\|^2 \le \sqrt{\E_m\|\gY[\theta_0]\|^4} < \infty$ by~\cref{asmp:score_integrability_theta0}.
\end{proof}

The more challenging component is to verify the \textit{design-based} conditions (conditions (4)--(5)).  The remainder of this section is structured as follows. In~\cref{app:ulln} we establish some results regarding a design-based uniform law of large numbers, which will be useful for establishing condition (4). In~\cref{app:design_variance_limiting} we establish design-based asymptotic normality and consistency of scalar parameters (\cref{lemma:ht_design}), which we then extend to multivariate parameters (\cref{lemma:ht_design_vector}), and in~\cref{lemma:convergence_of_covariance} we establish convergence of the design-based variance to a limiting variance.  In~\cref{app:main_result} we put these components together to yield a proof of the following statement, which is our main result.

\begin{restatable}[Formal Statement of Asymptotic Normality]{theorem}{MainResult}\label{thm:main_result}
	For $G(\theta), G_N(\theta), \hat{G}_N(\theta), \theta_0, \theta_N, \hat{\theta}_N$ (see~\cref{def:estimating_equations}) defined on the design--model product space (see~\cref{def:rb_product_space}), under~\cref{asmp:model_assumptions} and~\cref{asmp:design_assumptions}, where $J := J(\theta_0)$, and assuming that $\Sigma_{\infty,\bullet}$ (defined below) is positive definite,
	\begin{equation}
		\sqrt{n} (\hat{\theta}_N - \theta_0) \cid \mathcal{N}(0, \Sigma), \text{ where } \Sigma = J^{-1}(\Sigma_d + r \cdot \Sigma_{m}) J^{-\top}
	\end{equation}
	where $\Sigma_m = \E_m[\gY[\theta_0] \gY[\theta_0]^\top]$, and where $\Sigma_d = r\,\Sigma_{\infty, \bullet}$ for $\bullet \in \{\textup{ps, cps}\}$ (with $r = \lim_{N \to \infty} n_N/N \in (0,1)$ defined in~\cref{asmp:design_assumption_constraints_on_pi}), where $w^\star := \gY[\theta_0] - \gF[\theta_0]$ and we write $\pis$ for $\pis(X)$, with $\pis$ from~\cref{asmp:design_assumption_pi_lim}.
	\begin{align*}
		\Sigma_{\infty,\textup{ps}}
		      & := \E_m\left[w^\star w^{\star\top}(1-\pis) / \pis\right],                                                       \\
		\Sigma_{\infty,\textup{cps}}
		      & := \E_m\!\left[\bigl(w^\star-\pis \beta\bigr)\bigl(w^\star-\pis \beta\bigr)^\top\,\frac{1-\pis}{\pis}\right], &
		\beta & := \frac{\E_m\bigl[w^\star\bigl(1-\pis\bigr)\bigr]}{\E_m\bigl[\pis\bigl(1-\pis\bigr)\bigr]}
	\end{align*}
\end{restatable}

\subsection{Design Lemmas}\label{app:design_lemmas}

\subsubsection{Uniform Laws of Large Numbers}
\label{app:ulln}

For Condition~(4) of~\cref{thm:rb_main}, we need uniform design consistency of the Jacobian of $\hat G_N(\theta)$ on a compact neighborhood of $\theta_0$. This follows from a general design ULLN: pointwise design consistency at each $\theta$, combined with a model-controlled Lipschitz envelope, yields uniform consistency by a standard finite-net (compactness) argument.

\begin{lemma}[Design ULLN]\label{lemma:design_ulln}
	Fix $\omega \in \Omega$ such that~\cref{asmp:design_assumption_constraints_on_pi} holds at $\omega$, and let $\*\pi^{(N)} = \pi^{(N)}(Z^N(\omega))$ denote the resulting inclusion probabilities at population size $N$. Let $\Theta \subset \R^\ell$ be compact (in our application, $\Theta = B(\theta_0)$, a compact neighborhood of $\theta_0$), and let $\{g_\theta : \theta \in \Theta\}$ be a class of measurable functions on $\mathcal X \times \mathcal Y$ taking values in $\R^\ell$. For each $\theta \in \Theta$, define
	\begin{equation*}
		\hat G_N(\theta) \;:=\; \frac{1}{N}\sum_{i=1}^N \frac{s_i}{\pi^{(N)}_i}\, g_\theta(X_i,Y_i),
		\qquad
		G_N(\theta) \;:=\; \frac{1}{N}\sum_{i=1}^N g_\theta(X_i,Y_i).
	\end{equation*}
	Suppose, on a model-probability-one set of $\omega$:
	\begin{enumerate}
		\item[\textup{(a)}] \emph{Pointwise Design Consistency.} For each $\theta\in\Theta$, $\hat G_N(\theta)-G_N(\theta)\cip 0$ in design probability.
		\item[\textup{(b)}] \emph{Lipschitz Envelope.} There exists a measurable
			$M:\mathcal X\times\mathcal Y\to[0,\infty)$ with
			\begin{equation*}
				\|g_\theta(x,y)-g_{\theta'}(x,y)\| \;\le\; M(x,y)\,\|\theta-\theta'\|
				\qquad \forall\,\theta,\theta'\in\Theta,
			\end{equation*}
			and $\bar M_N := N^{-1}\sum_{i=1}^N M(X_i,Y_i)$ is bounded along $P$-a.e.\ sample path; that is, $\limsup_N \bar M_N < \infty$, $P$-a.s. (A sufficient condition is $\E_m[M(X,Y)] < \infty$, in which case the SLLN gives $\bar M_N \to \E_m[M(X,Y)]$ $P$-a.s.)
	\end{enumerate}
	Then for $P$-almost every $\omega$,
	\begin{equation*}
		\sup_{\theta\in\Theta}\bigl\|\hat G_N(\theta)-G_N(\theta)\bigr\| \;\cip\;0 \qquad \text{in design probability.}
	\end{equation*}
\end{lemma}

\begin{proof}
	Fix $\omega$ in the model-probability-one set on which (a) and (b) hold and $\bar M_N$ is bounded; in what follows, all probabilities are with respect to the design conditional on $\omega$. Let $\epsilon>0$ and choose $\delta>0$ to be specified. By compactness, $\Theta$ admits a finite $\delta$-net $\{\theta^{(1)},\ldots,\theta^{(K_\delta)}\}\subset\Theta$. For arbitrary $\theta\in\Theta$, pick $\theta^{(k)}$ with $\|\theta-\theta^{(k)}\|\le\delta$. The triangle inequality applied to the finite sum, followed by the Lipschitz bound (b), gives
	\begin{align*}
		\bigl\|\hat G_N(\theta)-\hat G_N(\theta^{(k)})\bigr\|
		 & \;=\; \biggl\|\frac{1}{N}\sum_i \frac{s_i}{\pi^{(N)}_i}\bigl[g_\theta(X_i,Y_i) - g_{\theta^{(k)}}(X_i,Y_i)\bigr]\biggr\| \\
		 & \;\le\; \frac{1}{N}\sum_i \frac{s_i}{\pi^{(N)}_i}\,\bigl\|g_\theta(X_i,Y_i) - g_{\theta^{(k)}}(X_i,Y_i)\bigr\|           \\
		 & \;\le\; \frac{1}{N}\sum_i \frac{s_i}{\pi^{(N)}_i}\,M(X_i,Y_i)\,\|\theta-\theta^{(k)}\|.
	\end{align*}
	Since $s_i \in \{0,1\}$ and $\pi^{(N)}_i \ge \gamma$ by~\cref{asmp:design_assumption_constraints_on_pi}, we have $s_i / \pi^{(N)}_i \le 1/\gamma$; combined with $\|\theta - \theta^{(k)}\| \le \delta$, this gives
	\begin{equation*}
		\bigl\|\hat G_N(\theta)-\hat G_N(\theta^{(k)})\bigr\|
		\;\le\; \frac{\delta}{\gamma}\cdot\frac{1}{N}\sum_i M(X_i, Y_i) \;=\; \frac{\delta}{\gamma}\,\bar M_N.
	\end{equation*}
	The same triangle-and-Lipschitz argument, without the design-weighting factor (so without the $1/\gamma$), gives
	\begin{equation*}
		\bigl\|G_N(\theta)-G_N(\theta^{(k)})\bigr\| \;\le\; \frac{1}{N}\sum_i M(X_i, Y_i)\,\|\theta-\theta^{(k)}\| \;\le\; \delta\,\bar M_N.
	\end{equation*}
	Combining via the triangle inequality $\|\hat G_N(\theta) - G_N(\theta)\| \le \|\hat G_N(\theta) - \hat G_N(\theta^{(k)})\| + \|\hat G_N(\theta^{(k)}) - G_N(\theta^{(k)})\| + \|G_N(\theta^{(k)}) - G_N(\theta)\|$ and taking the supremum over $\theta\in\Theta$,
	\begin{equation*}
		\sup_{\theta\in\Theta}\bigl\|\hat G_N(\theta)-G_N(\theta)\bigr\|
		\;\le\; \max_{1\le k\le K_\delta}\bigl\|\hat G_N(\theta^{(k)})-G_N(\theta^{(k)})\bigr\|
		\;+\; \delta\bigl(\gamma^{-1}+1\bigr)\bar M_N.
	\end{equation*}
	Choose $\delta = \delta(\epsilon,\omega)$ small enough that $\delta(\gamma^{-1}+1)\bar M_N(\omega) \le \epsilon/2$ for all $N$ large; this is possible since $\bar M_N$ is bounded along the sample path under (b). The first term is a maximum over finitely many points, each of which is $o_P(1)$ in design probability by (a), so it is $o_P(1)$. Hence $P_d\bigl(\sup_\theta\|\hat G_N(\theta)-G_N(\theta)\| > \epsilon\bigr) \to 0$, which is the claim.
\end{proof}

\subsubsection{Design Variance and Limiting Variance}
\label{app:design_variance_limiting}

We first state and prove~\cref{lemma:ht_design} for the scalar setting, which is the same as \cref{lemma:ht_design_informal} but with regularity conditions and measure-theoretic notation included.  We then show that this can be generalized straightforwardly to the multivariate vector setting via Cram\'er--Wold in~\cref{lemma:ht_design_vector}.  For CPS, the first is essentially a restatement of Theorem 1 of~\citet{Berger1998norm}, and the second is a fairly standard application of Cram\'er--Wold.

We note that more general results along these lines exist in the literature for functional CLTs in the CPS and PS designs \citep{bertail2017-a7,boistard2017-68}.  However, since our setting is simpler (we consider a finite-dimensional $\theta$, not a function class), we present the following lemmas to keep the presentation self-contained.  The more technically challenging and original part of our proof will come in~\cref{app:convergence_design_covariance}, where we establish convergence to the claimed design covariance.

\begin{lemma}[Design Consistency and Asymptotic Normality of HT Under (C)PS]\label{lemma:ht_design}
	Fix $\omega \in \Omega$ such that~\cref{asmp:design_assumption_constraints_on_pi} holds at $\omega$, and consider the resulting fixed sequence $\{(Y_i(\omega), X_i(\omega), Z_i(\omega))\}_{i=1}^\infty$.
	Let $w : \mathcal{Y} \times \mathcal{X} \to \R$ be measurable, write $w_i \coloneqq w(Y_i(\omega), X_i(\omega))$, and assume $\limsup_{N \to \infty} \frac{1}{N} \sum_{i=1}^N w_i^4 < \infty$. 
	Let $\hat{\mu}^w_{\*\pi}$ denote the Horvitz--Thompson (HT) estimator of $\mu^w_N \coloneqq \frac{1}{N} \sum_{i=1}^N w_i$, formed under either the Poisson design $\p_{\textup{ps}(\*\pi)}$ or the Conditional Poisson design $\p_{\textup{cps}(\*\pi)}$. Then, the following convergence statements hold in design probability $p_N(\cdot, Z^{N}(\omega))$:
	\begin{enumerate}
		\item \textup{(Consistency.)} $\hat{\mu}^w_{\*\pi} - \mu^w_N \cip 0$ as $N \to \infty$.
		\item \textup{(Asymptotic Normality.)} If $\sigma^2_\bullet(\*\pi, N) > 0$ is bounded away from zero,
		      \begin{equation*}
			      \frac{\sqrt{N}}{\sigma_\bullet(\*\pi, N)} \big(\hat{\mu}^w_{\*\pi} - \mu^w_N\big) \cid \mathcal{N}(0, 1) \quad \text{as } N \to \infty,
		      \end{equation*}
		      where the design-specific scaling $\sigma_\bullet(\*\pi, N)$ is given by
		      \begin{align*}
			      \sigma^2_{\textup{ps}}(\*\pi, N)  & = \frac{1}{N} \sum_{i=1}^N w_i^2 \, \frac{1 - \pi_i}{\pi_i},                                                                                             \\
			      \sigma^2_{\textup{cps}}(\*\pi, N) & = \frac{1}{N} \sum_{i=1}^N z_i^2 \, \frac{1 - \pi_i}{\pi_i}, 
				  \quad z_i \coloneqq w_i - \pi_i \beta, \quad \beta \coloneqq \frac{\sum_j w_j (1 - \pi_j)}{\sum_k \pi_k (1 - \pi_k)}.
		      \end{align*}
	\end{enumerate}
\end{lemma}

\begin{proof}
	We show this result for Poisson Sampling and Conditional Poisson Sampling separately.  Recall that $\hat{\mu}^w_{\*\pi} \coloneqq \frac{1}{N} \sum_{i=1}^N \frac{s_i w_i}{\pi_i} $ and $\mu^w_N := \frac{1}{N} \sum_{i=1}^{N} w_i$.

	\textbf{Poisson Sampling}: Writing
	\begin{equation*}
		\hat{\mu}^w_{\*\pi} - \mu^w_N \;=\; \frac{1}{N} \sum_{i=1}^{N} T_{N,i}, \qquad T_{N,i} \;:=\; \frac{w_i (s_i - \pi_i)}{\pi_i},
	\end{equation*}
	the $\{T_{N,i}\}$ are mean-zero because $\E_d[s_i] = \pi_i$, independent because this is Poisson Sampling, and not identically distributed because the coins are different. 
	The variance at unit $i$ is $\V_d(T_{N,i}) = w_i^2(1-\pi_i)/\pi_i$, so by independence and using $\pi_i \ge \gamma$ from~\cref{asmp:design_assumption_constraints_on_pi} we get
	\begin{align*}
		\V_d(\hat\mu^w_{\*\pi} - \mu^w_N)
		\;=\; 
		\frac{1}{N^2}\sum_{i=1}^N w_i^2\,\frac{1-\pi_i}{\pi_i} 
		\;\le\; 
		\frac{1-\gamma}{\gamma} \frac{1}{N^2}\sum_{i=1}^N w_i^2
		\;\le\; 
		\frac{1}{\gamma} \frac{1}{N^2}\sum_{i=1}^N w_i^2.
	\end{align*}

	By Jensen's inequality, $\frac{1}{N}\sum_i w_i^2 \le \bigl(\frac{1}{N}\sum_i w_i^4\bigr)^{1/2} = O(1)$ for this fixed $\omega$ as $N$ grows, by the assumption $\limsup_{N \to \infty} \frac{1}{N} \sum_{i=1}^N w_i^4 < \infty$. Hence the design variance above is $O(1/N) \to 0$, and Chebyshev's inequality gives consistency.

	For asymptotic normality, we verify the Lyapunov condition at $\delta = 2$, which implies the Lindeberg condition. With $|s_i - \pi_i| \le 1$ and $\pi_i \ge \gamma$,
	\begin{equation*}
		\sum_{i=1}^N \E_d\biggl|\frac{T_{N,i}}{\sqrt N}\biggr|^4 \;\le\; \frac{1}{N^2 \gamma^4}\sum_{i=1}^N w_i^4 \cdot \E_d|s_i - \pi_i|^4 \;\le\; \frac{1}{N \gamma^4} \cdot \frac{1}{N}\sum_{i=1}^N w_i^4 \;=\; O(N^{-1}) \;\to\; 0,
	\end{equation*}
	using the Bernoulli moment bound $\E_d|s_i - \pi_i|^4 \le 1$. The standardized Lyapunov ratio is $\sum_i \E_d|T_{N,i}/\sqrt N|^4 / \sigma_{\textup{ps}}^4(\*\pi,N) \to 0$ since $\sigma^2_{\textup{ps}}(\*\pi, N)$ is bounded away from zero by hypothesis, so the Lindeberg--Feller CLT applies.

	\textbf{Conditional Poisson Sampling}: The proof follows from Theorem 1 of~\citet{Berger1998norm}, which is itself a restatement of a theorem of~\citet{Hajek1964cps}, and verifying that the conditions hold in our setting.  We restate that result here, before showing how it applies in our case, while aligning the notation slightly to our setting.

	\begin{lemma}[Theorem 1 of~\citet{Berger1998norm}]\label{thm:berger}
		Consider a sequence of populations of size $N$ with outcomes $Y_i$ and corresponding inclusion probabilities $\pi_i$ and total $T = \sum_{i} Y_i$, and let $\*s$ denote the indices selected via Conditional Poisson Sampling with probabilities $\pi_i$.  Define $t_{\pi} = \sum_{i} \frac{s_i Y_i}{\pi_i}$ to be the total estimator, and let $\tilde{t}_{\pi} = (t_{\pi} - T) / \sigma$ be the normalized error, where $\sigma^2 := \sum_{i=1}^{N} Z_i^2 (1 - \pi_i) / \pi_i$, $Z_i := Y_i - \beta \pi_i$, $\beta = \frac{1}{d(\pi)} \sum_{i=1}^{N} Y_i [1 - \pi_i]$, and $d(\pi) = \sum_{i=1}^{N} \pi_i (1 - \pi_i)$. If $d(\pi) \rightarrow \infty$ and the following Lindeberg condition holds for every $\epsilon > 0$,
		\begin{equation}
			\frac{1}{\sigma^2} \sum_{i=1}^{N} Z_i^2 (1 - \pi_i) / \pi_i \mathbf{1}\{|Z_i| / \pi_i > \epsilon \sigma\} \rightarrow 0,
		\end{equation}
		then $\tilde{t}_{\pi} \cid \mathcal{N}(0, 1)$.
	\end{lemma}

	We require the above CLT because CPS has non-independent indicators $s_i$. 
	It requires two assumptions: 
	(A) $d(\pi) \to \infty$, and (B) the Lindeberg condition on the centered values.
	Recall from~\cref{lemma:ht_design} that $z_i \coloneqq w_i - \pi_i \beta$ and $\beta \coloneqq \frac{\sum_j w_j (1 - \pi_j)}{\sum_k \pi_k (1 - \pi_k)}$.
	With $Y_i$ replaced by $w_i$, Berger's $Z_i$ is our $z_i$, and $\sigma^2 = N \sigma^2_{\textup{cps}}(\*\pi, N)$.
	Since $\hat{\mu}^w_{\*\pi} = t_\pi / N$, $\mu^w_N = T / N$, and $\sigma = \sqrt{N}\sigma_{\textup{cps}}(\*\pi, N)$, Berger's theorem applies to our normalized error:
	\al{
	    & \frac{\sqrt{N}}{\sigma_{\textup{cps}}(\*\pi, N)}\bigl(\hat{\mu}^w_{\*\pi} - \mu^w_N\bigr)
	    = \frac{t_\pi - T}{\sqrt{N}\sigma_{\textup{cps}}(\*\pi, N)}
	    = \frac{t_\pi - T}{\sigma}
	    = \tilde{t}_\pi.
	}
	\cref{thm:berger} is stated in design probability with the population held fixed, which is our setting, since $\omega$ is fixed.
	We will now show that Berger's assumptions hold in our setting. 
	Throughout, let $C > \limsup_{N} \frac{1}{N}\sum_j w_j^4$ as assumed in~\cref{lemma:ht_design}, so that $\frac{1}{N}\sum_j w_j^4 \le C$ for all $N$ large enough.

	\textit{(A) holds.} Since $\gamma \le \pi_i \le 1 - \gamma$, each term of $d(\pi) = \sum_i \pi_i(1-\pi_i)$ is at least $\gamma(1-\gamma)$, so $d(\pi) \ge N\gamma(1-\gamma) \to \infty$.

	\textit{(B) holds.} We show that the Lindeberg sum is $O(1/N)$ in three parts: $\beta$ is bounded (part 1), so the $z_i$ have bounded average fourth moment (part 2), which bounds the Lindeberg sum (part 3).

	\textit{Part 1: $\beta$ is bounded.} Its numerator is $O(N)$ and its denominator is at least $N\gamma(1-\gamma)$. For all $N$ large enough,
	\flal{
		&   |\beta|
			= \frac{\bigl|\sum_j w_j (1 - \pi_j)\bigr|}{\sum_k \pi_k (1 - \pi_k)}
			\le \frac{\sum_j |w_j|}{\sum_k \pi_k (1 - \pi_k)}
			\le \frac{\sum_j |w_j|}{N \gamma (1 - \gamma)}
			= \frac{\frac{1}{N}\sum_j |w_j|}{\gamma (1 - \gamma)}
				&&\tag{$\gamma \le \pi_k \le 1 - \gamma$}
		\\& \le \frac{\bigl(\frac{1}{N}\sum_j w_j^4\bigr)^{1/4}}{\gamma (1 - \gamma)}
				&&\tag{Jensen}
		\\& \le \frac{C^{1/4}}{\gamma (1 - \gamma)}
				&&\tag{Assumption}
	}

	\textit{Part 2: the $z_i$ have bounded average fourth moment.} For all $N$ large enough,
	\flal{
		&   \frac{1}{N}\sum_i z_i^4
			= \frac{1}{N}\sum_i (w_i - \pi_i \beta)^4
			\le \frac{1}{N}\sum_i \bigl(|w_i| + |\beta|\bigr)^4
				&&\tag{$0 \le \pi_i \le 1$}
		\\& \le \frac{1}{N}\sum_i 8 \bigl(w_i^4 + \beta^4\bigr)
		= 8 \cdot \frac{1}{N}\sum_i w_i^4 + 8 \beta^4
				&&\tag{\cref{lemma:power_mean} at $p = 4$}
		\\& \le 8C + 8\beta^4
				&&\tag{Assumption}
		\\& \le 8C + \frac{8C}{\gamma^4 (1 - \gamma)^4}
			=: C_z
				&&\tag{part 1}
	}

	\textit{Part 3: the Lindeberg sum vanishes.} The only new tool is a bound on an indicator: for any $x \ge 0$ and $a > 0$, $\mathbf{1}\{x > a\} \le x^2 / a^2$, since if $x > a$ the right-hand side exceeds $1$, and otherwise it is non-negative. 
	Recall that $\sigma^2_{\textup{cps}}(\*\pi, N)$ is bounded away from zero, as assumed in~\cref{lemma:ht_design}, i.e.,~there is $c > 0$ with $\sigma^2_{\textup{cps}}(\*\pi, N) \ge c$ for all $N$ large enough. Writing the Lindeberg sum of~\cref{thm:berger} in our notation ($Z_i = z_i$), for all $N$ large enough,
	\flal{
		&   \frac{1}{\sigma^2}\sum_i z_i^2 \,\frac{1-\pi_i}{\pi_i}\,\mathbf{1}\bigl\{|z_i|/\pi_i > \epsilon\sigma\bigr\}
			&&
		\\& \le \frac{1}{\sigma^2}\sum_i z_i^2 \,\frac{1-\pi_i}{\pi_i}\cdot\frac{z_i^2}{\epsilon^2\sigma^2\pi_i^2}
				&&\tag{indicator bound with $x = |z_i|/\pi_i$, $a = \epsilon\sigma$}
		\\& = \frac{1}{\epsilon^2\sigma^4}\sum_i z_i^4 \,\frac{1-\pi_i}{\pi_i^3}
				&&
		\\& \le \frac{1}{\epsilon^2\sigma^4\gamma^3}\sum_i z_i^4
				&&\tag{$1 - \pi_i \le 1$ and $\pi_i \ge \gamma$}
		\\& = \frac{N}{\epsilon^2\sigma^4\gamma^3}\cdot\frac{1}{N}\sum_i z_i^4
				&&
		\\& = \frac{1}{\epsilon^2\gamma^3}\cdot\frac{\frac{1}{N}\sum_i z_i^4}{N\sigma^4_{\textup{cps}}(\*\pi, N)}
				&&\tag{$\sigma^2 = N\sigma^2_{\textup{cps}}(\*\pi, N)$, so $\sigma^4 = N^2\sigma^4_{\textup{cps}}(\*\pi, N)$}
		\\& \le \frac{1}{\epsilon^2\gamma^3}\cdot\frac{C_z}{N\sigma^4_{\textup{cps}}(\*\pi, N)}
				&&\tag{part 2}
		\\& \le \frac{1}{\epsilon^2\gamma^3}\cdot\frac{C_z}{N c^2}
				&&\tag{$\sigma^2_{\textup{cps}}(\*\pi, N) \ge c$}
		\\& \to 0
				&&\tag{only $N$ varies}
	}

	Thus (A) and (B) hold, so~\cref{thm:berger} applies.

	\textit{Consistency under CPS.} \Cref{thm:berger} gives consistency only when $\sigma^2_{\textup{cps}}(\*\pi, N)$ is bounded away from zero, so we prove it directly.
	The plan is to bound the CPS design variance by twice the PS design variance, and then apply Chebyshev's inequality as in the Poisson case.
	The key fact is that CPS indicators are negatively correlated: $\pi_{ij} \le \pi_i \pi_j$ for $i \neq j$~\citep[Theorem 3]{Branden2012negative}, which is intuitive because the underlying Poisson samples are independent and CPS samples without replacement. 
	Negative correlation only bounds the variance of a sum with same-sign coefficients, so we split $w_i = w_i^+ - w_i^-$ with $w_i^+ := \max(w_i, 0)$ and $w_i^- := \max(-w_i, 0)$.
	Write $a_i^\pm := w_i^\pm / \pi_i \ge 0$.
	\flal{
	    &   \V_d\bigparens{\hat{\mu}^w_{\*\pi} - \mu^w_N}
	        = \frac{1}{N^2} \V_d\bigparens{\sum_i s_i a_i^+ - \sum_i s_i a_i^-}
	        &&\tag{$\mu^w_N$ is fixed under the design; $w_i = w_i^+ - w_i^-$}
	    \\& \le \frac{2}{N^2} \bigbracks{\V_d\bigparens{\sum_i s_i a_i^+} + \V_d\bigparens{\sum_i s_i a_i^-}}
	            &&\tag{$V(X - Y) \le 2V(X) + 2V(Y)$}
	    \\& \V_d\bigparens{\sum_i s_i a_i^\pm}
	        = \sum_i (a_i^\pm)^2 \pi_i (1 - \pi_i) + \sum_i \sum_{j \neq i} a_i^\pm a_j^\pm (\pi_{ij} - \pi_i \pi_j)
	            &&\tag{$\V_d(s_i) = \pi_i(1-\pi_i)$, $\mathrm{Cov}_d(s_i, s_j) = \pi_{ij} - \pi_i \pi_j$}
	    \\& \le \sum_i (a_i^\pm)^2 \pi_i (1 - \pi_i)
	        = \sum_i (w_i^\pm)^2 \frac{1 - \pi_i}{\pi_i}
	            &&\tag{$a_i^\pm a_j^\pm \ge 0$ and $\pi_{ij} \le \pi_i \pi_j$}
	    \\& \implies \V_d\bigparens{\hat{\mu}^w_{\*\pi} - \mu^w_N}
	        \le \frac{2}{N^2} \sum_i \bigparens{(w_i^+)^2 + (w_i^-)^2} \frac{1 - \pi_i}{\pi_i}
	        = \frac{2}{N^2} \sum_i w_i^2 \frac{1 - \pi_i}{\pi_i}
	            &&\tag{$(w_i^+)^2 + (w_i^-)^2 = w_i^2$}
	    \\& \le \frac{2}{\gamma N} \cdot \frac{1}{N} \sum_i w_i^2
	        = O(1/N)
	            &&\tag{$\pi_i \ge \gamma$; $\frac{1}{N}\sum_i w_i^2 = O(1)$ as in the Poisson case}
	}
	The estimator is design-unbiased, since $\E_d[s_i] = \pi_i$.
	Chebyshev's inequality then gives $\hat{\mu}^w_{\*\pi} - \mu^w_N \cip 0$, with no condition on $\sigma^2_{\textup{cps}}(\*\pi, N)$.
\end{proof}

\begin{lemma}[Cram\'er--Wold Device]\label{lemma:cramer_wold}
	Let $\{X_N\}_{N\ge 1}$ and $X$ be random vectors in $\R^q$. Then $X_N \cid X$ if and only if $a^\top X_N \cid a^\top X$ for every $a \in \R^q$.
\end{lemma}

\begin{restatable}[Vector Form of \cref{lemma:ht_design}]{lemma}{VectorAsympNormality}\label{lemma:ht_design_vector}
	Fix $\omega\in\Omega$ such that~\cref{asmp:design_assumption_constraints_on_pi} holds at $\omega$.
	Given measurable $w:\mathcal Y\times\mathcal X\to\R^q$, write $w_i := w(Y_i(\omega),X_i(\omega))$, $\hat\mu^w_{\*\pi} := N^{-1}\sum_{i=1}^N (s_i/\pi_i) w_i$, and $\mu^w_N := N^{-1}\sum_{i=1}^N w_i$, where $\*s$ is drawn from either $\p_{\textup{ps}(\*\pi)}$ or $\p_{\textup{cps}(\*\pi)}$.
	Assume that $\limsup_{N\to\infty} \frac{1}{N}\sum_{i=1}^N \|w_i\|^4 < \infty$, and define the design covariance matrices
	\begin{align*}
		\Sigma_{\textup{ps}}(\*\pi,N)
		 & \;=\; \frac{1}{N}\sum_{i=1}^{N} w_i w_i^\top \,\frac{1-\pi_i}{\pi_i}, \\
		\Sigma_{\textup{cps}}(\*\pi,N)
		 & \;=\; \frac{1}{N}\sum_{i=1}^{N} z_i z_i^\top \,\frac{1-\pi_i}{\pi_i},
		\qquad
		z_i \;:=\; w_i - \pi_i\,\frac{\sum_j w_j(1-\pi_j)}{\sum_k \pi_k(1-\pi_k)} \in\R^q.
	\end{align*}
	Then, in design probability $p_N(\cdot,Z^N(\omega))$:
	\begin{enumerate}
		\item \textup{(Consistency.)}
		      $\hat\mu^w_{\*\pi} - \mu^w_N \cip 0$
		      as $N\to\infty$.
		\item \textup{(Asymptotic Normality.)}
		      If $\Sigma_\bullet(\*\pi,N)$ (where $\bullet \in \{\textup{ps, cps}\}$) converges to a positive definite limit
			$\Sigma_\infty\in\R^{q\times q}$, then
		      \begin{equation*}
			      \sqrt N\bigl(\hat\mu^w_{\*\pi}-\mu^w_N\bigr)
			      \;\cid\;
			      \mathcal N_q(0,\Sigma_\infty)
			      \qquad \text{as } N\to\infty.
		      \end{equation*}
	\end{enumerate}
\end{restatable}

\begin{proof}
	
	Consistency follows from the scalar lemma applied to each of the finitely many coordinates.
	For $k = 1 \commadots q$, write $w_{ik} := e_k^\top w_i$ for the $k$th coordinate of $w_i$. The $k$th coordinate of $\hat\mu^w_{\*\pi} - \mu^w_N$ is $\hat\mu^{w_k}_{\*\pi} - \mu^{w_k}_N$, the error of the scalar Horvitz--Thompson estimator for the values $w_{1k} \commadots w_{Nk}$. Since $|w_{ik}| \le \|w_i\|$, we have 
	\al{
	    & \limsup_{N} \frac{1}{N}\sum_i w_{ik}^4 \le \limsup_{N} \frac{1}{N}\sum_i \|w_i\|^4 < \infty.
	}
	So the consistency part of~\cref{lemma:ht_design} gives $\hat\mu^{w_k}_{\*\pi} - \mu^{w_k}_N \cip 0$ for each $k$. Because there are finitely many coordinates, this gives convergence of the vector: for any $\epsilon > 0$,
	\flal{
	    &   P_d\bigparens{\|\hat\mu^w_{\*\pi} - \mu^w_N\| > \epsilon}
	        \le P_d\bigparens{\sum_{k=1}^q |\hat\mu^{w_k}_{\*\pi} - \mu^{w_k}_N| > \epsilon}
	        &&\tag{triangle inequality: $\|v\| \le \sum_k |v_k|$}
	    \\& \le P_d\bigparens{|\hat\mu^{w_k}_{\*\pi} - \mu^{w_k}_N| > \epsilon/q \text{ for some } k}
	            &&\tag{if the sum is $> \epsilon$, then at least 1 term is $> \epsilon/q$}
	    \\& \le \sum_{k=1}^q P_d\bigparens{|\hat\mu^{w_k}_{\*\pi} - \mu^{w_k}_N| > \epsilon/q}
	            &&\tag{union bound}
	    \\& \to 0
	            &&\tag{each of the $q$ terms $\to 0$ by~\cref{lemma:ht_design}}
	}

	For asymptotic normality, the Cram\'er--Wold device~(\cref{lemma:cramer_wold}) lets us show the vector claim by showing $\sqrt N\, a^\top(\hat\mu^w_{\*\pi} - \mu^w_N) \cid \mathcal N(0, a^\top \Sigma_\infty a)$ for every $a \in \R^q$. 
	We can show this for $\|a\| = 1$, which implies it for any $a$ since scaling $a$ by $c$ scales both sides by $c$.

	For $w_i^{(a)} := a^\top w_i$ we have $a^\top \hat\mu^w_{\*\pi} = \frac{1}{N}\sum_i \frac{s_i}{\pi_i} a^\top w_i$. This is the HT estimator for $w_i^{(a)}$, and $a^\top \mu^w_N$ is their mean. So~\cref{lemma:ht_design} applies to $a^\top(\hat\mu^w_{\*\pi} - \mu^w_N)$ with $w_i^{(a)}$ in place of $w_i$, as long as $\limsup_N \frac{1}{N}\sum_i (w_i^{(a)})^4 < \infty$. We only assumed the vector version, $\limsup_N \frac{1}{N}\sum_i \|w_i\|^4 < \infty$, but it implies the scalar version:
	\flal{
	    &   \frac{1}{N}\sum_i (w_i^{(a)})^4
	        = \frac{1}{N}\sum_i |a^\top w_i|^4
	        \le \frac{1}{N}\sum_i \|a\|^4 \|w_i\|^4
	        &&\tag{Cauchy--Schwarz: $|a^\top w_i| \le \|a\|\,\|w_i\|$}
	    \\& = \frac{1}{N}\sum_i \|w_i\|^4
	            &&\tag{$\|a\| = 1$}
	}
	so taking $\limsup_N$ of both sides, the assumption holds. 

	The other hypothesis of~\cref{lemma:ht_design} is that its scalar variance is bounded away from zero. Write $\sigma^2_\bullet(\*\pi, N; a)$ for that variance computed from the values $w_i^{(a)}$. We cannot assume anything about it directly; what we did assume is that the matrices $\Sigma_\bullet(\*\pi, N)$ converge to $\Sigma_\infty$. So we rewrite the scalar variance in terms of the matrix: $\sigma^2_\bullet(\*\pi, N; a) = a^\top \Sigma_\bullet(\*\pi, N)\, a$.

	\textit{PS.} Plugging $w_i^{(a)}$ into the definition of $\sigma^2_{\textup{ps}}$ in~\cref{lemma:ht_design},
	\flal{
	    &   \sigma^2_{\textup{ps}}(\*\pi, N; a)
	        = \frac{1}{N}\sum_i (a^\top w_i)^2 \,\frac{1-\pi_i}{\pi_i}
	        &&
	    \\& = \frac{1}{N}\sum_i a^\top w_i w_i^\top a \,\frac{1-\pi_i}{\pi_i}
	            &&\tag{$(a^\top w_i)^2 = a^\top w_i \, w_i^\top a$}
	    \\& = a^\top \bigbracks{\frac{1}{N}\sum_i w_i w_i^\top \,\frac{1-\pi_i}{\pi_i}} a
	        = a^\top \Sigma_{\textup{ps}}(\*\pi, N)\, a
	            &&\tag{$a$ does not depend on $i$}
	}

	\textit{CPS.} The only new ingredient is that the centering is linear in the values. Write $\beta := \frac{\sum_j w_j(1-\pi_j)}{\sum_k \pi_k(1-\pi_k)} \in \R^q$, so that $z_i = w_i - \pi_i \beta$. The scalar centering ratio for the values $w_i^{(a)}$ is
	\al{
	    & \beta^{(a)} = \frac{\sum_j a^\top w_j(1-\pi_j)}{\sum_k \pi_k(1-\pi_k)} = a^\top \beta,
	    \qquad\text{so}\qquad
	    z_i^{(a)} = a^\top w_i - \pi_i\, a^\top \beta = a^\top z_i.
	}
	The same three steps as for PS, with $z_i$ in place of $w_i$, give $\sigma^2_{\textup{cps}}(\*\pi, N; a) = a^\top \Sigma_{\textup{cps}}(\*\pi, N)\, a$.

	\textit{Bounded away from zero.} By assumption $\Sigma_\bullet(\*\pi, N) \to \Sigma_\infty$, so $\sigma^2_\bullet(\*\pi, N; a) = a^\top \Sigma_\bullet(\*\pi, N)\, a \to a^\top \Sigma_\infty a$, and $a^\top \Sigma_\infty a > 0$ because $\Sigma_\infty$ is positive definite. So the sequence $N \mapsto \sigma^2_\bullet(\*\pi, N; a)$ converges to the positive number $L := a^\top \Sigma_\infty a$. Taking $\epsilon = L/2$ in the definition of this limit, $\sigma^2_\bullet(\*\pi, N; a) \ge L/2$ for all $N$ large enough, so it is bounded away from zero.

	\textit{Conclusion.} Both conditions hold, so~\cref{lemma:ht_design} gives $\sqrt N\, a^\top(\hat\mu^w_{\*\pi} - \mu^w_N) / \sigma_\bullet(\*\pi, N; a) \cid \mathcal N(0, 1)$. Since $\sigma_\bullet(\*\pi, N; a) \to (a^\top \Sigma_\infty a)^{1/2}$, Slutsky's theorem gives $\sqrt N\, a^\top(\hat\mu^w_{\*\pi} - \mu^w_N) \cid \mathcal N(0, a^\top \Sigma_\infty a)$. This holds for every unit $a$, so the Cram\'er--Wold device~(\cref{lemma:cramer_wold}) gives $\sqrt N(\hat\mu^w_{\*\pi} - \mu^w_N) \cid \mathcal N_q(0, \Sigma_\infty)$.
\end{proof}

\subsubsection{Convergence of Design Covariance}
\label{app:convergence_design_covariance_main}

We now show that $\Sigma_{\bullet}(\*\pi, N)$ from~\cref{lemma:ht_design_vector} converges to a particular non-random limit. This is the primary non-standard and non-trivial part of our proof, since we need to demonstrate convergence $P$-almost surely, while accounting for the fact that both $\*\pi^{(N)}$ and $\theta_N$ are functions of the finite population.

\paragraph{Helper Lemmas.}
\label{app:helper_lemmas}

We will make use of the following helper lemmas. The first two lemmas are elementary perturbation bounds that we will invoke repeatedly when showing convergence of the design covariance.  Both follow from the algebraic identity $aa^\top - bb^\top = (a-b)a^\top + b(a-b)^\top$ and Cauchy--Schwarz on the empirical inner product.

\begin{lemma}[Outer-Product Mean Perturbation]\label{lemma:outer_product_perturbation}
	Let $a_1, \ldots, a_N, b_1, \ldots, b_N \in \R^q$ and $\alpha_1, \ldots, \alpha_N \in [0, M]$ for some $M \geq 0$. Then
	\begin{equation*}
		\Big\|\frac{1}{N}\sum_{i=1}^N \alpha_i \bigl(a_i a_i^\top - b_i b_i^\top\bigr)\Big\|
		\;\leq\; M\,\sqrt{\frac{1}{N}\sum_{i=1}^N \|a_i - b_i\|^2}\,\sqrt{\frac{1}{N}\sum_{i=1}^N \bigl(\|a_i\| + \|b_i\|\bigr)^2}.
	\end{equation*}
\end{lemma}
\begin{proof}
	The identity $a_i a_i^\top - b_i b_i^\top = (a_i - b_i) a_i^\top + b_i (a_i - b_i)^\top$ together with the bound $\|uv^\top\| \leq \|u\|\|v\|$ and the triangle inequality yields
	\begin{equation*}
		\bigl\|a_i a_i^\top - b_i b_i^\top\bigr\| \;\leq\; \|a_i - b_i\|\bigl(\|a_i\| + \|b_i\|\bigr).
	\end{equation*}
	Applying this term-by-term and using $|\alpha_i| \leq M$,
	\begin{equation*}
		\Big\|\frac{1}{N}\sum_i \alpha_i \bigl(a_i a_i^\top - b_i b_i^\top\bigr)\Big\| \;\leq\; \frac{M}{N}\sum_i \|a_i - b_i\|\bigl(\|a_i\| + \|b_i\|\bigr),
	\end{equation*}
	and Cauchy--Schwarz on the empirical inner product completes the proof.
\end{proof}

\begin{lemma}[Weighted Mean Under Weight Perturbation]\label{lemma:weight_perturbation}
	Let $a_1, \ldots, a_N \in \R^q$ and $\alpha_1, \ldots, \alpha_N, \alpha'_1, \ldots, \alpha'_N \in \R$. Then
	\begin{equation*}
		\Big\|\frac{1}{N}\sum_{i=1}^N a_i a_i^\top \bigl(\alpha_i - \alpha'_i\bigr)\Big\|
		\;\leq\; \sqrt{\frac{1}{N}\sum_{i=1}^N \|a_i\|^4}\,\sqrt{\frac{1}{N}\sum_{i=1}^N \bigl(\alpha_i - \alpha'_i\bigr)^2}.
	\end{equation*}
\end{lemma}
\begin{proof}
	Since $\|a_i a_i^\top\| = \|a_i\|^2$, the triangle inequality gives
	\begin{equation*}
		\Big\|\frac{1}{N}\sum_i a_i a_i^\top \bigl(\alpha_i - \alpha'_i\bigr)\Big\| \;\leq\; \frac{1}{N}\sum_i \|a_i\|^2\,|\alpha_i - \alpha'_i|,
	\end{equation*}
	and Cauchy--Schwarz on the empirical inner product completes the proof.
\end{proof}

The next lemma is a Lipschitz bound on the inverse-probability weight $\alpha(\pi) := (1-\pi)/\pi$, which we invoke whenever comparing finite-population quantities formed under different inclusion probabilities.

\begin{lemma}[Lipschitz Inverse-Probability Weight]\label{lemma:phi_lipschitz}
	Fix $\gamma \in (0, 1/2)$ and let $\alpha(\pi) := (1-\pi)/\pi$. Then $|\alpha(a) - \alpha(b)| \le |a-b|/\gamma^2$ for all $a, b \in [\gamma, 1-\gamma]$.
\end{lemma}
\begin{proof}
	Combining fractions, $|\alpha(a) - \alpha(b)| = |\frac{(1-a)b - (1-b)a}{ab}| = |\frac{b - a}{ab}|$. The bound $a, b \ge \gamma$ gives $ab \ge \gamma^2 > 0$, hence $|\alpha(a) - \alpha(b)| \le |a-b|/\gamma^2$.
\end{proof}

We will also use the following elementary convexity inequality to bound moments of sums.

\begin{lemma}[Power-Mean Inequality]\label{lemma:power_mean}
	For $a, b \in \R$ and $p \ge 1$, $|a + b|^p \le 2^{p-1}\bigl(|a|^p + |b|^p\bigr)$. In particular, $(a+b)^p \le 2^{p-1}(a^p + b^p)$ for $a, b \ge 0$.
\end{lemma}
\begin{proof}
	The map $x \mapsto |x|^p$ is convex on $\R$ for $p \ge 1$, so Jensen's inequality gives $\bigl|\tfrac{a+b}{2}\bigr|^p \le \tfrac{|a|^p + |b|^p}{2}$; multiplying by $2^p$ yields the claim.
\end{proof}

The next lemma shows that the various CPS centering vectors $\beta$ are stable under perturbations of either the inputs $u_i$ or the inclusion-probability weights $\eta_i$.

\begin{lemma}[Stability of $\beta$]\label{lemma:cps_centering_stability}
	Fix $\gamma \in (0, 1/2)$. Let $\{u_i\}_{i=1}^N, \{u'_i\}_{i=1}^N \subset \R^q$ and $\{\eta_i\}_{i=1}^N, \{\eta'_i\}_{i=1}^N \subset [\gamma, 1-\gamma]$. Define the empirical ratios
	\begin{equation*}
		\beta \;:=\; \frac{\frac{1}{N}\sum_j u_j (1 - \eta_j)}{\frac{1}{N}\sum_k \eta_k (1 - \eta_k)}, \qquad
		\beta' \;:=\; \frac{\frac{1}{N}\sum_j u'_j (1 - \eta'_j)}{\frac{1}{N}\sum_k \eta'_k (1 - \eta'_k)}.
	\end{equation*}
	Both denominators are bounded below by $\gamma(1-\gamma) > 0$, so $\beta, \beta'$ are well-defined and satisfy the deterministic bound
	\begin{equation}\label{eq:beta_deterministic_bound}
		\|\beta\| \;\le\; \frac{1}{\gamma(1-\gamma)}\sqrt{\tfrac{1}{N}\sum_j \|u_j\|^2}
	\end{equation}
	(and analogously for $\beta'$). Furthermore, if
	\begin{equation*}
		\frac{1}{N}\sum_j \|u_j - u'_j\|^2 \;=\; o_P(1), \qquad
		\frac{1}{N}\sum_j (\eta_j - \eta'_j)^2 \;=\; o_P(1), \qquad
		\frac{1}{N}\sum_j \|u'_j\|^2 \;=\; O_P(1),
	\end{equation*}
	then $\|\beta - \beta'\| = o_P(1)$.
\end{lemma}
\begin{proof}
	Set $A := \frac{1}{N}\sum_j u_j(1-\eta_j)$, $B := \frac{1}{N}\sum_k \eta_k(1-\eta_k)$, and analogously define $A', B'$, so that $\beta = A / B$ and $\beta' = A' / B'$.  The bound in~\cref{eq:beta_deterministic_bound} is obtained as follows: Since $\eta_k \in [\gamma, 1-\gamma]$ implies $\eta_k(1-\eta_k) \geq \gamma(1-\gamma)$, we have $B, B' \geq \gamma(1-\gamma) > 0$.  Since $|1-\eta_j| \le 1$, we have $\| \sum_j u_j (1 - \eta_j) \| \leq \sum_j \|u_j\| |1 - \eta_j| \leq \sum_j \|u_j\|$ and so $\|A\| \leq \frac{1}{N} \sum_j \|u_j\|$, and Cauchy--Schwarz on the empirical inner product gives $\frac{1}{N}\sum_j \|u_j\| \le \sqrt{\frac{1}{N}\sum_j \|u_j\|^2}$. Combined with $1/B \le 1/[\gamma(1-\gamma)]$, this yields~\cref{eq:beta_deterministic_bound}; the bound for $\beta'$ is identical.

	For the remaining conclusion on the convergence of $\beta$ to $\beta'$, we start by subtracting and adding $A' / B$ to obtain the difference
	\begin{equation*}
		\beta - \beta' \;=\; \frac{A}{B} - \frac{A'}{B'} \;=\; \frac{A - A'}{B} + A'\!\left(\frac{1}{B} - \frac{1}{B'}\right) \;=\; \frac{A - A'}{B} - \frac{A'}{B'}\cdot\frac{B - B'}{B},
	\end{equation*}
	so by the triangle inequality and the upper bound $1 / B \leq 1 / [\gamma(1-\gamma)]$,
	\begin{equation}\label{eq:ratio_identity}
		\|\beta - \beta'\| \;\leq\; \frac{\|A - A'\|}{\gamma(1-\gamma)} + \frac{\|\beta'\|\,|B - B'|}{\gamma(1-\gamma)}.
	\end{equation}
	It remains to bound $\|A - A'\|$, $|B - B'|$, and $\|\beta'\|$.

	\emph{Bounding $\|A - A'\|$.} Writing $u_j(1-\eta_j) - u'_j(1-\eta'_j) = (u_j - u'_j)(1-\eta_j) + u'_j(\eta'_j - \eta_j)$ (adding and subtracting $u'_j \eta_j$), the triangle inequality and $|1 - \eta_j| \leq 1$ give
	\begin{equation*}
		\|A - A'\| \;\leq\; \frac{1}{N}\sum_j \|u_j - u'_j\| + \frac{1}{N}\sum_j \|u'_j\|\,|\eta_j - \eta'_j|.
	\end{equation*}
	Cauchy--Schwarz on the empirical inner product (with weights $1/N$) yields
	\begin{equation}\label{eq:AA_bound}
		\|A - A'\| \;\leq\; \sqrt{\tfrac{1}{N}\sum_j \|u_j - u'_j\|^2} + \sqrt{\tfrac{1}{N}\sum_j \|u'_j\|^2}\,\sqrt{\tfrac{1}{N}\sum_j (\eta_j - \eta'_j)^2}.
	\end{equation}

	\emph{Bounding $|B - B'|$.} The map $\pi \mapsto \pi(1-\pi)$ has derivative $1 - 2\pi \in [-1, 1]$, hence is $1$-Lipschitz on $[0, 1]$, so $|\eta_k(1-\eta_k) - \eta'_k(1-\eta'_k)| \leq |\eta_k - \eta'_k|$. Cauchy--Schwarz gives
	\begin{equation}\label{eq:BB_bound}
		|B - B'| \;\leq\; \frac{1}{N}\sum_k |\eta_k - \eta'_k| \;\leq\; \sqrt{\tfrac{1}{N}\sum_k (\eta_k - \eta'_k)^2}.
	\end{equation}

	\emph{Bounding $\|\beta'\|$.} By the deterministic bound in~\cref{eq:beta_deterministic_bound} applied to $\beta'$ and the assumption $\frac{1}{N}\sum_j \|u'_j\|^2 = O_P(1)$, $\|\beta'\| = O_P(1)$.

	Both bounds are deterministic inequalities. Under the three rate conditions, \cref{eq:AA_bound} gives $\|A - A'\| = o_P(1) + O_P(1)\cdot o_P(1) = o_P(1)$, and \cref{eq:BB_bound} gives $|B - B'| = o_P(1)$.
	Substituting into~\cref{eq:ratio_identity}, $\|\beta - \beta'\| \leq O(1)\cdot o_P(1) + O_P(1)\cdot o_P(1) = o_P(1)$, which completes the proof.
\end{proof}

\paragraph{Main Statement.}
\label{app:convergence_design_covariance}

Recall that~\cref{lemma:ht_design_vector} defines the design covariance matrices
\begin{align*}
	\Sigma_{\textup{ps}}(\*\pi,N)
	 & \;=\; \frac{1}{N}\sum_{i=1}^{N} w_i w_i^\top \,\frac{1-\pi_i}{\pi_i}, \\
	\Sigma_{\textup{cps}}(\*\pi,N)
	 & \;=\; \frac{1}{N}\sum_{i=1}^{N} z_i z_i^\top \,\frac{1-\pi_i}{\pi_i},
	\qquad
	z_i \;:=\; w_i - \pi_i\,\frac{\sum_j w_j(1-\pi_j)}{\sum_k \pi_k(1-\pi_k)} \in\R^q.
\end{align*}
\Cref{lemma:ht_design_vector} states that \textit{if each converges $P$-almost surely to a positive definite limit ($\Sigma_{\infty,\textup{ps}}$ and $\Sigma_{\infty,\textup{cps}}$, respectively)}, then $\sqrt{N}(\hat{\mu}^w_{\*\pi} - \mu^{w}_N)$ converges in distribution to a normal distribution with the resulting covariance.  For the use of this result in the context of verifying~\cref{thm:rb_main}, we will also need that this limit is non-stochastic in the super-population (i.e., for any $\omega$, and hence any sequence of observations, we converge to the same target).

We now set out to demonstrate this convergence of the design covariance matrices, $P$-a.s. This demonstration is complicated by the fact that there will be two sources of variation across different finite populations indexed by $\omega$.  First, our $w_i$ depends on the parameter $\theta_N$ via $w_i = w_i(\theta_N) := g_{\theta_N}(X_i, Y_i) - g_{\theta_N}(X_i, f(X_i))$, and so depends on the particular realization $\omega$.  Second, our $\*\pi^{(N)}$ depends on the particular realization $\omega$ as well, and further depends on the entire finite population $\{X_i\}_{i=1}^{N}$, so we need the condition that it converges not only to a fixed function $\pis$ (from~\cref{asmp:design_assumption_pi_lim}), but also that it does so in a way that still admits convergence of the covariance matrix to a non-random limit that does not depend on $\omega$.

With all of that in mind, we extend the notation of~\cref{lemma:ht_design_vector} by writing $\Sigma_{\textup{ps}}(\*\pi, \theta, N), \Sigma_{\textup{cps}}(\*\pi, \theta, N)$ for the matrix obtained when $w_i$ is evaluated at $\theta$ and the inclusion-probability vector is $\*\pi = (\pi_1, \ldots, \pi_N)$ (where, when a function $\pis: \mathcal{X} \to [\gamma, 1-\gamma]$ is supplied in place of $\*\pi$, we read $\pi_i = \pis(X_i)$).  Suppressing the explicit dependence on $\omega$, the quantity to control is then $\Sigma_\bullet(\*\pi^{(N)}, \theta_N, N)$, given by
\begin{align}
	\Sigma_{\textup{ps}}(\*\pi^{(N)}, \theta_N, N)
	 & \;=\; \frac{1}{N}\sum_{i=1}^{N} w_i(\theta_N) w_i(\theta_N)^\top \,\frac{1-\pi^{(N)}_i}{\pi^{(N)}_i}, \label{eq:sigma_ps}                                          \\
	\Sigma_{\textup{cps}}(\*\pi^{(N)}, \theta_N, N)
	 & \;=\; \frac{1}{N}\sum_{i=1}^{N} (w_i(\theta_N) - \pi^{(N)}_i \beta^{(N)}) (w_i(\theta_N) - \pi^{(N)}_i \beta^{(N)})^\top\,\frac{1-\pi^{(N)}_i}{\pi^{(N)}_i}, \label{eq:sigma_cps}
\end{align}
where $\beta^{(N)} \;:=\; \frac{\sum_j w_j(\theta_N) (1-\pi^{(N)}_j)}{\sum_k \pi^{(N)}_k(1-\pi^{(N)}_k)}$.

\begin{lemma}[Convergence of Design Covariance]\label{lemma:convergence_of_covariance}
	Under~\cref{asmp:design_assumptions,asmp:model_assumptions}, define $w^\star := \gY[\theta_0] - \gF[\theta_0]$ and write $\pis$ for $\pis(X)$, with $\pis$ from~\cref{asmp:design_assumption_pi_lim}.   Further define the matrices
	\begin{align*}
		\Sigma_{\infty,\textup{ps}}
		      & := \E_m\left[w^\star w^{\star\top}(1-\pis) / \pis\right],                                                       \\
		\Sigma_{\infty,\textup{cps}}
		      & := \E_m\!\left[\bigl(w^\star-\pis \beta\bigr)\bigl(w^\star-\pis \beta\bigr)^\top\,\frac{1-\pis}{\pis}\right], &
		\beta & := \frac{\E_m\bigl[w^\star\bigl(1-\pis\bigr)\bigr]}{\E_m\bigl[\pis\bigl(1-\pis\bigr)\bigr]}.
	\end{align*}
	Then, for $\bullet \in \{\textup{ps}, \textup{cps}\}$,
	\begin{equation}
		\Sigma_\bullet(\*\pi^{(N)}, \theta_N, N) \xrightarrow{P\text{-a.s.}} \Sigma_{\infty,\bullet} \qquad \text{as } N \to \infty,
	\end{equation}
	where $\Sigma_{\textup{ps}}, \Sigma_{\textup{cps}}$ are defined in~\cref{eq:sigma_ps,eq:sigma_cps}. Note that since $\Sigma_\bullet(\*\pi^{(N)}, \theta_N, N)$ does not involve the sample $\*s$, it depends only on $\omega$; the convergence is therefore an ordinary limit of matrices for $P$-a.e.\ $\omega$, with no design randomness.
\end{lemma}
\begin{proof}
	The candidate limits are well-defined: the overlap bound $\gamma \le \pis(X) \le 1-\gamma$ from~\cref{asmp:design_assumption_pi_lim} controls $1/\pis$ and $1-\pis$ uniformly, and~\cref{asmp:score_integrability_theta0} together with $\|w^\star\| \le \|\gY[\theta_0]\| + \|\gF[\theta_0]\|$ ensures $\E_m \|w^\star\|^4 < \infty$, and hence $\E_m \|w^\star\|^2 < \infty$.

	We prove $\Sigma_\bullet(\*\pi^{(N)}, \theta_N, N) \xrightarrow{P\text{-a.s.}} \Sigma_{\infty,\bullet}$ in three steps, telescoping through $\Sigma_\bullet(\*\pi^{(N)}, \theta_0, N)$ (which uses the realized inclusion probabilities but evaluates $w_i$ at $\theta_0$, giving $w_i^\star$) and $\Sigma_\bullet(\pis, \theta_0, N)$ (which further replaces $\pi^{(N)}_i$ with $\pis(X_i)$).

	\paragraph{Step 0: Preliminaries.}
	Steps 1 and 2 use three facts along $P$-a.e.\ sample path, for $k \in \{1, 2, 4\}$: (a) $w_i$ is close to $w_i^\star$ on average, (b) the moments of $w_i^\star$ are bounded, and (c) the moments of $w_i$ are bounded.
	Throughout, fix $\omega$ in the intersection of the following probability-one events: the events in~\cref{asmp:design_assumption_constraints_on_pi,asmp:design_assumption_pi_lim}, the event $\theta_N \to \theta_0$ from~\cref{asmp:z_estimator_consistency}, and the events on which the SLLN holds for every average used below.
	There are countably many such events, so their intersection has probability one.
	We also take $N$ large enough that $N \ge N_0(\omega)$, so that $\gamma \le \pi^{(N)}_i \le 1-\gamma$ for all $i$, and that $\theta_N \in B(\theta_0)$.
	Fact (a) follows from the Lipschitz assumption once $\theta_N$ is close to $\theta_0$: by~\cref{asmp:z_estimator_consistency}, $\theta_N \to \theta_0$ $P$-a.s., so $\theta_N \in B(\theta_0)$ for all $N$ large enough.
	Fact (b) follows from the strong law of large numbers, and fact (c) combines (a) and (b).
	\flal{
	    &   \text{(a)} \quad \frac{1}{N}\sum_i \|w_i - w_i^\star\|^k
	        &&
	    \\& = \frac{1}{N}\sum_i \bigl\| \bigl[\giY[\theta_N] - \giY[\theta_0]\bigr] - \bigl[\giF[\theta_N] - \giF[\theta_0]\bigr] \bigr\|^k
	            &&\tag{definition of $w_i, w_i^\star$}
	    \\& \le \frac{1}{N}\sum_i \bigparens{\|\giY[\theta_N] - \giY[\theta_0]\| + \|\giF[\theta_N] - \giF[\theta_0]\|}^k
	            &&\tag{triangle inequality}
	    \\& \le \|\theta_N - \theta_0\|^k \, \frac{1}{N}\sum_i \bigparens{M(X_i, Y_i) + M(X_i, f(X_i))}^k
	            &&\tag{\cref{asmp:score_lipschitz}, $\theta_N \in B(\theta_0)$}
	    \\& \le \|\theta_N - \theta_0\|^k \cdot 2^{k-1} \bigparens{\frac{1}{N}\sum_i M(X_i, Y_i)^k + \frac{1}{N}\sum_i M(X_i, f(X_i))^k}
	            &&\tag{\cref{lemma:power_mean}}
	    \\& = o(1) \cdot O(1) = o(1) \quad P\text{-a.s.}
	            &&\tag{\cref{asmp:z_estimator_consistency}; SLLN with $\E_m[M^k] < \infty$ by \cref{asmp:score_lipschitz}}
	    \\& \text{(b)} \quad \E_m\|w^\star\|^4
	            &&
	    \\& \le \E_m \bigparens{\|\gY[\theta_0]\| + \|\gF[\theta_0]\|}^4
	            &&\tag{definition of $w^\star$, triangle inequality}
	    \\& \le 8 \bigparens{\E_m\|\gY[\theta_0]\|^4 + \E_m\|\gF[\theta_0]\|^4} < \infty
	            &&\tag{\cref{lemma:power_mean}, \cref{asmp:score_integrability_theta0}}
	    \\& \implies \frac{1}{N}\sum_i \|w_i^\star\|^k \xrightarrow{P\text{-a.s.}} \E_m\|w^\star\|^k < \infty, \text{ so } \frac{1}{N}\sum_i \|w_i^\star\|^k = O(1) \quad P\text{-a.s.}
	            &&\tag{SLLN; $k < 4$ by Jensen}
	    \\& \text{(c)} \quad \frac{1}{N}\sum_i \|w_i\|^k
	            &&
	    \\& \le 2^{k-1} \bigparens{\frac{1}{N}\sum_i \|w_i^\star\|^k + \frac{1}{N}\sum_i \|w_i - w_i^\star\|^k}
	            &&\tag{$w_i = w_i^\star + (w_i - w_i^\star)$, \cref{lemma:power_mean}}
	    \\& = O(1) + o(1) = O(1) \quad P\text{-a.s.}
	            &&\tag{(b), (a)}
	}

	\paragraph{Step 1: $\theta_N \to \theta_0$ Replacement.}
	The strong law of large numbers needs an average of i.i.d.\ terms, but $w_i = w_i(\theta_N)$ is not i.i.d.\ across $i$, because $\theta_N$ depends on all $N$ units.
	We therefore replace $w_i$ by $w_i^\star = w_i(\theta_0)$, which depends only on $(X_i, Y_i)$, and show that the replacement changes $\Sigma_\bullet$ by $o(1)$ $P$-a.s.
	Since $\Sigma_\bullet$ is an average of weighted outer products, the change factors as (average distance between the vectors) $\times$ (average size of the vectors), and Step 0 shows that the first factor is $o(1)$ and the second is $O(1)$.
	Throughout, write $\alpha_i := (1 - \pi^{(N)}_i)/\pi^{(N)}_i$, and note that $0 \le \alpha_i \le 1/\gamma$ because $\pi^{(N)}_i \in [\gamma, 1-\gamma]$ by~\cref{asmp:design_assumption_constraints_on_pi}.

	\textit{Poisson case.} The PS covariance is built directly from $w_i$, so one application of~\cref{lemma:outer_product_perturbation} suffices.
	\flal{
	    &   \bigl\|\Sigma_{\textup{ps}}(\*\pi^{(N)}, \theta_N, N) - \Sigma_{\textup{ps}}(\*\pi^{(N)}, \theta_0, N)\bigr\|
	        &&
	    \\& = \Bigl\|\frac{1}{N}\sum_i \alpha_i \bigl(w_i w_i^\top - w_i^\star w_i^{\star\top}\bigr)\Bigr\|
	            &&\tag{\cref{eq:sigma_ps}}
	    \\& \le \frac{1}{\gamma}\sqrt{\frac{1}{N}\sum_i \|w_i - w_i^\star\|^2}\,\sqrt{\frac{1}{N}\sum_i \bigl(\|w_i\| + \|w_i^\star\|\bigr)^2}
	            &&\tag{\cref{lemma:outer_product_perturbation}, $\alpha_i \le 1/\gamma$}
	    \\& \le \frac{1}{\gamma}\sqrt{\frac{1}{N}\sum_i \|w_i - w_i^\star\|^2}\,\sqrt{\frac{2}{N}\sum_i \|w_i\|^2 + \frac{2}{N}\sum_i \|w_i^\star\|^2}
	            &&\tag{\cref{lemma:power_mean}}
	    \\& = o(1) \cdot O(1) = o(1) \quad P\text{-a.s.}
	            &&\tag{Step 0 (a), (c), (b)}
	}

	\textit{CPS case.} The CPS covariance is built from the centered vectors $z_i := w_i - \pi^{(N)}_i \beta^{(N)}$, where
	\al{
	    & \beta^{(N)} := \frac{\frac{1}{N}\sum_j w_j (1 - \pi^{(N)}_j)}{\frac{1}{N}\sum_k \pi^{(N)}_k (1 - \pi^{(N)}_k)},
	}
	and we define $z_i^\star := w_i^\star - \pi^{(N)}_i \beta^{(N)}_\star$, with $\beta^{(N)}_\star$ the same ratio with $w_j^\star$ in place of $w_j$.
	Replacing $\theta_N$ by $\theta_0$ changes every $w_i$ and also the centering $\beta^{(N)}$, which averages over all units.
	We therefore show (i) $\beta^{(N)}$ is close to $\beta^{(N)}_\star$, then (ii) $z_i$ is close to $z_i^\star$ on average, then (iii) the $z_i$ and $z_i^\star$ have bounded size on average, and finally (iv) apply~\cref{lemma:outer_product_perturbation} to $z$ as in the Poisson case.
	Both ratios have the same denominator, which is at least $\gamma(1-\gamma)$ because $\pi^{(N)}_k \in [\gamma, 1-\gamma]$.
	\flal{
	    &   \text{(i)} \quad \bigl\|\beta^{(N)} - \beta^{(N)}_\star\bigr\|
	        &&
	    \\& = \frac{\bigl\|\frac{1}{N}\sum_j (w_j - w_j^\star)(1 - \pi^{(N)}_j)\bigr\|}{\frac{1}{N}\sum_k \pi^{(N)}_k (1 - \pi^{(N)}_k)}
	            &&\tag{same denominator}
	    \\& \le \frac{1}{\gamma(1-\gamma)} \cdot \frac{1}{N}\sum_j \|w_j - w_j^\star\|
	            &&\tag{triangle inequality, $|1 - \pi^{(N)}_j| \le 1$}
	    \\& \le \frac{1}{\gamma(1-\gamma)} \sqrt{\frac{1}{N}\sum_j \|w_j - w_j^\star\|^2} = o(1) \quad P\text{-a.s.}
	            &&\tag{Jensen; Step 0 (a)}
	}
	\flal{
	    &   \text{(ii)} \quad \frac{1}{N}\sum_i \|z_i - z_i^\star\|^2
	            &&
	    \\& = \frac{1}{N}\sum_i \bigl\|(w_i - w_i^\star) - \pi^{(N)}_i \bigl(\beta^{(N)} - \beta^{(N)}_\star\bigr)\bigr\|^2
	            &&\tag{definition of $z_i, z_i^\star$}
	    \\& \le \frac{2}{N}\sum_i \|w_i - w_i^\star\|^2 + 2\,\bigl\|\beta^{(N)} - \beta^{(N)}_\star\bigr\|^2
	            &&\tag{\cref{lemma:power_mean}, $\pi^{(N)}_i \le 1$}
	    \\& = o(1) \quad P\text{-a.s.}
	            &&\tag{Step 0 (a), (i)}
	}
	\flal{
	    &   \text{(iii)} \quad \frac{1}{N}\sum_i \|z_i\|^2
	            &&
	    \\& \le \frac{2}{N}\sum_i \|w_i\|^2 + 2\,\bigl\|\beta^{(N)}\bigr\|^2
	            &&\tag{\cref{lemma:power_mean}, $\pi^{(N)}_i \le 1$}
	    \\& \le \Bigl(2 + \frac{2}{\gamma^2(1-\gamma)^2}\Bigr) \frac{1}{N}\sum_i \|w_i\|^2 = O(1) \quad P\text{-a.s.}
	            &&\tag{\cref{eq:beta_deterministic_bound}; Step 0 (c)}
	    \\& \text{and likewise } \frac{1}{N}\sum_i \|z_i^\star\|^2 = O(1) \quad P\text{-a.s.}
	            &&\tag{same steps; Step 0 (b)}
	}
	\flal{
	    &   \text{(iv)} \quad \bigl\|\Sigma_{\textup{cps}}(\*\pi^{(N)}, \theta_N, N) - \Sigma_{\textup{cps}}(\*\pi^{(N)}, \theta_0, N)\bigr\|
	            &&
	    \\& = \Bigl\|\frac{1}{N}\sum_i \alpha_i \bigl(z_i z_i^\top - z_i^\star z_i^{\star\top}\bigr)\Bigr\|
	            &&\tag{\cref{eq:sigma_cps}}
	    \\& \le \frac{1}{\gamma}\sqrt{\frac{1}{N}\sum_i \|z_i - z_i^\star\|^2}\,\sqrt{\frac{2}{N}\sum_i \|z_i\|^2 + \frac{2}{N}\sum_i \|z_i^\star\|^2}
	            &&\tag{\cref{lemma:outer_product_perturbation}, \cref{lemma:power_mean}}
	    \\& = o(1) \cdot O(1) = o(1) \quad P\text{-a.s.}
	            &&\tag{(ii), (iii)}
	}

	\paragraph{Step 2: $\*\pi^{(N)} \to \pis$ Replacement.}
	After Step 1 the vectors $w_i^\star$ are i.i.d., but the inclusion probabilities $\pi^{(N)}_i$ are not, because the design computes them from the whole population.
	We therefore replace $\pi^{(N)}_i$ by $\pis(X_i)$, which depends only on $X_i$, and show that the replacement changes $\Sigma_\bullet$ by $o(1)$ $P$-a.s.
	The input is~\cref{asmp:design_assumption_pi_lim}, which states that $\frac{1}{N}\sum_i (\pi^{(N)}_i - \pis(X_i))^2 \to 0$ $P$-a.s.
	Write $\alpha_i := (1 - \pi^{(N)}_i)/\pi^{(N)}_i$ as in Step 1 and $\alpha_i^\star := (1 - \pis(X_i))/\pis(X_i)$.
	Since $\alpha$ is a Lipschitz function of $\pi$ on $[\gamma, 1-\gamma]$, the weights are close on average.
	\flal{
	    &   \frac{1}{N}\sum_i (\alpha_i - \alpha_i^\star)^2
	        &&
	    \\& \le \frac{1}{\gamma^4} \cdot \frac{1}{N}\sum_i \bigl(\pi^{(N)}_i - \pis(X_i)\bigr)^2
	            &&\tag{\cref{lemma:phi_lipschitz}, $\pi^{(N)}_i, \pis(X_i) \in [\gamma, 1-\gamma]$}
	    \\& = o(1) \quad P\text{-a.s.}
	            &&\tag{\cref{asmp:design_assumption_pi_lim}}
	}

	\textit{Poisson case.} The PS covariance depends on $\pi$ only through the weights $\alpha_i$, so the vectors $w_i^\star$ stay fixed and only the weights change.
	\flal{
	    &   \bigl\|\Sigma_{\textup{ps}}(\*\pi^{(N)}, \theta_0, N) - \Sigma_{\textup{ps}}(\pis, \theta_0, N)\bigr\|
	        &&
	    \\& = \Bigl\|\frac{1}{N}\sum_i w_i^\star w_i^{\star\top} \bigl(\alpha_i - \alpha_i^\star\bigr)\Bigr\|
	            &&\tag{\cref{eq:sigma_ps}}
	    \\& \le \sqrt{\frac{1}{N}\sum_i \|w_i^\star\|^4}\,\sqrt{\frac{1}{N}\sum_i (\alpha_i - \alpha_i^\star)^2}
	            &&\tag{\cref{lemma:weight_perturbation}}
	    \\& = O(1) \cdot o(1) = o(1) \quad P\text{-a.s.}
	            &&\tag{Step 0 (b) with $k = 4$; display above}
	}

	\textit{CPS case.} The CPS covariance depends on $\pi$ in two places: through the weights $\alpha_i$, as in the Poisson case, and through the centered vectors, since both the factor $\pi_i$ and the ratio $\beta$ in $z_i = w_i^\star - \pi_i \beta$ change.
	Let $\beta^{(N)}_\star$ be the centering ratio with inputs $w_j^\star$ and weights $\pi^{(N)}_j$, as in Step 1, and $\beta^{(N)}_{\star,\infty}$ the same ratio with weights $\pis(X_j)$.
	Write $z^{(N)}_i := w_i^\star - \pi^{(N)}_i \beta^{(N)}_\star$ and $z^{(\infty)}_i := w_i^\star - \pis(X_i)\,\beta^{(N)}_{\star,\infty}$.
	We change one thing at a time: adding and subtracting $\frac{1}{N}\sum_i \alpha_i z^{(\infty)}_i z^{(\infty)\top}_i$ splits the difference into a term $T_1$ where only the vectors change and a term $T_2$ where only the weights change.
	\al{
	    & \Sigma_{\textup{cps}}(\*\pi^{(N)}, \theta_0, N) - \Sigma_{\textup{cps}}(\pis, \theta_0, N)
	        = \underbrace{\frac{1}{N}\sum_i \alpha_i \bigl(z^{(N)}_i z^{(N)\top}_i - z^{(\infty)}_i z^{(\infty)\top}_i\bigr)}_{=:\, T_1}
	        + \underbrace{\frac{1}{N}\sum_i z^{(\infty)}_i z^{(\infty)\top}_i \bigl(\alpha_i - \alpha_i^\star\bigr)}_{=:\, T_2}
	}
	$T_1$ is handled as in the CPS case of Step 1: show (i) $\beta^{(N)}_\star$ is close to $\beta^{(N)}_{\star,\infty}$, then (ii) $z^{(N)}_i$ is close to $z^{(\infty)}_i$ on average, then (iii) both have bounded moments.
	$T_2$ is handled as in the Poisson case above, and needs the fourth moments from (iii).
	Unlike in Step 1, the two ratios in (i) have different denominators, so (i) uses the ratio bound in~\cref{eq:ratio_identity} from the proof of~\cref{lemma:cps_centering_stability}, with $u_j = u'_j = w_j^\star$, $\eta_j = \pi^{(N)}_j$, and $\eta'_j = \pis(X_j)$.
	The bounds in~\cref{eq:AA_bound,eq:BB_bound} are deterministic inequalities, so they hold along each sample path.
	\flal{
	    &   \text{(i)} \quad \bigl\|\beta^{(N)}_\star - \beta^{(N)}_{\star,\infty}\bigr\|
	        &&
	    \\& \le \frac{1}{\gamma(1-\gamma)} \Bigl(\|A - A'\| + \bigl\|\beta^{(N)}_{\star,\infty}\bigr\|\,|B - B'|\Bigr)
	            &&\tag{\cref{eq:ratio_identity}}
	    \\& \le \frac{1}{\gamma(1-\gamma)} \Bigl(\sqrt{\frac{1}{N}\sum_j \|w_j^\star\|^2} + \bigl\|\beta^{(N)}_{\star,\infty}\bigr\|\Bigr) \sqrt{\frac{1}{N}\sum_j \bigl(\pi^{(N)}_j - \pis(X_j)\bigr)^2}
	            &&\tag{bounds on $\|A - A'\|$, $|B - B'|$ with $u = u'$}
	    \\& \le \frac{1}{\gamma(1-\gamma)} \Bigl(1 + \frac{1}{\gamma(1-\gamma)}\Bigr) \sqrt{\frac{1}{N}\sum_j \|w_j^\star\|^2}\,\sqrt{\frac{1}{N}\sum_j \bigl(\pi^{(N)}_j - \pis(X_j)\bigr)^2}
	            &&\tag{\cref{eq:beta_deterministic_bound}}
	    \\& = O(1) \cdot o(1) = o(1) \quad P\text{-a.s.}
	            &&\tag{Step 0 (b); \cref{asmp:design_assumption_pi_lim}}
	}
	\flal{
	    &   \text{(ii)} \quad \frac{1}{N}\sum_i \bigl\|z^{(N)}_i - z^{(\infty)}_i\bigr\|^2
	        &&
	    \\& = \frac{1}{N}\sum_i \bigl\|\pi^{(N)}_i \bigl(\beta^{(N)}_\star - \beta^{(N)}_{\star,\infty}\bigr) + \bigl(\pi^{(N)}_i - \pis(X_i)\bigr) \beta^{(N)}_{\star,\infty}\bigr\|^2
	            &&\tag{definition of $z^{(N)}_i, z^{(\infty)}_i$}
	    \\& \le 2\,\bigl\|\beta^{(N)}_\star - \beta^{(N)}_{\star,\infty}\bigr\|^2 + 2\,\bigl\|\beta^{(N)}_{\star,\infty}\bigr\|^2 \cdot \frac{1}{N}\sum_i \bigl(\pi^{(N)}_i - \pis(X_i)\bigr)^2
	            &&\tag{\cref{lemma:power_mean}, $\pi^{(N)}_i \le 1$}
	    \\& = o(1) \quad P\text{-a.s.}
	            &&\tag{(i); \cref{eq:beta_deterministic_bound} with Step 0 (b); \cref{asmp:design_assumption_pi_lim}}
	}
	\flal{
	    &   \text{(iii)} \quad \frac{1}{N}\sum_i \bigl\|z^{(\infty)}_i\bigr\|^4
	        &&
	    \\& \le \frac{8}{N}\sum_i \|w_i^\star\|^4 + 8\,\bigl\|\beta^{(N)}_{\star,\infty}\bigr\|^4
	            &&\tag{\cref{lemma:power_mean}, $\pis(X_i) \le 1$}
	    \\& = O(1) \quad P\text{-a.s.}
	            &&\tag{Step 0 (b); \cref{eq:beta_deterministic_bound} with Step 0 (b)}
	    \\& \text{and likewise } \frac{1}{N}\sum_i \bigl\|z^{(N)}_i\bigr\|^2 = O(1), \quad \frac{1}{N}\sum_i \bigl\|z^{(\infty)}_i\bigr\|^2 = O(1) \quad P\text{-a.s.}
	            &&\tag{same steps with $k = 2$}
	}
	\flal{
	    &   \text{(iv)} \quad \|T_1\|
	        &&
	    \\& \le \frac{1}{\gamma}\sqrt{\frac{1}{N}\sum_i \bigl\|z^{(N)}_i - z^{(\infty)}_i\bigr\|^2}\,\sqrt{\frac{2}{N}\sum_i \bigl\|z^{(N)}_i\bigr\|^2 + \frac{2}{N}\sum_i \bigl\|z^{(\infty)}_i\bigr\|^2}
	            &&\tag{\cref{lemma:outer_product_perturbation}, $\alpha_i \le 1/\gamma$; \cref{lemma:power_mean}}
	    \\& = o(1) \cdot O(1) = o(1) \quad P\text{-a.s.}
	            &&\tag{(ii), (iii)}
	    \\& \|T_2\| \le \sqrt{\frac{1}{N}\sum_i \bigl\|z^{(\infty)}_i\bigr\|^4}\,\sqrt{\frac{1}{N}\sum_i (\alpha_i - \alpha_i^\star)^2}
	            &&\tag{\cref{lemma:weight_perturbation}}
	    \\& = O(1) \cdot o(1) = o(1) \quad P\text{-a.s.}
	            &&\tag{(iii); first display of Step 2}
	}
	Hence $\Sigma_{\textup{cps}}(\*\pi^{(N)}, \theta_0, N) - \Sigma_{\textup{cps}}(\pis, \theta_0, N) = T_1 + T_2 = o(1)$ $P$-a.s.

	\paragraph{Step 3: SLLN at the Static $\pis$ Expression.}

	\textit{Poisson case.} The summands of
	\begin{equation*}
		\Sigma_{\textup{ps}}(\pis, \theta_0, N) \;=\; \frac{1}{N}\sum_{i=1}^N w_i^\star (w_i^\star)^\top\,\frac{1-\pis(X_i)}{\pis(X_i)}
	\end{equation*}
	are i.i.d.\ measurable functions of $(X_i, Y_i)$, with finite first moment $\E_m\bigl[\|w^\star (w^\star)^\top\|\,\frac{1-\pis(X)}{\pis(X)}\bigr] \le \frac{1}{\gamma}\E_m\|w^\star\|^2 < \infty$ by~\cref{asmp:score_integrability_theta0} (giving $\E_m\|w^\star\|^2 < \infty$ via Cauchy--Schwarz from $L^4$) and the overlap bound. The model SLLN therefore gives
	\begin{equation*}
		\Sigma_{\textup{ps}}(\pis, \theta_0, N) \;\xrightarrow[N\to\infty]{P\text{-a.s.}}\; \Sigma_{\infty,\textup{ps}}.
	\end{equation*}

	\textit{CPS case.} By the SLLN applied to numerator and denominator of $\beta^{(N)}_{\star,\infty}$ (each having finite first moment under~\cref{asmp:score_integrability_theta0} and the overlap bound), continuous mapping with the lower bound $\E_m[\pis(1-\pis)] \ge \gamma(1-\gamma) > 0$ on the denominator gives
	\begin{equation*}
		\beta^{(N)}_{\star,\infty} \;\xrightarrow{P\text{-a.s.}}\; \beta_\infty \;=\; \beta.
	\end{equation*}
	Set $\tilde z_i := w_i^\star - \pis(X_i)\,\beta_\infty$. Then $z^{(\infty)}_i - \tilde z_i = -\pis(X_i)(\beta^{(N)}_{\star,\infty} - \beta_\infty)$ and $\pis(X_i) \le 1-\gamma$, so
	\begin{equation*}
		\frac{1}{N}\sum_i \|z^{(\infty)}_i - \tilde z_i\|^2 \;\le\; (1-\gamma)^2\,\|\beta^{(N)}_{\star,\infty} - \beta_\infty\|^2 \;=\; o(1) \qquad P\text{-a.s.}
	\end{equation*}
	Step 2 (iii) gives $\frac{1}{N}\sum_i \|z^{(\infty)}_i\|^2 = O(1)$ $P$-a.s.
	Since $\|\tilde z_i\|^2 \le 2\|w_i^\star\|^2 + 2\|\beta\|^2$ and $\beta$ is fixed, Step 0 (b) gives $\frac{1}{N}\sum_i \|\tilde z_i\|^2 = O(1)$ $P$-a.s.
	Then~\cref{lemma:power_mean} gives $\frac{1}{N}\sum_i (\|z^{(\infty)}_i\| + \|\tilde z_i\|)^2 = O(1)$ $P$-a.s., and~\cref{lemma:outer_product_perturbation} with $\alpha_i = \alpha(\pis(X_i)) \le 1/\gamma$ gives
	\begin{equation*}
		\Sigma_{\textup{cps}}(\pis, \theta_0, N) - \frac{1}{N}\sum_i \tilde z_i \tilde z_i^\top\,\frac{1-\pis(X_i)}{\pis(X_i)} \;=\; o(1) \qquad P\text{-a.s.}
	\end{equation*}
	The remaining sum is i.i.d.\ in the model with expectation exactly $\Sigma_{\infty,\textup{cps}}$ and finite first moment by the same moment argument as in the Poisson case, so the SLLN gives convergence to $\Sigma_{\infty,\textup{cps}}$ $P$-a.s. Combining,
	\begin{equation*}
		\Sigma_{\textup{cps}}(\pis, \theta_0, N) \;\xrightarrow[N\to\infty]{P\text{-a.s.}}\; \Sigma_{\infty,\textup{cps}}.
	\end{equation*}

	\paragraph{Combining the Three Steps.} The triangle inequality gives, for $\bullet\in\{\textup{ps},\textup{cps}\}$,
	\begin{align*}
		\bigl\|\Sigma_\bullet(\*\pi^{(N)}, \theta_N, N) - \Sigma_{\infty,\bullet}\bigr\|  \;\le\; & \underbrace{\bigl\|\Sigma_\bullet(\*\pi^{(N)}, \theta_N, N) - \Sigma_\bullet(\*\pi^{(N)}, \theta_0, N)\bigr\|}_{o(1)\ P\text{-a.s.\ by Step 1}} \\
		                                                                                        & + \underbrace{\bigl\|\Sigma_\bullet(\*\pi^{(N)}, \theta_0, N) - \Sigma_\bullet(\pis, \theta_0, N)\bigr\|}_{o(1)\ P\text{-a.s.\ by Step 2}}    \\
		                                                                                        & + \underbrace{\bigl\|\Sigma_\bullet(\pis, \theta_0, N) - \Sigma_{\infty,\bullet}\bigr\|}_{o(1)\ P\text{-a.s.\ by Step 3}}.
	\end{align*}
	Therefore $\Sigma_\bullet(\*\pi^{(N)}, \theta_N, N) \xrightarrow{P\text{-a.s.}} \Sigma_{\infty,\bullet}$, completing the proof.
\end{proof}

\subsection{Main Result}
\label{app:main_result}

We are now prepared to prove our main result.

\MainResult*
\begin{proof}
	The proof applies~\cref{thm:rb_main}. Conditions (1)--(3) hold by~\cref{lemma:model_condition_verification}, so it remains to show that conditions (4) and (5) hold.

	\textbf{Condition (4).} We show that $\nabla \hat G_N(\theta)$ converges uniformly on $B(\theta_0)$ in design probability to $J(\theta)$ (the same limit produced in Condition~(2), which is non-stochastic in the design), for $P$-almost every $\omega$. From~\cref{def:estimating_equations}, differentiation under the finite sum gives
	\begin{equation*}
		\nabla \hat G_N(\theta) \;=\; \frac{1}{N}\sum_{i=1}^N \frac{s_i}{\pi^{(N)}_i}\bigl[\hiy - \hif\bigr] + \frac{1}{N}\sum_{i=1}^N \hif,
	\end{equation*}
	so that, recalling $\nabla G_N(\theta) = \frac{1}{N}\sum_i \hiy$,
	\begin{equation*}
		\nabla \hat G_N(\theta) - \nabla G_N(\theta) \;=\; \frac{1}{N}\sum_{i=1}^N \!\left(\frac{s_i}{\pi^{(N)}_i} - 1\right)\bigl[\hiy - \hif\bigr].
	\end{equation*}
	By the triangle inequality,
	\begin{align*}
		\sup_{\theta \in B(\theta_0)} \bigl\|\nabla \hat G_N(\theta) - J(\theta)\bigr\|
		 & \;\le\; \sup_{\theta \in B(\theta_0)} \bigl\|\nabla \hat G_N(\theta) - \nabla G_N(\theta)\bigr\| + \sup_{\theta \in B(\theta_0)} \bigl\|\nabla G_N(\theta) - J(\theta)\bigr\|,
	\end{align*}
	and the second supremum is $o(1)$ $P$-a.s.\ by Condition~(2). It remains to show the first supremum is $o_P(1)$ in design probability for $P$-a.e.\ $\omega$.

	\textit{Application of the design ULLN.} We apply~\cref{lemma:design_ulln} to the function class $\{u_\theta : \theta \in B(\theta_0)\}$ defined by $u_\theta(x,y) := \nabla g_\theta(x,y) - \nabla g_\theta(x, f(x))$, since
	\begin{equation*}
		\nabla \hat G_N(\theta) - \nabla G_N(\theta) \;=\; \frac{1}{N}\sum_i \frac{s_i}{\pi^{(N)}_i}\,u_\theta(X_i, Y_i) - \frac{1}{N}\sum_i u_\theta(X_i, Y_i),
	\end{equation*}
	which has the form $\hat G^u_N(\theta) - G^u_N(\theta)$ in the notation of~\cref{lemma:design_ulln}.

	\textit{Hypothesis (a) (pointwise design consistency).}
	Hypothesis (a) asks for HT consistency at each fixed $\theta$, which the consistency clause of~\cref{lemma:ht_design_vector} gives once the empirical fourth moment of $u_\theta(X_i, Y_i)$ is bounded.
	We bound this moment by a quantity that does not depend on $\theta$, so a single $P$-probability-one set of $\omega$ serves every $\theta \in B(\theta_0)$.
	\flal{
	    &   \|u_\theta(X_i, Y_i)\|^4
	        &&
	    \\& \le \bigparens{M(X_i, Y_i) + M(X_i, f(X_i))}^4
	            &&\tag{\cref{cor:jac_bound}, triangle inequality}
	    \\& \le 8\bigparens{M(X_i, Y_i)^4 + M(X_i, f(X_i))^4}
	            &&\tag{\cref{lemma:power_mean}}
	    \\& \implies \sup_{\theta \in B(\theta_0)} \frac{1}{N}\sum_i \|u_\theta(X_i, Y_i)\|^4 \le \frac{8}{N}\sum_i \bigparens{M(X_i, Y_i)^4 + M(X_i, f(X_i))^4} = O(1) \quad P\text{-a.s.}
	            &&\tag{SLLN; $\E_m[M^4] < \infty$ by \cref{asmp:score_lipschitz}}
	}
	Fix $\omega$ in the intersection of this set with the set on which~\cref{asmp:design_assumption_constraints_on_pi} holds; both have $P$-probability one, and neither depends on $\theta$.
	For each $\theta \in B(\theta_0)$, the consistency clause of~\cref{lemma:ht_design_vector}, applied to the vectorization of $w_i := u_\theta(X_i, Y_i) \in \R^{\ell \times \ell}$, gives $\hat G^u_N(\theta) - G^u_N(\theta) \cip 0$ in design probability.

	\textit{Hypothesis (b) (Lipschitz envelope).} Applying~\cref{asmp:jacobian_lipschitz} at $(x,y)$ and at $(x, f(x))$ separately,
	\begin{equation*}
		\|u_\theta(x,y) - u_{\theta'}(x,y)\| \;\le\; \bigl[M_J(x,y) + M_J(x, f(x))\bigr]\,\|\theta - \theta'\|,
	\end{equation*}
	with $\E_m[M_J(X,Y) + M_J(X, f(X))] < \infty$ by~\cref{asmp:jacobian_lipschitz}. Setting $\tilde M(x,y) := M_J(x,y) + M_J(x, f(x))$, which plays the role of $M$ in~\cref{lemma:design_ulln}, the SLLN gives $\bar M_N := \frac{1}{N}\sum_i \tilde M(X_i, Y_i) \to \E_m[\tilde M(X,Y)]$ $P$-a.s., so $\bar M_N$ is bounded along $P$-a.e.\ sample path.

	\textit{Conclusion.}~\cref{lemma:design_ulln} therefore yields
	\begin{equation*}
		\sup_{\theta \in B(\theta_0)} \bigl\|\nabla \hat G_N(\theta) - \nabla G_N(\theta)\bigr\| \;\cip\; 0
	\end{equation*}
	in design probability for $P$-a.e.\ $\omega$. Combining with the bound on $\sup_\theta\|\nabla G_N - J\|$ from Condition~(2), we obtain $\sup_{\theta \in B(\theta_0)} \|\nabla \hat G_N(\theta) - J(\theta)\| \cip 0$ in design probability for $P$-a.e.\ $\omega$. The limit $J(\theta)$ is non-stochastic in the design, and $J(\theta_0)$ is non-singular by~\cref{asmp:nonsingular_jacobian}, so Condition~(4) of~\cref{thm:rb_main} holds.

	\textbf{Condition (5).} We establish $\sqrt n\,\hat G_N(\theta_N) \cid \mathcal{N}_\ell(0, \Sigma_d)$ in design law for $P$-a.e.\ $\omega$, with $\Sigma_d = r\,\Sigma_{\infty,\bullet}$ non-stochastic in the super-population. By~\cref{def:estimating_equations},
	\begin{equation*}
		\hat G_N(\theta_N) \;=\; \frac{1}{N}\sum_i \frac{s_i}{\pi^{(N)}_i}\bigl[g_{\theta_N}(X_i, Y_i) - g_{\theta_N}(X_i, f(X_i))\bigr] + \frac{1}{N}\sum_i g_{\theta_N}(X_i, f(X_i)).
	\end{equation*}
	Since $\theta_N$ solves $G_N(\theta_N) = 0$ by definition, $\frac{1}{N}\sum_i g_{\theta_N}(X_i, Y_i) = 0$, so
	\begin{align*}
		\frac{1}{N}\sum_i g_{\theta_N}(X_i, f(X_i)) \;=\; \frac{1}{N}\sum_i g_{\theta_N}(X_i, f(X_i)) - \underbrace{\frac{1}{N}\sum_i g_{\theta_N}(X_i, Y_i)}_{=0} \;=\; -\frac{1}{N}\sum_i w_i(\theta_N),
	\end{align*}
	where we define $w_i(\theta) := g_\theta(X_i, Y_i) - g_\theta(X_i, f(X_i))$.
	Substituting and combining,
	\begin{align*}
		\hat G_N(\theta_N) \;=\; \frac{1}{N}\sum_i \frac{s_i}{\pi^{(N)}_i}\,w_i(\theta_N) - \frac{1}{N}\sum_i w_i(\theta_N) \;=\; \frac{1}{N}\sum_i \!\left(\frac{s_i}{\pi^{(N)}_i} - 1\right) w_i(\theta_N) \;=\; \hat\mu^w_{\*\pi} - \mu^w_N,
	\end{align*}
	the Horvitz--Thompson residual of~\cref{lemma:ht_design_vector} with $w_i = w_i(\theta_N)$.

	\textit{Design CLT at rate $\sqrt N$.}
	\Cref{lemma:ht_design_vector} needs all of its hypotheses to hold at a single $\omega$, so we fix $\omega$ in the intersection of three $P$-probability-one sets: the set on which~\cref{asmp:design_assumption_constraints_on_pi} holds, the set on which $\frac{1}{N}\sum_i \|w_i(\theta_N)\|^4 = O(1)$ (by Step 0 (c) in the proof of~\cref{lemma:convergence_of_covariance}), and the set on which $\Sigma_\bullet(\*\pi^{(N)}, \theta_N, N) \to \Sigma_{\infty,\bullet}$ (by~\cref{lemma:convergence_of_covariance}).
	\Cref{lemma:ht_design_vector} is stated for a fixed $w$, but its proof uses, at each $N$, only the values $w_1, \ldots, w_N$ and the fourth-moment bound, so it applies to $w_i(\theta_N)$, whose values change with $N$.
	The limit $\Sigma_{\infty,\bullet}$ is positive definite by hypothesis, so the asymptotic-normality clause of~\cref{lemma:ht_design_vector}, applied to $w_i(\theta_N)$ at this $\omega$, gives
	\al{
	    & \sqrt N\,\hat G_N(\theta_N) = \sqrt N\,(\hat\mu^w_{\*\pi} - \mu^w_N) \cid \mathcal{N}_\ell(0, \Sigma_{\infty,\bullet}) \quad \text{in design law, for $P$-a.e.\ } \omega.
	}

	\textit{Conversion to rate $\sqrt n$.} Here $n = n_N$ is the budget in~\cref{asmp:design_assumption_constraints_on_pi}, which is deterministic under both designs. Under PS it is the expected sample size $\sum_i \pi^{(N)}_i$, not the realized size $\sum_i s_i$. Hence $\sqrt n\,\hat G_N(\theta_N) = \sqrt{n_N/N}\,\cdot\,\sqrt N\,\hat G_N(\theta_N)$, where $\sqrt{n_N/N} \to \sqrt r$ is deterministic, so the display above and Slutsky's lemma give
	\begin{equation*}
		\sqrt n\,\hat G_N(\theta_N) \;\cid\; \mathcal{N}_\ell(0, r\,\Sigma_{\infty,\bullet}) \qquad \text{in design law for $P$-a.e.\ } \omega,
	\end{equation*}
	verifying Condition~(5) of~\cref{thm:rb_main} with $\Sigma_d = r\,\Sigma_{\infty,\bullet}$ (non-stochastic in the super-population).

	\textit{Conclusion.} Conditions (1)--(5) of~\cref{thm:rb_main} are satisfied, so the theorem yields
	\begin{equation*}
		\sqrt n\,(\hat\theta_N - \theta_0) \;\cid\; \mathcal{N}\bigl(0,\,J^{-1}(\Sigma_d + r\,\Sigma_m)J^{-\top}\bigr)
	\end{equation*}
	on the product space, with $\Sigma_d, \Sigma_m$ as identified above, which concludes the proof.
\end{proof}

\subsection{Proofs of Corollaries}\label{app:proofs}

\subsubsection{Design Variance Reduction for the Mean}\label{app:mean_variance_reduction}

\VarianceReductionCPSMean*
\begin{proof}

	Recall from~\cref{lemma:ht_design} that $z_i = w_i - \pi_i \beta$, where
	\al{
		& \beta = \frac{\sum_{i=1}^N w_i (1 - \pi_i)}{d(\*\pi)},
		\qquad
		d(\*\pi) = \sum_{i=1}^N \pi_i (1 - \pi_i).
	}
	\flal{
	    &   \sigma^2_{\textup{ps}}(\*\pi, N) - \sigma^2_{\textup{cps}}(\*\pi, N)
	        = \frac{1}{N} \sum_{i=1}^N (w_i^2 - z_i^2) \frac{1-\pi_i}{\pi_i}
	        &&\tag{definitions in~\cref{lemma:ht_design}}
	    \\& = \frac{1}{N} \sum_{i=1}^N \bigparens{w_i^2 - (w_i - \pi_i \beta)^2} \frac{1-\pi_i}{\pi_i}
	            &&\tag{$z_i = w_i - \pi_i \beta$}
	    \\& = \frac{1}{N} \sum_{i=1}^N (2 w_i \pi_i \beta - \pi_i^2 \beta^2) \frac{1-\pi_i}{\pi_i}
	            &&\tag{expand the square}
	    \\& = \frac{2\beta}{N} \sum_{i=1}^N w_i (1-\pi_i)
	            - \frac{\beta^2}{N} \sum_{i=1}^N \pi_i(1-\pi_i)
	            &&\tag{distribute, $\pi_i$ cancels}
	    \\& = \frac{2\beta}{N} \cdot \beta\, d(\*\pi) - \frac{\beta^2}{N} \cdot d(\*\pi)
	            &&\tag{$\sum_{i=1}^N w_i(1-\pi_i) = \beta\, d(\*\pi)$ and $\sum_{i=1}^N \pi_i(1-\pi_i) = d(\*\pi)$}
	    \\& = \frac{\beta^2\, d(\*\pi)}{N}
	            &&
	    \\& = \frac{\bigparens{\sum_{i=1}^N w_i(1-\pi_i)}^2}{N\, d(\*\pi)}
	            &&\tag{write out $\beta$}
	    \\& = \frac{\bigparens{\frac{1}{N}\sum_{i=1}^N w_i(1-\pi_i)}^2}{\frac{1}{N}\sum_{i=1}^N \pi_i(1-\pi_i)}
	        \ \geq\ 0
	            &&\tag{divide top and bottom by $N^2$}
	}

\end{proof}

\subsubsection{Design Covariance Reduction for Z-Estimation}\label{app:M_variance_reduction}

Recall the following corollary from the main text.
\MatrixVarianceReductionCPS*

\begin{proof}
Write $A := \E_m[w^\star(1-\pis)]$ and $b := \E_m[\pis(1-\pis)]$, so that $\beta = A/b$.
\flal{
    &   \Sigma_{\infty,\textup{ps}} - \Sigma_{\infty,\textup{cps}}
        = \E_m \bigbracks{
            \bigl(w^\star w^{\star\top} - (w^\star-\pis\beta)(w^\star-\pis\beta)^{\top}\bigr)
            \tfrac{1-\pis}{\pis}
            }
        &&\tag{definition of $\Sigma_{\infty,\bullet}$}
    \\& = \E_m \bigbracks{
            \pis\bigl(\beta w^{\star\top} + w^\star\beta^{\top} - \pis\beta\beta^{\top}\bigr)
            \tfrac{1-\pis}{\pis}
            }
            &&\tag{expand; $w^\star w^{\star\top}$ cancels}
    \\& = \E_m \bigbracks{
            \bigl(\beta w^{\star\top} + w^\star\beta^{\top} - \pis\beta\beta^{\top}\bigr)(1-\pis)
            }
            &&\tag{$\pis$ cancels}
    \\& = \beta A^{\top} + A\beta^{\top} - \beta\beta^{\top} b
            &&\tag{linearity; $\beta$ is a constant}
    \\& = \tfrac{1}{b}AA^{\top} + \tfrac{1}{b}AA^{\top} - \tfrac{1}{b}AA^{\top}
            &&\tag{$\beta = A/b$}
    \\& = \frac{AA^{\top}}{b}
        \;\succeq\; 0
            &&\tag{positive semi-definite of rank at most one, $b > 0$}
}
\end{proof}

\subsection{Derivation of the Covariance Between the HT Estimator and the PS Sample Size}\label{app:covariance}

Let $\*s$ be drawn from the Poisson Sampling design $\p_{\textup{ps}(\*\pi)}$ with $\sum_{i=1}^N \pi_i = n$, let $\tilde n := \sum_i s_i$ denote the realized sample size, and let $\hat\mu^w_{\*\pi} := \frac{1}{N}\sum_i \frac{w_i s_i}{\pi_i}$ denote the HT estimator of $\mu^w_N = \frac{1}{N}\sum_i w_i$. Their design covariance is
\begin{flalign*}
    &   \cov(\hat\mu^w_{\*\pi}, \tilde{n})
        = \cov\bigparens{\frac{1}{N} \sum_i \frac{w_i s_i}{\pi_i}, \sum_i s_i}
        = \frac 1N \sum_i \sum_j \frac{w_i }{\pi_i} \E_d[ s_is_j] - \frac{n}{N}\sum_iw_i
        &&\tag{$\cov(A,B) = \E[AB] - \E[A]\E[B]$, $\E_d[\hat\mu^w_{\*\pi}] = \frac1N\sum_i w_i$, $\E_d[\tilde n] = n$}
    \\& \text{where } \sum_i\sum_j \frac{w_i}{\pi_i} \E_d[s_is_j]
            = \sum_i\sum_{j\neq i} \frac{w_i}{\pi_i} \E_d[s_is_j]
            + \sum_i \frac{w_i}{\pi_i} \E_d[s_i^2]
            &&
    \\& \qquad = \sum_i\sum_{j\neq i} \frac{w_i}{\pi_i}\pi_i\pi_j + \sum_i w_i
            &&\tag{$s_i \ci_{\textup{ps}} s_j$ and $\E_d[s_i^2] = \E_d[s_i] = \pi_i$}
    \\& \qquad = \sum_i w_i (n-\pi_i) + \sum_i w_i
            &&\tag{$\sum_j \pi_j = n$}
    \\& \qquad = n\sum_i w_i - \sum_i w_i\pi_i + \sum_i w_i
            &&
    \\& \implies \cov(\hat\mu^w_{\*\pi}, \tilde{n})
            = \frac1N\bigparens{n\sum_i w_i - \sum_i w_i\pi_i + \sum_i w_i} - \frac nN\sum_i w_i
            = \frac1N\sum_i w_i(1-\pi_i)
            &&
\end{flalign*}

\end{document}